\documentclass[acmsmall,screen]{acmart}
\usepackage{array}
\usepackage{textcomp}
\usepackage{stfloats}
\usepackage{url}
\usepackage{verbatim}
\usepackage{graphicx}
\usepackage{textcomp}
\usepackage{xcolor}
\usepackage{bm}
\usepackage{mathtools}
\usepackage{algorithm,algpseudocode}
\usepackage{footnote}
\definecolor{g}{rgb}{0.925, 0.957, 0.831}
\definecolor{p}{rgb}{0.980,0.910,0.922}
\definecolor{b}{rgb}{0.776,0.918,0.980}
\usepackage{multirow}
\usepackage{colortab}
\usepackage{colortbl}
\usepackage{arydshln}
\usepackage{url}
\usepackage{tablefootnote}

\newcommand{\swin}{Fuzzy-UCS$_\text{SWIN}$}
\newcommand{\vote}{Fuzzy-UCS$_\text{VOTE}$}
\newcommand{\ds}{Fuzzy-UCS$_\text{DS}$}
\newcommand{\all}{Fuzzy-UCS$_*$}
\newcommand{\bhline}[1]{\noalign{\hrule height #1}} 
\usepackage{color}

\AtBeginDocument{%
  }

\setcopyright{acmlicensed}
\copyrightyear{2025}
\acmYear{2025}
\acmDOI{10.1145/3717613}

\acmJournal{TELO}
\acmVolume{1}
\acmNumber{1}
\acmArticle{1}
\acmMonth{2}

\begin{document}

\title{A Class Inference Scheme With Dempster-Shafer Theory for Learning Fuzzy-Classifier Systems}


\author{Hiroki Shiraishi}
\affiliation{%
\department{Faculty of Engineering}
  \institution{Yokohama National University}
  \city{Yokohama 240-8501}
  \country{Japan}}
\email{shiraishi-hiroki-yw@ynu.jp}
\orcid{0000-0001-8730-1276}

\author{Hisao Ishibuchi}
\authornote{Corresponding authors.}
\affiliation{%
\department{Department of Computer Science and Engineering}
  \institution{Southern University of Science and Technology}
  \city{Shenzhen 518055}
  \country{China}
}
\email{hisao@sustech.edu.cn}
\orcid{0000-0001-9186-6472}

\author{Masaya Nakata}
\authornotemark[1]
\affiliation{%
\department{Faculty of Engineering}
  \institution{Yokohama National University}
  \city{Yokohama 240-8501}
  \country{Japan}}
\email{nakata-masaya-tb@ynu.ac.jp}
\orcid{0000-0003-3428-7890}


\begin{abstract}

The decision-making process {significantly influences the predictions of machine learning models. This is especially important in rule-based systems such as Learning Fuzzy-Classifier Systems (LFCSs) where the selection and application of rules directly determine prediction accuracy and reliability. LFCSs combine evolutionary algorithms with supervised learning to optimize fuzzy classification rules, offering enhanced interpretability and robustness. Despite these advantages,} research on improving decision-making mechanisms (i.e., class inference schemes) in LFCSs remains limited. Most LFCSs use voting-based or single-winner-based inference schemes. These schemes rely on classification performance on training data and may not perform well on unseen data, risking overfitting. To address these limitations, this article introduces a novel class inference scheme for LFCSs based on the Dempster-Shafer Theory of Evidence (DS theory). The proposed scheme handles uncertainty well. By using the DS theory, the scheme calculates belief masses (i.e., measures of belief) for each specific class and the ``I don't know'' state from each fuzzy rule and infers a class from these belief masses. Unlike the conventional schemes, the proposed scheme also considers the ``I don't know'' state that reflects uncertainty, thereby improving the transparency and reliability of LFCSs. Applied to a variant of LFCS (i.e., Fuzzy-UCS), {the proposed scheme demonstrates statistically significant improvements in terms of test {macro F1 scores} across 30 real-world datasets compared to conventional voting-based and single-winner-based {fuzzy inference} schemes.} It forms smoother decision boundaries, provides reliable confidence measures, and enhances the robustness and generalizability of LFCSs in real-world applications.
Our implementation is available at \url{https://github.com/YNU-NakataLab/jUCS}. An extended abstract related to this work is available at \url{https://doi.org/10.36227/techrxiv.174900814.43483543/v1} \cite{shiraishi2025evidential}.
\end{abstract}


\received{3 June 2024}
\received[revised]{22 October 2024, 5 February 2025}
\received[accepted]{9 February 2025}

\maketitle

\section{Introduction}

The decision-making process is the foundation of machine learning models and plays a crucial role in their effectiveness and efficiency \cite{breiman2001random}. Accurate decision-making ensures that a model can generalize from training data and correctly classify new data points. This is especially important in rule-based systems where the selection and application of rules directly impact the accuracy and reliability of predictions \cite{wilson1995xcs}.

\textit{Learning Classifier Systems} (LCSs) \cite{holland1986possibilities} are a prominent evolutionary machine learning paradigm that combines evolutionary algorithms with supervised learning to optimize a set of rules for classification tasks. An LCS can construct adaptive and interpretable rules (i.e., local models) that learn from their environment and adapt over time. LCSs have been applied in various fields, such as data mining \cite{zhang2021lcs}, automatic design of ensemble modeling \cite{preen2021autoencoding}, {health informatics \cite{yazdani2020diagnosing}, and bioinformatics \cite{woodward2024survival}}, demonstrating their versatility and effectiveness.
However, LCSs typically utilize crisp rules, which, although effective, may lack the interpretability and robustness necessary to handle real-world data characterized by uncertainty and noise \cite{shiraishi2023fuzzy}. To address these limitations, we focus on \textit{Learning Fuzzy-Classifier Systems} (LFCSs) \cite{valenzuela1991fuzzy}, an extension of LCS that optimizes fuzzy rules instead of crisp ones.
Fuzzy rules offer several advantages over crisp rules. Firstly, fuzzy rules use linguistic terms, such as \textit{small}, \textit{medium}, and \textit{large}, to represent the antecedents of the rules, providing higher interpretability compared to crisp rules \cite{ishibuchi2004classification,fernandez2019evolutionary}. {In this article, we use the term \textit{interpretability} to refer to the ease with which humans can understand and intuitively grasp the meaning of fuzzy rules. Zadeh \shortcite{zadeh1975concept} suggested that linguistic labels used in fuzzy rules are often more intuitive for humans to comprehend compared to precise numerical ranges (i.e., quantitative terms) used in crisp rules.} This interpretability is crucial in fields where transparency is essential, such as healthcare and finance \cite{adadi2018peeking}. Secondly, fuzzy rules exhibit greater robustness when dealing with uncertainty and noise in the data \cite{shoeleh2011towards}. This robustness is vital for real-world applications where data is often incomplete and imprecise.

Various improvements have been made to LFCSs to date. For example, Shiraishi et al. \shortcite{shiraishi2023fuzzy} proposed a mechanism that evolutionary optimizes the shapes of membership functions for each rule according to the problem being solved. Additionally, Nojima et al. \shortcite{nojima2023fuzzy} introduced a reject option to enhance LFCS reliability by discarding low-confidence inference results. Recently, several LFCSs have been proposed to address multi-label classification problems \cite{omozaki2020multiobjective,omozaki2022evolutionary}, significantly expanding the applicability of LFCSs.  However, research on improving the decision-making mechanism, i.e., class inference schemes, is still limited. Even with optimal rules, an inappropriate class inference scheme can lead to incorrect outputs. Conversely, an appropriate scheme can produce correct results even without optimal rules. Thus, focusing on the decision-making mechanism is important for enhancing LFCS performance.

In LFCSs, there are two primary schemes for class inference: a \textit{voting-based inference} \cite{bardossy1995fuzzy} and a \textit{single-winner-based inference} \cite{ishibuchi1999performance}. The voting-based inference is an ensemble-based approach that makes decisions based on the consensus of multiple rules, potentially enhancing robustness and accuracy. In contrast, the single-winner-based inference relies solely on the rule that performs the best on training data, making it straightforward. 
To the best of our knowledge, most LFCSs (e.g., \cite{li2024fuzzy,shoeleh2011towards,tadokoro2021xcs,nojima2015simple,orriols2011fuzzy}) employ either the voting-based inference or single-winner-based inference. 
However, both schemes infer a class based on the quality of rules, as measured by a fitness function that reflects classification accuracy on training data.
While these schemes work well for training data, they may not necessarily translate to high performance on unseen data. There is a risk that rules optimized for training data may overfit, capturing noise and specific patterns that do not generalize well to new data points.

To mitigate these issues, this article introduces a novel class inference scheme for LFCSs based on the \textit{Dempster-Shafer Theory of Evidence} (DS theory) \cite{dempster1967upper,shafer1976mathematical}, which is a robust framework for reasoning and decision making under uncertainty and conflicting information \cite{yager2008classic}. 
The DS theory assigns a measure of belief, known as \textit{belief mass}, to information from various independent knowledge sources and combines them using \textit{Dempster's rule of combination} \cite{shafer1976mathematical}. 
The DS theory can be a good candidate for increasing the robustness and generalizability of LFCS, especially since it deals with uncertainty and is adept at aggregating multiple lines of evidence.

Our goal is to integrate the DS theory into the LFCS decision-making process to improve the system's ability to make reliable and accurate predictions for unseen data points. Specifically, in the proposed class inference scheme, when a new data point is presented, belief masses are calculated from the membership degree and weight vector of each fuzzy rule. The inference result is a consensus derived from combining these belief masses using Dempster's rule of combination. Unlike the conventional voting- and single-winner-based schemes, which focus solely on which specific class each rule supports, our proposed scheme also considers ``floating votes'', i.e., ``I don't know'' belief masses indicating uncertainty. This feature enhances the transparency and reliability of the LFCS inference results.
In this article, the proposed class inference scheme is applied to the \textit{sUpervised Fuzzy-Classifier System} (Fuzzy-UCS) \cite{orriols2008fuzzy}, an LFCS designed for single-label classification problems.

The main contributions of this article are as follows:
\begin{enumerate}
\item To the best of our knowledge, this article is the first to incorporate the DS theory into LFCSs. 

\item Our experimental results demonstrate that Fuzzy-UCS with the proposed class inference scheme significantly improves classification performance for real-world problems characterized by uncertainty and incomplete information. This improvement is notable when compared to two variants of Fuzzy-UCS that use the conventional voting- and single-winner-based class inference schemes.

\item The proposed class inference scheme forms smoother decision boundaries compared to conventional class inference schemes. This feature helps to achieve more accurate classification results in Fuzzy-UCS.

\item The proposed class inference scheme provides belief masses indicating the reliability of the classification result. For instance, if a new data point is presented in a region with sparse training data points, the inference scheme assigns a larger ``I don't know'' belief mass, reflecting lower confidence in the inference.
\end{enumerate}

The structure of this article is as follows: Section \ref{sec: related work} reviews relevant literature. Sections \ref{sec: Fuzzy-UCS} and \ref{sec: ds theory} provide overviews of Fuzzy-UCS and the DS theory, respectively. Section \ref{sec: proposed method} details the proposed class inference scheme for LFCSs. Section \ref{sec: experiment} conducts comparative experiments using 30 real-world datasets for classification tasks. The experimental results are then evaluated and discussed. Section \ref{sec: analysis} provides analytical results to investigate the characteristics of each class inference scheme. Finally, Section \ref{sec: concluding remarks} concludes this article.

\section{Related Work}
\label{sec: related work}
\subsection{Class Inference Schemes in LFCSs}
\label{ss: related work class inference}

In LFCSs, two predominant class inference schemes have been established \cite{ishibuchi2004classification,orriols2008fuzzy}. The first, a \textit{voting-based inference} \cite{bardossy1995fuzzy}, involves a decision-making process that relies on the aggregated product of membership degrees and fitness values from multiple fuzzy rules. The second, a \textit{single-winner-based inference} \cite{ishibuchi1999performance}, designates the class of the single fuzzy rule with the largest product of a membership degree and a fitness value as the output. These schemes have found contemporary applications in modern LFCSs, such as \cite{omozaki2022evolutionary} and \cite{shiraishi2023fuzzy}. Notably, the voting-based inference has been adapted for nonfuzzy-classifier systems and is now a standard class inference scheme in leading LCSs, such as XCS \cite{wilson1995xcs}, UCS \cite{bernado2003accuracy}, ExSTraCS \cite{urbanowicz2015exstracs}, and ASCS \cite{liu2020absumption}.

\subsection{Dempster-Shafer Theory in Rule-Based Systems}
{The integration of the DS theory into rule-based systems has been explored across various domains. 
While this article focuses on its application within LFCSs, the DS theory has also been employed in expert systems to enhance decision-making under uncertainty.

For example, Guth \shortcite{guth1988uncertainty} applied the DS theory to analyze uncertainty propagation in rule-based expert systems, demonstrating its effectiveness in handling both probabilistic rule assignments and uncommitted belief in Boolean logic frameworks.
Beynon et al. \shortcite{beynon2001expert} developed an expert system using the DS theory combined with the Analytic Hierarchy Process for multi-criteria decision-making, showcasing its application in real estate property evaluation. Similarly, Hutasuhut et al. \shortcite{hutasuhut2018expert} developed an expert system using the DS theory to detect stroke, demonstrating its effectiveness in medical diagnostics by providing optimal diagnoses based on patient symptoms.

These examples highlight how the DS theory can improve decision-making in various fields. 
This wide applicability provides a solid basis for using the DS theory in LFCSs, as we proposed in this article.
}
\subsection{Dempster-Shafer Theory in Machine Learning}
The integration of the DS theory into machine learning was first exemplified by Den{\oe}ux \shortcite{denoeux1995k}, who enhanced the traditional k-nearest neighbors (kNN) method by combining pieces of evidence from each data point's neighborhood. Following this pioneering work, numerous machine learning techniques, such as decision trees \cite{quinlan1993c}, support vector machines \cite{boser1992training}, fuzzy c-means \cite{bezdek2013pattern},  Na{\"i}ve Bayes \cite{john1995estimating}, and LCS, have leveraged the strengths of the DS theory in managing uncertainty and merging evidence from diverse sources \cite{li2019evidential,zhou2015structural,mulyani2016new,masson2008ecm,ferjani2022evidential}.

A notable recent advance is the introduction of the DS theory into deep learning by Sensoy et al. \shortcite{sensoy2018evidential}, which quantifies the uncertainty in classification results. Their methodology differs from traditional neural networks in that it produces parameters of a Dirichlet distribution, based on the DS theory, to model both predictions and uncertainties in classification. These parameters facilitate the calculation of predictive probabilities for each class and uncertainty.

The DS theory has been employed not only in individual machine learning models but also for integrating outputs from multiple models \cite{rahmati2016application,nachappa2020flood,belmahdi2023application}. For instance, Belmahdi et al. \shortcite{belmahdi2023application} demonstrated that combining the inference results of three classifiers (artificial neural network, support vector machine, and random forest \cite{breiman2001random}) using the DS theory can lead to improved classification results compared to those relying on a single classifier.

Although the DS theory has been successfully applied in machine learning, its use in LFCSs has not been explored.

\section{Fuzzy-UCS}
\label{sec: Fuzzy-UCS}
Fuzzy-UCS \cite{orriols2008fuzzy} is a Michigan-style LFCS that integrates supervised learning with a steady-state genetic algorithm (GA) \cite{goldberg2002design} to evolve fuzzy rules online. Essentially, Fuzzy-UCS can be viewed as a ``fuzzified'' adaptation of UCS \cite{bernado2003accuracy}. This system alternates between two primary phases: the training (\textit{exploration}) phase and the test (\textit{exploitation}) phase. Within the training phase, Fuzzy-UCS searches for accurate and maximally general rules. Conversely, during the test phase, the system applies these acquired rules to infer a class for a new unlabeled data point.
This section briefly explains Fuzzy-UCS for a $d$-dimensional $n$-class single-label classification problem with a class label set $\mathcal{C}=\{c_i\}_{i=1}^n$.
For detailed explanations, kindly refer to \cite{orriols2008fuzzy}.

\subsection{Knowledge Representation}
\begin{figure}[t]
\centerline{\includegraphics[width=0.6\linewidth]{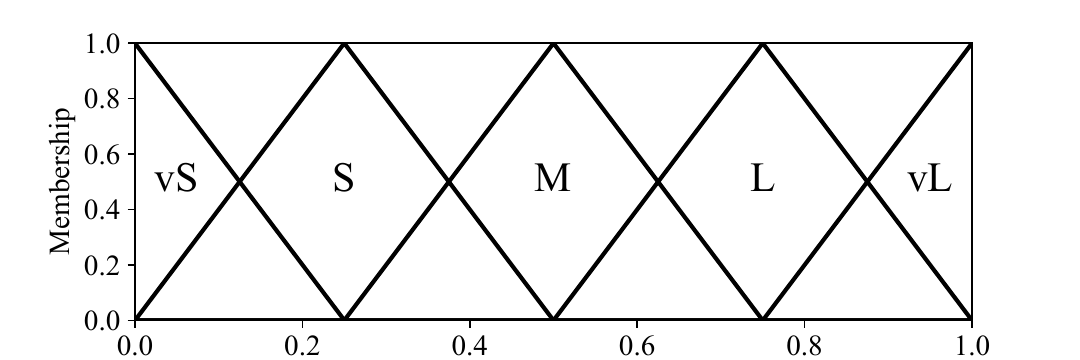}}
\caption{An example of homogeneous linguistic discretization of the domain interval $[0,1]$. The meaning of each term is as follows: vS: \textit{very small}, S: \textit{small}, M: \textit{medium}, L: \textit{large}, vL: \textit{very large}.}
\label{fig: triangular_membership}
\end{figure}

A $d$-dimensional fuzzy rule $k$  is expressed by:
\begin{eqnarray}
\label{eq: rule_repr}
    {\rm \textbf{IF}} \; x_1 \; \text{is} \; A_1^k \; \text{and} \; \cdot\cdot\cdot\; \text{and} \; x_d \; \text{is} \; A_d^k \; {\rm \textbf{THEN}} \; c^k \; {\rm \textbf{WITH}} \; w^k,
\end{eqnarray}
where $\mathbf{A}^k=(A_1^k,..., A_d^k)$ is a \textit{rule-antecedent} vector, $c^k\in\mathcal{C}=\{c_1,...,c_n\}$ is a \textit{rule-consequent} class, and $w^k\in[0,1]$ is a \textit{rule-weight}. 
Each variable $x_i$ is conditioned by a fuzzy set $A_i$ for $\forall i \in \{1,...,d\}$.
In this article, as the antecedent set, we use a triangular-shaped membership function with five linguistic terms in Fig. \ref{fig: triangular_membership}.
{The Fuzzy-UCS framework can optimize the shape of membership functions by using the GA to search for the optimal combination of linguistic labels for each rule.
			}

The \textit{membership degree} $\mu_{\mathbf{A}^k}(\mathbf{x})\in[0,1]$ of an input vector $\mathbf{x}\in[0,1]^d$ with the rule $k$ in Eq. (\ref{eq: rule_repr}) is calculated by the product operator: 
\begin{equation}
\label{eq: matching_degree}
    \mu_{\mathbf{A}^k}(\mathbf{x})=\prod_{i=1}^d{\mu_{A^k_i}}(x_i),
\end{equation}
where ${\mu_{A^k_i}}:[0,1] \rightarrow [0,1]$ is the membership function of the fuzzy set $A_i^k$. 
If the value of $x_i$ is unknown, the system handles the missing value by specifying $\mu_{{A^k_i}}(x_i)=1$.

Each rule $k$ has six primary parameters: 
(i) a \textit{fitness} $F^k\in(-1, 1]$, which reflects the classification accuracy of rule $k$\footnote{In LCSs (such as UCS and ExSTraCS), fitness values range between (0, 1]. In contrast, in LFCSs like Fuzzy-UCS, fitness values range between (-1, 1] due to the fitness calculation method introduced by \cite{ishibuchi2005rule}, as described in Section \ref{sss: training phase Fuzzy-UCS}. \cite{ishibuchi2005rule} demonstrated that employing fitness values within the (-1, 1] range enhances the test classification accuracy of an LFCS compared to the (0, 1] range. Consequently, Fuzzy-UCS also adopts fitness values within the (-1, 1] range in their original work \cite{orriols2008fuzzy}. For further details, kindly refer to Appendix \ref{sec: sup analysis of fitness range impact}.}; 
(ii) a \textit{weight vector} $\mathbf{v}^k = (v_i^k)_{i=1}^n \in [0, 1]^n$, indicating the confidence with which rule $k$ predicts each class $c_i$ for a matched input (the largest element of $\mathbf{v}^k$ is used as $w^k$ in Eq. (\ref{eq: rule_repr}));
(iii) a \textit{correct matching vector} $\mathbf{cm}^k = ({\rm cm}_i^k)_{i=1}^{n}\in(\mathbb{R}_0^+)^n$, where each element ${\rm cm}_i^k$ is the summation of the membership degree for training data points from class $c_i$;
(iv) an \textit{experience} ${\rm exp}^k\in\mathbb{R}^+_0$, which computes the accumulated contribution of rule $k$ in order to classify training data points; 
(v) a \textit{numerosity} ${\rm num}^k\in\mathbb{N}_0$, which indicates the number of rules that have been subsumed (i.e., the number of successful \textit{subsumption} operations, detailed in Section \ref{sss: training phase Fuzzy-UCS});
and (vi) a \textit{time stamp} ${\rm ts}^k\in\mathbb{N}$, which denotes the time-step of the last execution of a steady-state GA to create new rules, where rule $k$ was considered as a candidate to be a parent.
These parameters are continuously updated throughout the training process.

\subsection{Algorithm}
\label{ss: fuzzy-ucs mechanism}

\begin{figure}[t]
\centerline{\includegraphics[width=\linewidth]{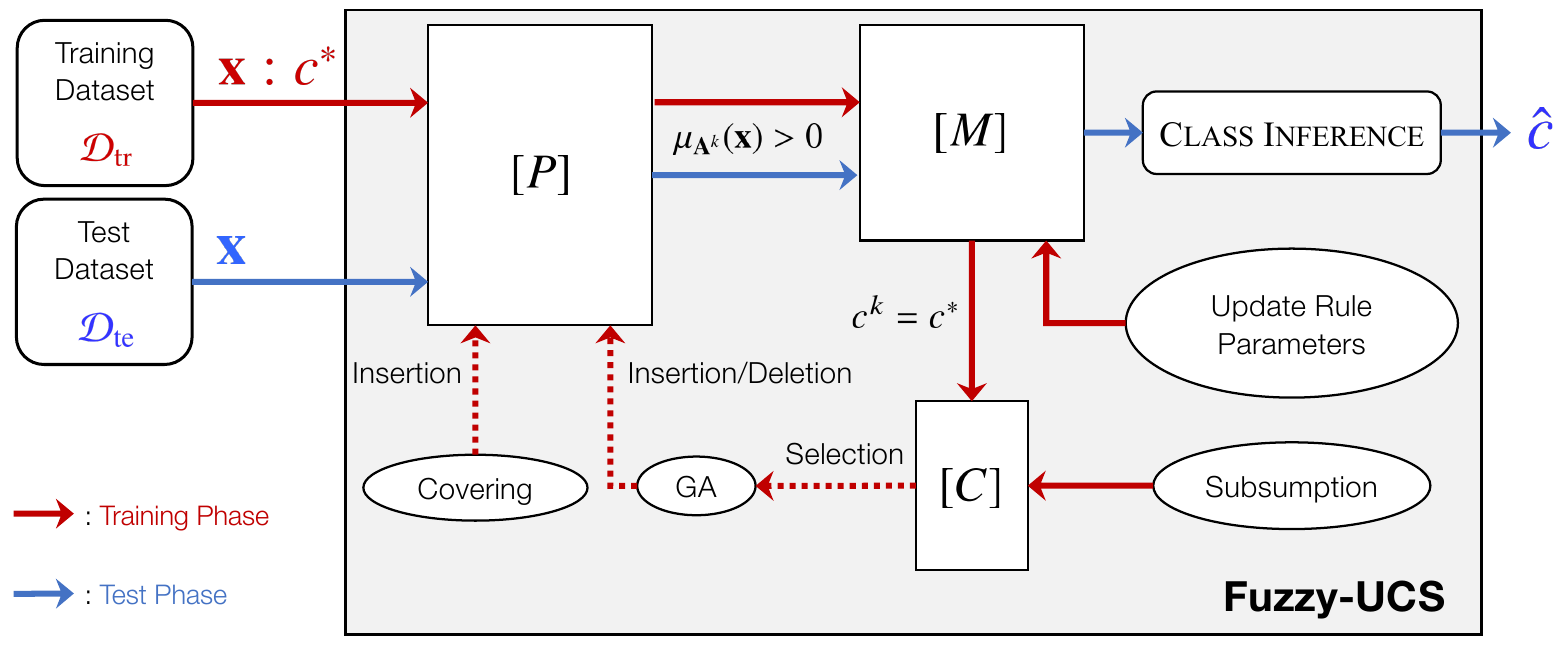}}
\caption{Schematic illustration of Fuzzy-UCS. The run cycle depends on the type of run: training or test. Upon receiving each data point $\mathbf{x}$, operations indicated by solid arrows are always performed, while operations indicated by dashed arrows (i.e., covering and GA) are executed only when specific conditions are met (cf. Section \ref{sss: training phase Fuzzy-UCS}).}
\label{fig: fuzzy-ucs}
\end{figure}

{Fig. \ref{fig: fuzzy-ucs} schematically illustrates Fuzzy-UCS.}

\subsubsection{Training Phase}
\label{sss: training phase Fuzzy-UCS}
\begin{algorithm}[t]
    \caption{\normalsize{Fuzzy-UCS training phase, adapted from \cite{orriols2008fuzzy}} \hspace{\fill} \normalsize{$\triangleright$ Section \ref{sss: training phase Fuzzy-UCS}}}
    \label{alg: fuzzy-ucs training}
    \begin{algorithmic}[1]
 \renewcommand{\algorithmicrequire}{\textbf{Input:}}
 \renewcommand{\algorithmicensure}{\textbf{Output:}}
 \Require the training dataset $\mathcal{D}_{\text{tr}}$;
 \Ensure the ruleset ${{[P]}}$;
    \State Initialize time $t$ as $t\leftarrow 0$;
    \State Initialize ruleset ${{[P]}}$ as ${{[P]}}\leftarrow \emptyset$;
    \While{termination criteria are not met}
        \State Observe a data point $\mathbf{x}\in\mathcal{D}_{\text{tr}}$;
                    \State Create match set ${{[M]}} \coloneqq \{k \in {{[P]}} \mid \mu_{\mathbf{A}^k}(\mathbf{x}) > 0\}$;

            \State Observe correct class $c^*\in\mathcal{C}$ associated with $\mathbf{x}$;
            \State Create correct set ${{[C]}}\coloneqq\{k\in{{[M]}} \mid c^k=c^*\}$;
            \If{$\sum_{k\in{{[C]}}}\mu_{\mathbf{A}^k}(\mathbf{x})<1$}
                \State Execute covering operator to generate a new rule $k_\text{cov}$;
            \EndIf
            
            \State Update ${\rm exp}^k$, $\mathbf{cm}^k$, $\mathbf{v}^k$, and $F^k$ for $\forall k\in{{[M]}}$ as in Eqs. (\ref{eq: update_experience})-(\ref{eq: update_fitness});
            \State Do correct set subsumption if necessary;
            \If{$t - \sum_{k\in{{[C]}}}{\rm num}^k \cdot {\rm ts}^k / {\sum_{k\in{{[C]}}}{\rm num}^k}>\theta_\text{GA}$}
            \State Update ${\rm ts}^k$ for $\forall k\in{{[C]}}$ as ${\rm ts}^k\leftarrow t$;
            \State Run GA on ${{[C]}}$;
            \State Do GA subsumption if necessary;
            \EndIf
            \State Update $t$ as $t \leftarrow t + 1$;
        \EndWhile

    \end{algorithmic}
\end{algorithm}

Algorithm \ref{alg: fuzzy-ucs training} describes the whole procedure of Fuzzy-UCS during the training phase.
At time $t=0$, the \textit{ruleset} ${{[P]}}$ is initialized as an empty set (lines 1-2).
Each training data point in a training dataset $\mathcal{D}_{\text{tr}}$ is used for a pre-specified number of epochs (e.g., 10 epochs) in a random order, which is the termination criteria of the training phase.

All rules are stored in ${{[P]}}$. At time $t$, the system receives a data point $\mathbf{x}$ from $\mathcal{D}_{\text{tr}}$ (line 4). Subsequently, a \textit{match set} ${{[M]}} = \{k \in {{[P]}} \mid \mu_{\mathbf{A}^k}(\mathbf{x}) > 0\}$ is constructed (line 5).
Let $c^*$ be the class of this data point (line 6).
After ${{[M]}}$ is formed, the system forms a \textit{correct set} ${{[C]}}=\{k\in{{[M]}} \mid c^k = c^*\}$ (line 7).
If $\sum_{k\in{{[C]}}}{\mu_{\mathbf{A}^k}}(\mathbf{x})<1$ is satisfied\footnote{{If no rules match a data point (i.e., $[M] = \emptyset$), then $[C] = \emptyset$ and $\sum_{k \in [C]} \mu_{\mathbf{A}^k}(\mathbf{x}) = 0$, satisfying the condition $\sum_{k \in [C]} \mu_{\mathbf{A}^k}(\mathbf{x}) < 1$. This triggers the covering operator to generate a new rule, ensuring the data point is covered.}}, the \textit{covering} operator generates a new rule $k_\text{cov}$ such that $\mu_{\mathbf{A}^{k_\text{cov}}}(\mathbf{x})=1$, $c^{k_\text{cov}}=c^*$, and $v_i^{k_\text{cov}}=1$ if $c_i=c^*$ else $0$ for $\forall i \in \{1,...,n\}$ (lines 8-10).

After ${{[C]}}$ is formed, the parameters of all rules in ${{[M]}}$ are updated (line 11).
First, the experience of each rule $k$ in ${{[M]}}$ is updated according to the current membership degree using\footnote{Unlike the membership degree-based update method in Fuzzy-UCS, in LCSs (such as UCS and ExSTraCS), the experience is updated incrementally using $\text{exp}_{t+1}^k = \text{exp}_{t}^k + 1$, uniformly increasing the experience of all matched rules. 
Our comparative experiments showed that Fuzzy-UCS with membership degree-based updates consistently achieved higher accuracy than the method that incrementally increased the experience by one. For further details, kindly refer to Appendix \ref{sec: sup analysis of experience update methods}.}:
\begin{equation}
\label{eq: update_experience}
    {\rm exp}^k_{t+1} = {\rm exp}^k_t+\mu_{\mathbf{A}^k}(\mathbf{x}).
\end{equation}
Next, the fitness is updated. 
To accomplish this, the system first updates the correct matching vector for $\forall c_i\in\mathcal{C}$ as:
\begin{equation}  \label{eq: update_cm}
{\rm cm}^k_{i_{t+1}}=
    \begin{cases}
       {\rm cm}^k_{i_t}+ \mu_{\mathbf{A}^k}(\mathbf{x})   &   \text{if } c_i=c^*,\\
        {\rm cm}^k_{i_t}& \text{otherwise.}
    \end{cases}
\end{equation}
Then, the system updates the weight for $\forall c_i\in\mathcal{C}$ using:
\begin{equation}
\label{eq: update_weight}
    v^k_{i_{t+1}}=\frac{{\rm cm}^k_{i_{t+1}}}{{\rm exp}^k_{t+1}},
\end{equation}
where $v_i^k$ indicates the soundness of which rule $k$ predicts class $c_i$ for a matched input. 
Note that the value of $\sum_{i=1}^{n}{v_i^k}$ is always $1$.
After that, the system updates the fitness using:
\begin{equation}
\label{eq: update_fitness}
    F^k_{t+1}=v^k_{\text{max}_{t+1}} - \sum_{i|i\neq\text{max}}{v^k_{i_{t+1}}},
\end{equation}
where 
the system subtracts the values of the other weights from the weight with maximum value $v^k_\text{max}$
\cite{ishibuchi2005rule}.
Finally, the highest weight $v_{\max}^k$ in the weight vector $\mathbf{v}^k$ and its corresponding class label $c_{\max}$ are designated as the rule-weight $w^k$ and the rule-consequent class $c^k$ in Eq. (\ref{eq: rule_repr}), respectively.

After the rule update, a steady-state GA is applied to ${{[C]}}$ (lines 13-17).
Michigan-style LFCSs (e.g., Fuzzy-XCS \cite{casillas2007fuzzy} and Fuzzy-UCS) are designed to call the GA after an interval of a given number of inputs controlled by the hyperparameter $\theta_\text{GA}$ \cite{nakata2020learning}. More specifically, the GA is executed when $t-\frac{\sum_{k\in{{[C]}}}{\rm num}^k \cdot {\rm ts}^k}{{\sum_{k\in{{[C]}}}{\rm num}^k}}>\theta_\text{GA}$ is satisfied.
In this case, the two parent rules from ${{[C]}}$ excluding any rules with negative fitness are selected through tournament selection, with the tournament size determined by the hyperparameter $\tau$. 
{This mechanism enables the GA to guide the system in systematically eliminating inaccurate rules, thereby enhancing the performance of the system.}
The two parent rules are replicated as two child rules, and then crossover and mutation are applied to the two child rules with probabilities $\chi$ and $p_\text{mut}$, respectively.
The two child rules are inserted into ${{[P]}}$ and two rules are deleted if the number of rules in ${{[P]}}$, $\sum_{k\in{{[P]}}}{\rm num}^k$, exceeds the ruleset size $N$.

The \textit{subsumption} operator
\cite{wilson1998generalization} 
can be employed to prevent inserting over-specific rules into the ruleset
\cite{liu2020absumption}.
Subsumption is triggered after ${{[C]}}$ is formed or after GA is executed, and is referred to as \textit{Correct Set Subsumption} (line 12) and \textit{GA Subsumption} (line 16), respectively. 
Specifically, for the two rules $k_\text{sub}$ and $k_\text{tos}$, if 
(i) $k_\text{sub}$ is more general than $k_\text{tos}$
(i.e., $k_\text{sub}$ has at least the same linguistic terms per variable as $k_\text{tos}$);
(ii) $k_\text{sub}$ is accurate (i.e., $F^{k_\text{sub}}$ is greater than the hyperparameter $F_0$); and 
(iii) $k_\text{sub}$ is sufficiently experienced (i.e., ${\rm exp}^{k_\text{sub}}$ is greater than the hyperparameter $\theta_\text{sub}$), 
then $k_\text{tos}$ is removed from ${{[P]}}$, and the numerosity of ${k_\text{sub}}$ is updated using ${\rm num}_{t+1}^{k_\text{sub}} = {\rm num}_t^{k_\text{sub}} + {\rm num}_t^{k_\text{tos}}$ to record the number of subsumed rules.

\subsubsection{Test Phase}
\label{sss: test phase Fuzzy-UCS}

\begin{algorithm}[t]
    \caption{\normalsize{Fuzzy-UCS test phase, adapted from \cite{orriols2008fuzzy}} \hspace{\fill} \normalsize{$\triangleright$ Section \ref{sss: test phase Fuzzy-UCS}}}
    \label{alg: fuzzy-ucs test}
    \begin{algorithmic}[1]
    \renewcommand{\algorithmicrequire}{\textbf{Input:}}
 \renewcommand{\algorithmicensure}{\textbf{Output:}}
 \Require a data point $\mathbf{x}$ from the test dataset $\mathcal{D}_{\text{te}}$, the ruleset ${{[P]}}$;
 \Ensure the inferred class label $\hat{c}$ of $\mathbf{x}$;
    \State Create match set ${{[M]}} \coloneqq \{k \in {{[P]}} \mid \mu_{\mathbf{A}^k}(\mathbf{x}) > 0\}$;
            \State Filter ${{[M]}}$ as ${{[M]}}\leftarrow\{k\in{{[M]}} \mid {\rm exp}^k>\theta_\text{exploit}\}$;
            \If{${{[M]}}\neq\emptyset$}
            \State $\hat{c} \leftarrow$ \textsc{ClassInference}(${{[M]}}, \mathbf{x}$);
            \Else
            \State $\hat{c} \leftarrow$ a randomly chosen class;
            \EndIf

    \end{algorithmic}
\end{algorithm}
Algorithm \ref{alg: fuzzy-ucs test} describes the whole procedure of Fuzzy-UCS during the test phase.

Given a data point $\mathbf{x}$ from a test dataset $\mathcal{D}_{\text{te}}$, the system's output $\hat{c}$ is determined through class inference using multiple rules in the match set ${{[M]}}$. 
This subsection outlines two inference schemes (cf. Section \ref{ss: related work class inference} and line 4).
It is important to note that only rules with sufficient experience (i.e., ${\rm exp}^k$ is greater than the hyperparameter $\theta_\text{exploit}$) are considered for inference (line 2).

\textbf{Voting-based inference.} The system's output $\hat{c}$ is determined using:
\begin{equation}
\label{eq: vote}
\hat{c} = \arg \max_{c\in\mathcal{C}}\sum_{k\in[{M}_c]}{\mu_{\mathbf{A}^k}(\mathbf{x})\cdot F^k\cdot{\rm num}^k},
\end{equation}
where $[{M}_c]$ represents the subset of rules in ${{[M]}}$ with a consequent class of $c$. Here, $\mu_{\mathbf{A}^k}(\mathbf{x})\cdot F^k\cdot{\rm num}^k$ represents the votes a rule $k$ has. The class with the highest number of votes is then the output \(\hat{c}\).

\textbf{Single-winner-based inference.} 
The system's output $\hat{c}$ is determined using:
\begin{equation}
\hat{c} = c^{k^*}, \quad \text{where} \quad k^* = \arg \max_{k \in {{[M]}}} \left( \mu_{\mathbf{A}^k}(\mathbf{x}) \cdot F^k \right).
\end{equation}
Here, \(k^*\) is the single-winner rule that maximizes \(\mu_{\mathbf{A}^k}(\mathbf{x}) \cdot F^k\) among all the rules in \({{[M]}}\). The consequent class \(c^{k^*}\) of this rule \(k^*\) is then the output \(\hat{c}\).

It is worth noting that, for both inference schemes, the system's output $\hat{c}$ is determined randomly if there are no rules that allow participation in inference (i.e., ${{[M]}}=\emptyset$, line 6). In case of a tie, such as when the total votes for $c_1$ and $c_2$ are equal in the voting-based inference, $\hat{c}$ is randomly selected from the tied classes (i.e., $c_1$ and $c_2$).

\section{Dempster-Shafer Theory of Evidence}
\label{sec: ds theory}

The \textit{Dempster-Shafer Theory of Evidence} (DS theory) was introduced by Shafer \shortcite{shafer1976mathematical} as a reformation of Dempster's earlier work \shortcite{dempster1967upper}. It offers a robust framework for reasoning with uncertainty. This section outlines the fundamental concepts of the DS theory that are important for understanding the proposed class inference scheme.

\subsection{Frame of Discernment}

The \textit{frame of discernment}, denoted as $\mathbf{\Theta}=\{\theta_1,\theta_2,...,\theta_n\}$, represents a finite set of distinct, mutually exclusive hypotheses $\theta_i$. Each subset of $\mathbf{\Theta}$ can be viewed as a composite hypothesis, leading to a total of $|2^\mathbf{\Theta}|$ possible hypotheses, where $2^\mathbf{\Theta}$ denotes the power set of $\mathbf{\Theta}$. For instance, if $\mathbf{\Theta}=\{\theta_1, \theta_2, \theta_3\}$, then $2^\mathbf{\Theta}$ encompasses the following eight subsets: $\{\emptyset, \{\theta_1\}, \{\theta_2\}, \{\theta_3\}, \{\theta_1, \theta_2\}, \{\theta_1, \theta_3\}, \{\theta_2, \theta_3\}, \mathbf{\Theta}\}$.

In the context of an $n$-class classification problem with a class label set $\mathcal{C}=\{c_i\}_{i=1}^n$, a subset such as $\{\theta_1\}$ represents the hypothesis ``the class of a given data point is $c_1$'', implying a specific class determination. A subset like $\{\theta_1, \theta_2\}$ suggests ``the class of the given data point is either $c_1$ or $c_2$'', denoting partial ignorance or a meta-class. $\mathbf{\Theta}$ as a whole implies complete uncertainty about the class (i.e., any class in $\mathcal{C}$), representing a complete ignorance, uncertainty, or an ``I don't know'' state \cite{sensoy2018evidential}. Conversely, the empty set $\emptyset$ denotes a hypothesis that is considered impossible or improbable (i.e., no class in $\mathcal{C}$).

\subsection{Belief Mass Function}
Within the DS theory framework, the concept of a \textit{belief mass function} is important. This function, denoted as $m:2^\mathbf{\Theta} \rightarrow [0,1]$, assigns a degree of belief, referred to as \textit{belief mass}, to each hypothesis within the frame of discernment. The belief mass function adheres to specific criteria outlined in the following equations:
\begin{align}
\label{eq: belief1}
    & m(A) \geq 0, \;\forall A\in2^\mathbf{\Theta},\\
\label{eq: belief2}
    & m(\emptyset) = 0,\\
\label{eq: belief3}
    & \sum_{A\in2^\mathbf{\Theta}}{m(A)}=1.
\end{align}
The belief mass $m(A)$ is interpreted as the specific degree of belief committed solely to the hypothesis $A$ and not to any of its subsets. In this context, subsets $A$ of $\mathbf{\Theta}$ for which $m(A)>0$ are termed \textit{focal elements} of $m$. The collection of all such focal elements constitutes the \textit{core} of $m$ \cite{liu2013new}. The condition $m(\emptyset)=0$ is a fundamental requirement in the DS theory. Shafer \cite{shafer1976mathematical} introduced it as the normality condition. It ensures that no belief is assigned to the empty set (i.e., to an impossible hypothesis).

\subsection{Combination of Belief Masses}

The DS theory employs a specific operator known as \textit{Dempster's rule of combination} \cite{shafer1976mathematical} for fusing information from distinct, independent sources of knowledge. This process involves the following steps.

Initially, the \textit{conjunctive sum} of two belief mass functions $m_1$ and $m_2$, corresponding to every hypothesis $\forall A\in 2^\mathbf{\Theta}$, is computed. This is achieved through the following formula called \textit{conjunctive rule of combination} \cite{smets1990combination}:
\begin{equation}
\label{eq: conjuctive sum operator}
    (m_1 \cap m_2)(A)\triangleq\sum_{B\cap C = A}{m_1(B)\cdot m_2(C)}.
\end{equation}

Subsequently, using Dempster's rule of combination, the functions $m_1$ and $m_2$ are combined as follows:
\begin{equation}
\label{eq: dempster rule}
    (m_1 \oplus m_2)(A)\triangleq\frac{1}{1-\mathcal{K}_{12}}(m_1 \cap m_2)(A),
\end{equation}
where $\mathcal{K}_{12}\triangleq(m_1 \cap m_2)(\emptyset)$ quantifies the \textit{conflict} between $m_1$ and $m_2$. The term $\mathcal{K}_{12}$ is referred to as the \textit{mass of conflict} \cite{kessentini2015dempster}. A higher value of $\mathcal{K}_{12}$, signifies greater inconsistency between the two sets of information. The normalization factor $\frac{1}{1-\mathcal{K}_{12}}$ reallocates the belief mass that is lost due to contradictions, distributing it among the non-contradictory elements (i.e., focal elements). This adjustment ensures that the conditions in Eqs. (\ref{eq: belief2}) and (\ref{eq: belief3}) remain valid. 

It is worth noting that both the conjunctive rule of combination, $\cap$, and Dempster's rule of combination, $\oplus$, exhibit properties of commutativity and associativity \cite{lefevre2002belief,smets1990combination}. This allows for flexibility in the order and grouping of information combinations.

\subsection{Pignistic Transform}
\label{ss: pignistic transform}
{The \textit{transferable belief model} (TBM) \cite{smets1994transferable} is a framework for reasoning under uncertainty that distinguishes between belief representation and decision-making. Within this framework, the \textit{pignistic transform} \cite{smets1994transferable} plays a crucial role in bridging these two aspects. The term ``pignistic'' comes from the Latin word ``pignus,'' meaning a bet, emphasizing its role in decision-making processes \cite{smets2000data}.}

The pignistic transform is a process that converts the belief mass function $m$ into a form of probability, called \textit{pignistic probability}.
{This transformation is necessary because while belief functions are excellent for representing uncertainty and partial knowledge, they are not directly suitable for decision-making. The resulting pignistic probability provides a basis for making decisions using the expected utility theory \cite{smets2005decision}.}
The pignistic probability for any hypothesis within the frame of discernment $\forall \theta\in\mathbf{\Theta}$ is calculated as follows:
\begin{equation}
\label{eq: betp}
    {\rm BetP}(\theta)\triangleq\sum_{\theta\in A \subseteq \mathbf{\Theta}}{\frac{m(A)}{|A|}},
\end{equation}
where $|A|$ represents the cardinality (i.e., number of elements) of the subset $A$.
The pignistic probability is computed by distributing the belief mass $m(A)$ of each subset $A$ of $\mathbf{\Theta}$ equally among the elements in $A$. This distribution process is based on the concept of fairness, especially in the context of ambiguity or partial ignorance \cite{kessentini2015dempster}.

Pignistic probabilities are particularly valuable in decision-making under uncertainty or when dealing with incomplete information. By transforming belief masses into a more interpretable and actionable form, the pignistic transform helps to derive clear and rational decisions from complex and uncertain data \cite{liu2013new}.

\section{The Proposed Class Inference Scheme}
\label{sec: proposed method}
\subsection{Overview}

This section proposes a novel class inference scheme for the test phase of LFCSs. The proposed scheme effectively incorporates the DS theory into the LFCS framework. It achieves this by quantifying and combining multiple pieces of information, namely the membership degrees of rules and their associated weight vectors. These elements are derived from various independent knowledge sources, namely fuzzy rules acquired by an LFCS, and are treated as belief masses within the DS theory framework. This approach aims to enhance the LFCS’s ability to perform more accurate and reliable class inference, especially in situations characterized by complexity and uncertainty.

\subsection{Algorithm}
\label{ss: proposed algorithm}

{Fig. \ref{fig: ds-inference} schematically illustrates the algorithm.}
Algorithm \ref{alg: proposed inference} describes the whole procedure of the proposed class inference scheme. 
The algorithm is designed for an $n$-class single-label classification problem with a class label set $\mathcal{C}=\{c_i\}_{i=1}^n$. In the algorithm, $\mathbf{\Theta}=\{\theta_i\}_{i=1}^n$ is the frame of discernment. This frame consists of a finite set of hypotheses $\theta_i$ each of which corresponds to a specific class label $c_i$. The proposed inference scheme follows these steps:
\begin{figure}[b]
\centerline{\includegraphics[width=\linewidth]{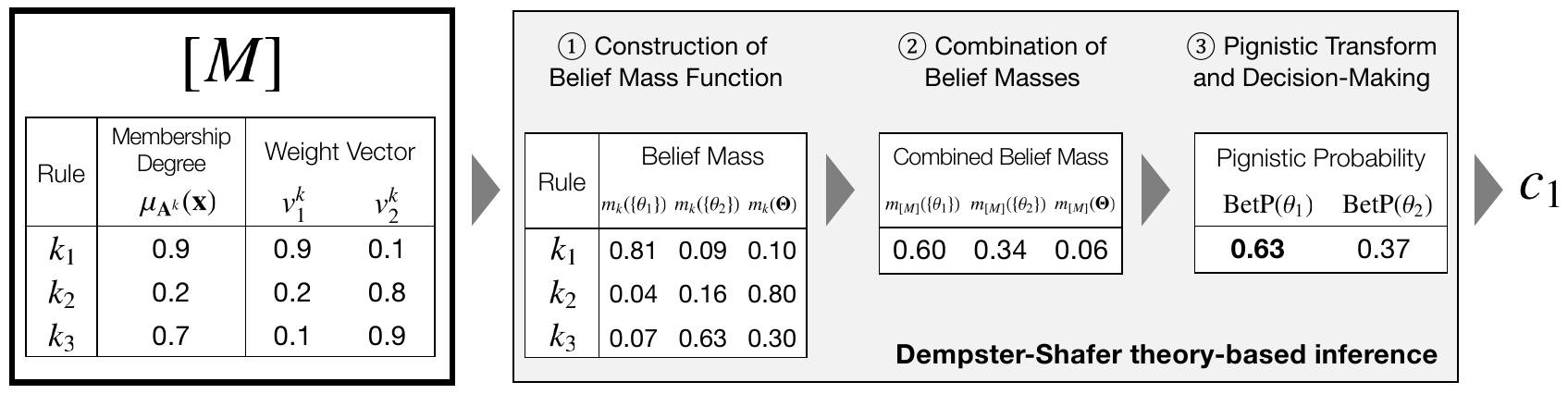}}
\caption{An example of how the proposed class inference scheme works for an input data point $\mathbf{x}$ on a binary classification problem with a class label set $\mathcal{C}=\{c_1,c_2\}$.}
\label{fig: ds-inference}
\end{figure}
\begin{algorithm}[b]
    \caption{\normalsize{Dempster-Shafer theory-based inference} \hspace{\fill} \normalsize{$\triangleright$ Section \ref{ss: proposed algorithm}}}
    \label{alg: proposed inference}
    \begin{algorithmic}[1]
    \Function{ClassInference}{${{[M]}}$, $\mathbf{x}$} 
    \hspace{\fill} {\normalsize $\triangleright$ cf. Alg. \ref{alg: fuzzy-ucs test}, line 4}

    \State Construct belief mass function $m_k(A)$ for $\forall k \in{{[M]}}$ as in Eq. (\ref{eq: proposed mass});
    \State Calculate the mass of conflict $\mathcal{K}_{{[M]}}$ as in Eq. (\ref{eq: mass of conflict});
    \State Calculate the combined belief mass $m_{{[M]}}(A)$:
    \If{$\mathcal{K}_{{[M]}}\neq 1$}
    \State Using Dempster's rule of combination as in Eq. (\ref{eq: proposed dempster});
    \Else
    \State Using Yager's rule as in Eq. (\ref{eq: yager});
    \EndIf
    \State Construct pignistic probability ${\rm BetP}(\theta)$ for $\forall \theta\in\mathbf{\Theta}$ as in Eq. (\ref{eq: betp});
    \State \textbf{return} the best hypothesis in $\mathbf{\Theta}$ as in Eq. (\ref{eq: proposed betp});
    \EndFunction

    \end{algorithmic}
\end{algorithm}

\begin{enumerate}
    \item \textbf{Construction of Belief Mass Function.} Formulate a belief mass function for each fuzzy rule in the match set ${{[M]}}$. This function is based on the rule’s membership degree $\mu_{\mathbf{A}^k}(\mathbf{x})$ and a weight vector $\mathbf{v}^k$ (detailed in Section \ref{sss: construct mass}).

    \item \textbf{Combination of Belief Masses.} Integrate the belief masses of all fuzzy rules in ${{[M]}}$ to form a combined belief mass, denoted as $m_{{[M]}}$ (detailed in Section \ref{sss: combine mass}).

    \item \textbf{Pignistic Transform and Decision-Making.} Apply the pignistic transform to $m_{{[M]}}$ and determine the class with the highest pignistic probability. This class is then selected as the inference result $\hat{c}$ (detailed in Section \ref{sss: betP vector}).
\end{enumerate}

\subsubsection{Construction of Belief Mass Function}
\label{sss: construct mass}

The belief mass function $m_k(A)$ for a given rule $k\in{{[M]}}$, corresponding to every hypothesis $\forall A\in 2^\mathbf{\Theta}$, is defined as follows (line 2):
\begin{equation}  \label{eq: proposed mass}
m_k(A)=
    \begin{cases}
        \mu_{\mathbf{A}^k}(\mathbf{x})\cdot v_i^k   & \text{if } A\in\{\{\theta_i\}\}_{i=1}^n,\\
        \displaystyle 1-\sum_{i=1}^n{m_k(\{\theta_i\})} & \text{if } A=\mathbf{\Theta},\\
        0& \text{otherwise.}
    \end{cases}
\end{equation}

This definition assigns the belief mass to each hypothesis as follows:
\begin{itemize}
    \item For any specific class hypothesis $A\in\{\{\theta_i\}\}_{i=1}^n$ (a singleton set), the belief mass is the product of the membership degree $\mu_{\mathbf{A}^k}(\mathbf{x})$ and the class-specific weight $v_i^k$.
    \item For the hypothesis $\mathbf{\Theta}$, which represents complete ignorance or uncertainty about the class, the belief mass is set to the remainder of the belief mass after accounting for all specific classes. This is computed as $1$ minus the sum of belief masses for all specific class hypotheses.
    \item For any other hypotheses (combinations of classes or meta-classes), the belief mass is set to $0$.
\end{itemize}

This approach ensures that the sum of belief masses across all hypotheses equals $1$, as required by the DS theory (cf. Eq. (\ref{eq: belief3})). By allocating the residual belief to the hypothesis of complete ignorance $\mathbf{\Theta}$, the system maintains a balance between certainty (specific class hypotheses) and uncertainty (complete ignorance), reflecting a comprehensive view of the class distribution for the given input. 

This approach is similar to methods used in machine learning techniques that apply the DS theory to single-label classification problems. Examples include \cite{sensoy2018evidential} and \cite{belmahdi2023application}. In these techniques, the core's maximum size is typically set to $n+1$. This size accounts for $n$ specific class hypotheses and one hypothesis representing complete ignorance.

\subsubsection{Combination of Belief Masses}
\label{sss: combine mass}

This process combines the belief mass functions of all fuzzy rules within ${{[M]}}$ into a single belief mass function $m_{{[M]}}$. 
The goal of this step is to obtain the belief mass $m_{{[M]}}(A)$ for every hypothesis $\forall A\in2^\mathbf{\Theta}$.

Initially, the mass of conflict $\mathcal{K}_{{[M]}}$ among all rules in ${{[M]}}$ is calculated using the conjunctive sum operator (line 3):
\begin{equation}
\label{eq: mass of conflict}
    \mathcal{K}_{{[M]}}=\left(\bigcap_{k\in{{[M]}}}m_k\right)\left(\emptyset\right).
\end{equation}

If $\mathcal{K}_{{[M]}} < 1$, which means that the belief masses of the rules in ${{[M]}}$ do not entirely contradict one another, the combined belief mass $m_{{[M]}}(A)$ is calculated using Dempster's rule of combination (line 6):
\begin{align}
\label{eq: proposed dempster}
    m_{{[M]}}(A) =\left(\bigoplus_{k\in{{[M]}}}{m_k}\right)\left(A\right) 
    =\frac{1}{1-\mathcal{K}_{[M]}}\left(\bigcap_{k\in{{[M]}}}m_k\right)(A).
\end{align}

Conversely, if $\mathcal{K}_{{[M]}} = 1$, which means the complete contradiction in the information from the rules in ${{[M]}}$, Dempster's rule of combination becomes inapplicable due to the issue of division by zero. In this case, \textit{Yager's rule} \cite{yager1987dempster} is utilized to address this contradiction (line 8):
\begin{equation}
\label{eq: yager}
  m_{{[M]}}(A)=
   \begin{cases}
        1& \text{if } A=\mathbf{\Theta},\\
        0& \text{otherwise.}
    \end{cases}
\end{equation}
Eq. (\ref{eq: yager}) handles a complete contradiction between independent knowledge sources as a state of ``complete lack of knowledge that resolves the contradiction.''\cite{yager1987dempster} It resolves this contradiction by reassigning the mass of conflict $\mathcal{K}_{{[M]}}=1$ to the belief mass of complete ignorance $\mathbf{\Theta}$. This approach manages contradictions by acknowledging them as a lack of sufficient information to resolve the conflict.

\subsubsection{Pignistic Transform and Decision-Making}
\label{sss: betP vector}

To make the final decision, the proposed class inference scheme utilizes a pignistic transform on the combined belief mass function $m_{{[M]}}$. This transformation yields the pignistic probability ${\rm BetP}(\theta)$ for each hypothesis $\theta$ within the frame of discernment $\mathbf{\Theta}$, as formulated in Eq. (\ref{eq: betp}) (line 10). 

The decision-making process involves selecting the class $c_i$ associated with the hypothesis $\{\theta_i\}$ that has the highest pignistic probability. This is represented as follows (line 11):
\begin{equation}
\label{eq: proposed betp}
    \hat{c} = \arg \max_{\theta\in\mathbf{\Theta}}{\rm BetP}(\theta),
\end{equation}
where $\hat{c}$ denotes the inferred class label, which is determined based on the maximum pignistic probability among all considered hypotheses (i.e., all classes). This approach ensures that class inference is based on a comprehensive assessment of the evidence gathered from the fuzzy rules in ${{[M]}}$.

When there are ties with multiple hypotheses having the same highest pignistic probability, the system randomly chooses the output $\hat{c}$ from the tied classes, similar to the conventional class inference schemes (cf. Section \ref{sss: test phase Fuzzy-UCS}). Such situations typically arise when $m_{{[M]}}(\mathbf{\Theta})=1$ according to Yager's rule. In these cases, the pignistic probability for each hypothesis $\theta_i$ is equally distributed according to Eq. (\ref{eq: betp}). This approach ensures that the system can still generate a randomly selected output in cases where information is either completely uncertain or contradictory.

\section{Experiments}
\label{sec: experiment}

This section utilizes 30 real-world datasets in Table \ref{tb: dataset} from the \textit{UCI Machine Learning Repository} \cite{dua2019uci} and \textit{Kaggle Dataset} collection. These datasets are selected due to their inherent challenges in data classification, such as the presence of missing values and class imbalance issues \cite{Thumpati2023Towards}.

Our objective is to evaluate the impact of different class inference schemes on the classification accuracy of LFCSs. To accomplish this, we compare the performance of three Fuzzy-UCS$_*$ variants. Here, $*$ is the abbreviation of the inference scheme used in Fuzzy-UCS. The abbreviations are defined as follows: (i) VOTE, for the voting-based inference scheme; (ii) SWIN, for the single-winner-based inference scheme; and (iii) DS, for the proposed DS theory-based inference scheme. It is important to note that all three Fuzzy-UCS$_*$ variants utilize the same ruleset, ${{[P]}}$, produced during the training phase. This ensures a fair comparison across the different inference schemes.
Additionally, to provide a comprehensive evaluation of the proposed scheme, we include a well-known LCS, the UCS nonfuzzy-classifier system \cite{bernado2003accuracy}, in our comparison.

\begin{table}[t]
\begin{center}
\caption{Properties of the 30 Real-World Datasets. The Columns Describe: the Identifier (ID.), the Name (Name), the Number of Instances ($\#$\textsc{Inst.}), the Total Number of Features ($\#$\textsc{Fea.}), the Number of Classes ($\#$\textsc{Cl.}), the Percentage of Missing Attributes ($\%$\textsc{Mis.}), {the Percentage of Instances of the Majority Class ($\%$\textsc{Maj.}), the Percentage of Instances of the Minority Class ($\%$\textsc{Min.})}, and the Source (Ref.).}
\label{tb: dataset}
\footnotesize
\scalebox{0.9}{
\begin{tabular}{c l c c c c c c c}
\bhline{1pt}
ID. & Name  & $\#$\textsc{Inst.}  & $\#$\textsc{Fea.} &$\#$\textsc{Cl.} & $\%$\textsc{Mis.} & {$\%$\textsc{Maj.}} & {$\%$\textsc{Min.}} & Ref.
\\
\bhline{1pt}
\texttt{bnk} &Banknote authentication & 1372 & 4 & 2 & 0 & {76.24} & {23.76} &  \cite{dua2019uci}\\
\texttt{can} &Cancer & 569 & 30 & 2 & 0 & {62.74} & {37.26} &\tablefootnote{{\url{https://www.kaggle.com/datasets/erdemtaha/cancer-data} (\today)}} \\
\texttt{car} &Car acceptance prediction & 50 & 4 & 2 & 0 & {68.00} & {32.00} &\tablefootnote{\url{https://www.kaggle.com/datasets/deepatharwani/car-acceptance-datasets} (\today)} \\
\texttt{col} &Column 3C weka & 310 & 6 & 3 & 0 & {46.13} & {7.74}& \cite{dua2019uci}  \\
\texttt{dbt} &Diabetes & 768 & 8 & 2 & 0 & {65.10} & {34.90}& \cite{dua2019uci} \\
\texttt{ecl} &Ecoli & 336 & 7 & 8 & 0 & {42.86} & {1.49}& \cite{dua2019uci}\\
\texttt{frt} &Fruit & 898 & 34 & 7 & 0 & {22.72} & {7.24} &\cite{koklu2021classification}\\
\texttt{gls} &Glass & 214 & 9 & 6 & 0 & {35.51} & {2.34}& \cite{dua2019uci} \\
\texttt{hrt} &Heart disease & 303 & 13 & 2 & 0 & {54.46} & {45.54}& \cite{dua2019uci}\\
\texttt{hpt} &Hepatitis & 155 & 19 & 2 & 5.67 & {79.35} & {20.65}& \cite{dua2019uci}\\
\texttt{hcl} &Horse colic & 368 & 22 & 2 & 23.80 & {62.77} & {37.23}& \cite{dua2019uci}\\
\texttt{irs} &Iris & 150 & 4 & 3 & 0 & {33.33} & {33.33}& \cite{dua2019uci} \\
\texttt{lnd} &Land mines & 338 & 3 & 5 & 0 & {21.01} & {19.23}&\cite{dua2019uci}\\
\texttt{mam} &Mammographic masses & 961 & 5 & 2 & 3.37 & {53.69} & {46.31}& \cite{dua2019uci}\\
\texttt{pdy} &Paddy leaf & 6000 & 3 & 4 & 0 & {25.00} & {25.00}& \tablefootnote{{\url{https://www.kaggle.com/datasets/torikul140129/paddy-leaf-images-aman} (\today)}} \\
\texttt{pis} &Pistachio & 2148 & 16 & 2 & 0 & {57.36} & {42.64}& \cite{singh2022classification}\\
\texttt{pha} &Pharyngitis in children & 676 & 18 & 2 & 1.86 & {54.88} & {45.12}&  \tablefootnote{{\url{https://www.kaggle.com/datasets/yoshifumimiya/pharyngitis/data} (\today)}}\\
\texttt{pre} &Predicting response based on engagement KPIs & 5000 & 3 & 3 & 0 & {33.34} & {33.32}& \tablefootnote{{\url{https://www.kaggle.com/datasets/micheldc55/predicting-response-based-on-engagement-kpis} (\today)}}\\
\texttt{pmp} &Pumpkin & 2499 & 12 & 2 & 0 & {52.00} & {48.00}& \cite{koklu2021use} \\
\texttt{rsn} &Raisin & 900 & 7 & 2 & 0 & {50.00} & {50.00} & \cite{ccinar2020classification}\\
\texttt{seg} &Segment & 2310 & 19 & 7 & 0 & {14.29} & {14.29}& \cite{dua2019uci} \\
\texttt{sir} &Sirtuin6 small molecules & 100 & 6 & 2 & 0 & {50.00} & {50.00} & \cite{dua2019uci}\\
\texttt{smk} &Smoker condition & 1023 & 7 & 2 & 0.20 & {60.51} & {30.49}& \tablefootnote{{\url{https://www.kaggle.com/datasets/devzohaib/smoker-condition} (\today)}}\\
\texttt{tae} &Teaching assistant evaluation & 151 & 5 & 3 & 0 & {34.44} & {32.45}& \cite{dua2019uci}\\
\texttt{tip} &Travel insurance prediction & 1987 & 8 & 2 & 0 & {64.27} & {35.73} & \tablefootnote{{\url{https://www.kaggle.com/datasets/tejashvi14/travel-insurance-prediction-data} (\today)}}\\
\texttt{tit} &Titanic & 891 & 6 & 2 & 3.31 & {61.62} & {38.38}&\tablefootnote{{\url{https://www.kaggle.com/datasets/yasserh/titanic-dataset} (\today)}}\\
\texttt{wne} &Wine & 178 & 13 & 3 & 0 & {39.89} & {26.97}& \cite{dua2019uci}\\
\texttt{wbc} &Wisconsin breast cancer & 699 & 9 & 2 & 0.25 & {65.52} & {34.48}&\cite{dua2019uci}  \\
\texttt{wpb} &Wisconsin prognostic breast cancer & 198 & 33 & 2 & 0.06 & {76.26} & {23.74}&\cite{dua2019uci} \\
\texttt{yst} &Yeast & 1484 & 8 & 10 & 0 & {31.20} & {0.34}& \cite{dua2019uci} \\
\bhline{1pt}
\end{tabular}
}
\end{center}
\end{table}

\subsection{Experimental Setup}
\label{ss: experimental setup}

For all of the considered problems, the hyperparameters for UCS and Fuzzy-UCS$_*$ are set to the values based on \cite{orriols2008fuzzy,tzima2013strength,urbanowicz2015exstracs}, as shown in Table \ref{tb: hyperparameters}. 
For more information on the hyperparameters, kindly refer to \cite{bernado2003accuracy} and \cite{orriols2008fuzzy}.
In UCS, the \textit{unordered bound hyperrectangular representation} \cite{stone2003real} is utilized for the rule antecedents. The uniform crossover is used for UCS and Fuzzy-UCS$_*$ in the GA. We use our own implementation of UCS and Fuzzy-UCS$_*$, both codified in Julia \cite{bezanson2017julia}. 

\begin{table}[t]
\begin{center}
\caption{Properties of the Hyperparameter Values.}
\label{tb: hyperparameters}
\footnotesize
\begin{tabular}{l c |  c c}
\bhline{1pt}
Description & Hyperparameter & UCS & Fuzzy-UCS$_*$ \\
\bhline{1pt}
Maximum {micro-}ruleset size{\tablefootnote{{In the context of L(F)CSs, the ruleset size has two distinct definitions \cite{urbanowicz2015exstracs}. The first definition is the \textit{micro}-ruleset size, which represents the sum of all rule numerosities in $[P]$, calculated as $\sum_{k\in[P]}{\rm num}^k$. The second definition is the \textit{macro}-ruleset size, which represents the number of unique rules (called macro-rules) currently in $[P]$. }\label{r1-1}}} & $N$ & 2000 & 2000\\
Accuracy threshold & ${\rm acc}_0$ & 0.99 & N/A\\
Fitness threshold & $F_0$ & N/A & 0.99\\
Learning rate & $\beta$ & 0.2 & N/A\\
Fitness exponent & $\nu$ &1 & 1\\
Crossover probability & $\chi$ & 0.8 & 0.8\\
Mutation probability & $p_\text{mut}$ & 0.04 & 0.04\\
Fraction of mean fitness for rule deletion & $\delta$ &0.1 & 0.1\\
Maximum change amount for mutation & $m_0$ & 0.1 & N/A\\
Maximum range for covering & $r_0$ & 1.0 & N/A\\
Time threshold for GA application in ${{[C]}}$ & $\theta_\text{GA}$ & 50 & 50\\
Experience threshold for rule deletion & $\theta_\text{del}$ & 50 & 50\\
Experience threshold for subsumption & $\theta_\text{sub}$ & 50 & 50\\
Experience threshold for class inference & $\theta_\text{exploit}$ & N/A & 10\\
Tournament size & $\tau$ & 0.4 & 0.4\\
Probability of \textit{Don't Care} symbol in covering & $P_\#$ & 0.33 & 0.33\\
Whether correct set subsumption is performed & $\mathit{doCorrectSetSubsumption}$ & $\mathit{yes}$ & $\mathit{yes}$\\
Whether GA subsumption is performed & $\mathit{doGASubsumption}$ & $\mathit{yes}$ & $\mathit{yes}$\\

\bhline{1pt}
\end{tabular}
\end{center}
\end{table}
To evaluate the impact of learning duration, we conduct experiments with two different training durations: a shorter duration of 10 epochs to simulate a situation with limited computation resources, and an extended duration of 50 epochs to ensure adequate learning time. {Here, an epoch refers to the process where the system trains on the entire training dataset once. }

We conduct 30 independent runs for each experiment, each initialized with a different random seed. The datasets are divided using shuffle-split cross-validation, allocating 90\% of the instances for training and the remaining 10\% for testing, as in \cite{preen2021autoencoding}. All attribute values in the datasets are normalized to the range $[0,1]$. 
The performance metric used to evaluate the systems is the average macro F1 scores, obtained from 30 runs of both the training and test data.

To analyze statistical significance, we first apply the Friedman test to the results of 30 runs on all four systems to determine a significance probability. If the Friedman test indicates a probability small enough, we then proceed with the Holm test, using the Wilcoxon signed-rank test as a post-hoc method. A difference is considered significant if the probability is less than 0.05 in both the Friedman and Holm post-hoc tests.

\subsection{Results}
\label{ss: results}
\begin{table*}[t]
\begin{center}
\caption{Results When the Number of Training Iterations is 10 Epochs, Displaying Average Macro F1 Scores Across 30 Runs.}
\label{tb: f1 score 10 epoch}
\scalebox{0.7}{
\begin{tabular}{c|c c c c|cccc}
\bhline{1pt}
\multicolumn{1}{c|}{\multirow{2}{*}{10 Epochs}} &\multicolumn{4}{c|}{\textsc{Training Macro F1 (\%)}} & \multicolumn{4}{c}{\textsc{Test Macro F1 (\%)}}\\
& UCS & Fuzzy-UCS$_\text{VOTE}$ & Fuzzy-UCS$_\text{SWIN}$ & Fuzzy-UCS$_\text{DS}$ & UCS & Fuzzy-UCS$_\text{VOTE}$ & Fuzzy-UCS$_\text{SWIN}$ & Fuzzy-UCS$_\text{DS}$\\
\bhline{1pt}
\texttt{bnk} & \cellcolor{p}70.02 $-$ & 89.44 $-$ & \cellcolor{g}91.83 $+$ & 89.88 & \cellcolor{p}68.96 $-$ & 89.00 $-$ & \cellcolor{g}91.27 $+$ & 89.54 \\
\texttt{can} & \cellcolor{g}93.53 $\sim$ & \cellcolor{p}93.26 $-$ & 93.37 $\sim$ & 93.40 & \cellcolor{p}89.13 $-$ & 92.68 $\sim$ & \cellcolor{g}92.82 $\sim$ & 92.68 \\
\texttt{car} & \cellcolor{p}72.72 $-$ & 78.28 $-$ & 78.64 $\sim$ & \cellcolor{g}79.91 & \cellcolor{p}58.31 $\sim$ & 60.76 $\sim$ & 61.61 $\sim$ & \cellcolor{g}62.35 \\
\texttt{col} & \cellcolor{p}50.65 $\sim$ & \cellcolor{g}58.00 $+$ & 53.29 $\sim$ & 51.89 & \cellcolor{p}46.96 $\sim$ & \cellcolor{g}53.36 $+$ & 48.01 $\sim$ & 48.66 \\
\texttt{dbt} & 56.90 $\sim$ & 59.90 $-$ & \cellcolor{p}56.39 $-$ & \cellcolor{g}60.30 & 54.91 $\sim$ & 56.13 $\sim$ & \cellcolor{p}53.56 $-$ & \cellcolor{g}56.32 \\
\texttt{ecl} & \cellcolor{p}39.15 $-$ & 52.26 $-$ & 49.49 $-$ & \cellcolor{g}53.30 & \cellcolor{p}36.60 $-$ & \cellcolor{g}56.93 $\sim$ & 54.78 $\sim$ & 56.75 \\
\texttt{frt} & \cellcolor{g}81.85 $+$ & 75.94 $-$ & \cellcolor{p}73.43 $-$ & 80.41 & 72.96 $\sim$ & 71.68 $-$ & \cellcolor{p}69.32 $-$ & \cellcolor{g}75.35 \\
\texttt{gls} & \cellcolor{p}40.08 $-$ & 40.62 $-$ & \cellcolor{g}52.92 $+$ & 47.53 & 33.67 $\sim$ & \cellcolor{p}32.35 $\sim$ & \cellcolor{g}40.35 $\sim$ & 38.49 \\
\texttt{hrt} & \cellcolor{p}82.88 $-$ & 88.02 $-$ & \cellcolor{g}89.40 $\sim$ & 89.31 & \cellcolor{p}76.09 $-$ & 80.98 $\sim$ & 80.02 $\sim$ & \cellcolor{g}81.58 \\
\texttt{hpt} & \cellcolor{g}84.67 $+$ & \cellcolor{p}74.06 $-$ & 75.50 $\sim$ & 75.05 & \cellcolor{p}52.96 $-$ & 59.68 $\sim$ & 59.31 $\sim$ & \cellcolor{g}60.27 \\
\texttt{hcl} & \cellcolor{g}74.67 $+$ & 62.85 $-$ & \cellcolor{p}61.57 $-$ & 64.54 & \cellcolor{g}57.25 $+$ & 49.90 $\sim$ & \cellcolor{p}45.71 $-$ & 50.98 \\
\texttt{irs} & \cellcolor{p}74.90 $-$ & 93.66 $-$ & 94.05 $\sim$ & \cellcolor{g}94.38 & \cellcolor{p}70.10 $-$ & 92.71 $\sim$ & \cellcolor{g}93.56 $\sim$ & 92.93 \\
\texttt{lnd} & 32.84 $-$ & \cellcolor{p}20.63 $-$ & 24.57 $-$ & \cellcolor{g}35.81 & 26.89 $\sim$ & 25.89 $\sim$ & \cellcolor{p}23.65 $-$ & \cellcolor{g}28.65 \\
\texttt{mam} & 79.71 $-$ & 80.32 $-$ & \cellcolor{p}77.12 $-$ & \cellcolor{g}80.45 & \cellcolor{g}78.95 $\sim$ & 78.57 $\sim$ & \cellcolor{p}75.93 $-$ & 78.84 \\
\texttt{pdy} & \cellcolor{g}72.86 $+$ & \cellcolor{p}34.78 $-$ & 34.87 $-$ & 38.92 & \cellcolor{g}71.90 $+$ & \cellcolor{p}34.81 $-$ & 35.12 $-$ & 39.30 \\
\texttt{pis} & 85.21 $-$ & \cellcolor{g}86.26 $\sim$ & \cellcolor{p}84.60 $-$ & 86.24 & 84.05 $-$ & \cellcolor{g}85.44 $\sim$ & \cellcolor{p}83.42 $-$ & 85.23 \\
\texttt{pha} & 75.77 $\sim$ & \cellcolor{p}74.85 $-$ & \cellcolor{g}77.88 $+$ & 76.25 & \cellcolor{p}58.27 $-$ & \cellcolor{g}63.59 $\sim$ & 61.91 $\sim$ & 63.08 \\
\texttt{pre} & \cellcolor{p}99.33 $\sim$ & 99.64 $\sim$ & \cellcolor{g}100.0 $+$ & 99.60 & \cellcolor{p}99.32 $\sim$ & 99.68 $\sim$ & \cellcolor{g}100.0 $+$ & 99.60 \\
\texttt{pmp} & 86.40 $\sim$ & \cellcolor{g}86.78 $\sim$ & \cellcolor{p}86.32 $-$ & 86.74 & \cellcolor{p}85.24 $\sim$ & 85.98 $\sim$ & 85.63 $-$ & \cellcolor{g}86.04 \\
\texttt{rsn} & \cellcolor{p}79.90 $-$ & 83.53 $-$ & \cellcolor{g}85.11 $+$ & 84.26 & \cellcolor{p}81.33 $-$ & 82.97 $-$ & \cellcolor{g}84.86 $+$ & 84.04 \\
\texttt{seg} & \cellcolor{p}87.68 $-$ & 88.52 $-$ & 88.90 $\sim$ & \cellcolor{g}88.95 & \cellcolor{p}87.18 $-$ & 87.92 $-$ & 88.34 $\sim$ & \cellcolor{g}88.43 \\
\texttt{sir} & \cellcolor{p}70.97 $-$ & 83.49 $\sim$ & \cellcolor{g}83.89 $\sim$ & 83.48 & \cellcolor{p}67.48 $-$ & \cellcolor{g}79.44 $\sim$ & 77.42 $\sim$ & 79.34 \\
\texttt{smk} & \cellcolor{g}84.54 $+$ & 71.95 $+$ & 83.42 $+$ & \cellcolor{p}67.57 & \cellcolor{g}83.73 $+$ & 71.17 $+$ & 82.71 $+$ & \cellcolor{p}67.28 \\
\texttt{tae} & \cellcolor{p}52.43 $-$ & 55.85 $-$ & 59.12 $-$ & \cellcolor{g}60.49 & \cellcolor{p}42.60 $-$ & 44.46 $-$ & 47.79 $\sim$ & \cellcolor{g}51.23 \\
\texttt{tip} & \cellcolor{g}76.30 $\sim$ & 76.20 $\sim$ & \cellcolor{p}74.62 $-$ & 76.23 & 74.40 $\sim$ & \cellcolor{g}75.00 $\sim$ & \cellcolor{p}72.84 $-$ & 74.88 \\
\texttt{tit} & \cellcolor{g}64.56 $+$ & 60.55 $\sim$ & 62.72 $+$ & \cellcolor{p}60.39 & \cellcolor{g}65.41 $+$ & 61.34 $\sim$ & 62.31 $\sim$ & \cellcolor{p}60.82 \\
\texttt{wne} & \cellcolor{p}87.67 $-$ & 97.19 $-$ & 97.54 $\sim$ & \cellcolor{g}97.61 & \cellcolor{p}81.85 $-$ & 94.96 $\sim$ & 93.52 $\sim$ & \cellcolor{g}95.34 \\
\texttt{wbc} & 95.28 $-$ & 96.08 $-$ & \cellcolor{p}95.20 $-$ & \cellcolor{g}96.26 & 94.26 $-$ & \cellcolor{g}95.81 $\sim$ & \cellcolor{p}93.61 $-$ & 95.48 \\
\texttt{wpb} & \cellcolor{p}76.83 $-$ & 77.56 $-$ & \cellcolor{g}86.74 $+$ & 82.61 & \cellcolor{p}50.21 $\sim$ & \cellcolor{g}53.13 $\sim$ & 52.19 $\sim$ & 53.00 \\
\texttt{yst} & \cellcolor{p}32.08 $-$ & 38.65 $-$ & 41.62 $-$ & \cellcolor{g}43.79 & \cellcolor{p}25.36 $-$ & 31.95 $-$ & \cellcolor{g}36.58 $\sim$ & 36.45 \\
\bhline{1pt}
Rank & \cellcolor{p}\textit{2.93}$\downarrow^\dag$ & \textit{2.77}$\downarrow^{\dag\dag}$ & \textit{2.40}$\downarrow$ & \cellcolor{g}\textit{1.90} & \cellcolor{p}\textit{3.20}$\downarrow^{\dag\dag}$ & \textit{2.32}$\downarrow^{\dag\dag}$ & \textit{2.63}$\downarrow^{\dag\dag}$ & \cellcolor{g}\textit{1.85} \\
Position & \textit{4} & \textit{3} & \textit{2} & \textit{1} & \textit{4} & \textit{2} & \textit{3} & \textit{1} \\
$+/-/\sim$ & 6/17/7 & 2/22/6 & 8/13/9 & - & 4/15/11 & 2/7/21 & 4/10/16 & - \\
\bhline{1pt}
$p$-value & 0.0327 & 0.000928 & 0.360 & - & 0.00322 & 0.0332 & 0.00578 & - \\
$p_\text{Holm}$-value & 0.0654 & 0.00278 & 0.360 & - & 0.00967 & 0.0332 & 0.0116 & - \\

\bhline{1pt}

\end{tabular}
}
\end{center}
\end{table*}

\begin{table*}[t]
\begin{center}
\caption{Results When the Number of Training Iterations is 50 Epochs, Displaying Average Macro F1 Scores Across 30 Runs.}
\label{tb: f1 score 50 epoch}
\scalebox{0.7}{
\begin{tabular}{c|c c c c|cccc}
\bhline{1pt}
\multicolumn{1}{c|}{\multirow{2}{*}{50 Epochs}} &\multicolumn{4}{c|}{\textsc{Training Macro F1 (\%)}} & \multicolumn{4}{c}{\textsc{Test Macro F1 (\%)}}\\
& UCS & Fuzzy-UCS$_\text{VOTE}$ & Fuzzy-UCS$_\text{SWIN}$ & Fuzzy-UCS$_\text{DS}$ & UCS & Fuzzy-UCS$_\text{VOTE}$ & Fuzzy-UCS$_\text{SWIN}$ & Fuzzy-UCS$_\text{DS}$\\
\bhline{1pt}
\texttt{bnk} & \cellcolor{g}94.84 $+$ & \cellcolor{p}92.55 $-$ & 93.56 $+$ & 92.87 & \cellcolor{g}94.30 $+$ & \cellcolor{p}92.21 $-$ & 92.87 $\sim$ & 92.51 \\
\texttt{can} & \cellcolor{g}96.98 $+$ & 94.83 $-$ & \cellcolor{p}94.31 $-$ & 94.95 & \cellcolor{p}91.00 $-$ & 94.12 $\sim$ & 93.79 $\sim$ & \cellcolor{g}94.17 \\
\texttt{car} & \cellcolor{p}75.85 $-$ & 85.22 $\sim$ & 85.08 $\sim$ & \cellcolor{g}85.58 & \cellcolor{p}58.10 $\sim$ & \cellcolor{g}64.60 $\sim$ & 63.02 $\sim$ & 61.92 \\
\texttt{col} & \cellcolor{g}66.28 $+$ & 62.95 $+$ & \cellcolor{p}56.13 $-$ & 59.76 & 57.78 $\sim$ & \cellcolor{g}59.23 $+$ & \cellcolor{p}52.04 $\sim$ & 55.78 \\
\texttt{dbt} & \cellcolor{g}73.76 $+$ & 70.11 $\sim$ & \cellcolor{p}62.46 $-$ & 70.07 & \cellcolor{g}67.66 $\sim$ & 66.44 $\sim$ & \cellcolor{p}57.81 $-$ & 66.16 \\
\texttt{ecl} & \cellcolor{p}59.77 $-$ & \cellcolor{g}70.47 $\sim$ & 65.33 $-$ & 69.62 & \cellcolor{p}55.93 $-$ & \cellcolor{g}66.03 $\sim$ & 58.65 $-$ & 66.00 \\
\texttt{frt} & \cellcolor{g}91.93 $+$ & 82.79 $-$ & \cellcolor{p}80.06 $-$ & 84.71 & 76.60 $-$ & 77.87 $-$ & \cellcolor{p}74.15 $-$ & \cellcolor{g}80.58 \\
\texttt{gls} & \cellcolor{p}58.00 $-$ & 66.95 $\sim$ & 65.51 $-$ & \cellcolor{g}67.73 & \cellcolor{p}41.58 $-$ & 53.82 $\sim$ & 53.59 $\sim$ & \cellcolor{g}54.87 \\
\texttt{hrt} & \cellcolor{p}90.42 $-$ & 94.60 $-$ & 94.28 $-$ & \cellcolor{g}94.99 & \cellcolor{p}80.36 $\sim$ & 81.13 $\sim$ & \cellcolor{g}82.20 $\sim$ & 81.09 \\
\texttt{hpt} & 89.93 $-$ & 91.22 $-$ & \cellcolor{p}89.48 $-$ & \cellcolor{g}92.12 & \cellcolor{p}51.99 $-$ & 62.03 $\sim$ & 59.07 $\sim$ & \cellcolor{g}62.72 \\
\texttt{hcl} & \cellcolor{g}88.73 $+$ & 84.13 $-$ & \cellcolor{p}79.55 $-$ & 84.84 & \cellcolor{g}59.28 $\sim$ & 59.25 $\sim$ & \cellcolor{p}53.48 $-$ & 59.18 \\
\texttt{irs} & \cellcolor{p}82.57 $-$ & 94.85 $-$ & \cellcolor{g}95.54 $\sim$ & 95.30 & \cellcolor{p}78.11 $-$ & 93.29 $\sim$ & \cellcolor{g}94.12 $\sim$ & 93.51 \\
\texttt{lnd} & \cellcolor{g}52.80 $+$ & \cellcolor{p}22.41 $-$ & 24.01 $-$ & 37.41 & \cellcolor{g}36.27 $+$ & 25.97 $\sim$ & \cellcolor{p}23.61 $-$ & 29.55 \\
\texttt{mam} & \cellcolor{g}82.99 $+$ & 80.84 $-$ & \cellcolor{p}77.42 $-$ & 81.02 & \cellcolor{g}80.49 $\sim$ & 79.55 $\sim$ & \cellcolor{p}76.20 $-$ & 79.86 \\
\texttt{pdy} & \cellcolor{g}89.18 $+$ & \cellcolor{p}35.73 $-$ & 36.09 $-$ & 41.93 & \cellcolor{g}88.20 $+$ & \cellcolor{p}35.79 $-$ & 36.42 $-$ & 42.43 \\
\texttt{pis} & \cellcolor{g}87.22 $+$ & 86.55 $\sim$ & \cellcolor{p}84.77 $-$ & 86.56 & \cellcolor{g}85.74 $\sim$ & 85.49 $\sim$ & \cellcolor{p}83.86 $-$ & 85.52 \\
\texttt{pha} & \cellcolor{p}78.51 $-$ & 81.39 $-$ & \cellcolor{g}86.59 $+$ & 85.23 & \cellcolor{p}60.98 $-$ & \cellcolor{g}65.36 $+$ & 63.61 $\sim$ & 64.64 \\
\texttt{pre} & 99.63 $\sim$ & \cellcolor{p}98.00 $-$ & \cellcolor{g}100.0 $+$ & 99.76 & 99.64 $\sim$ & \cellcolor{p}98.03 $-$ & \cellcolor{g}100.0 $+$ & 99.79 \\
\texttt{pmp} & \cellcolor{g}87.21 $\sim$ & 86.91 $\sim$ & \cellcolor{p}86.78 $-$ & 86.95 & 85.94 $\sim$ & 85.99 $\sim$ & \cellcolor{p}85.84 $\sim$ & \cellcolor{g}86.04 \\
\texttt{rsn} & 84.97 $\sim$ & \cellcolor{p}84.19 $-$ & \cellcolor{g}85.47 $+$ & 85.05 & 84.49 $\sim$ & \cellcolor{p}83.86 $-$ & \cellcolor{g}85.06 $\sim$ & 84.80 \\
\texttt{seg} & \cellcolor{g}95.42 $+$ & \cellcolor{p}89.69 $-$ & 90.05 $\sim$ & 90.07 & \cellcolor{g}92.90 $+$ & \cellcolor{p}89.18 $-$ & 89.25 $\sim$ & 89.68 \\
\texttt{sir} & \cellcolor{p}76.28 $-$ & 84.70 $\sim$ & \cellcolor{g}84.93 $\sim$ & 84.69 & \cellcolor{p}68.41 $-$ & 80.19 $\sim$ & 79.70 $\sim$ & \cellcolor{g}80.44 \\
\texttt{smk} & \cellcolor{g}97.59 $+$ & 95.07 $-$ & \cellcolor{p}92.88 $-$ & 95.42 & \cellcolor{g}96.38 $+$ & 94.61 $\sim$ & \cellcolor{p}92.46 $-$ & 94.66 \\
\texttt{tae} & \cellcolor{p}58.07 $-$ & 62.95 $-$ & 63.59 $\sim$ & \cellcolor{g}63.84 & \cellcolor{p}46.55 $-$ & 51.19 $\sim$ & 51.41 $\sim$ & \cellcolor{g}53.22 \\
\texttt{tip} & \cellcolor{g}79.69 $+$ & 77.16 $-$ & \cellcolor{p}76.03 $-$ & 77.26 & \cellcolor{g}76.83 $+$ & 75.57 $\sim$ & \cellcolor{p}73.98 $-$ & 75.79 \\
\texttt{tit} & \cellcolor{g}68.82 $\sim$ & 68.48 $\sim$ & \cellcolor{p}63.29 $-$ & 68.42 & \cellcolor{g}68.11 $\sim$ & 68.00 $\sim$ & \cellcolor{p}62.38 $-$ & 68.08 \\
\texttt{wne} & \cellcolor{p}93.72 $-$ & 98.33 $-$ & 98.59 $\sim$ & \cellcolor{g}98.79 & \cellcolor{p}88.25 $-$ & 96.01 $\sim$ & 95.44 $\sim$ & \cellcolor{g}96.16 \\
\texttt{wbc} & \cellcolor{g}97.32 $+$ & 96.81 $\sim$ & \cellcolor{p}95.99 $-$ & 96.76 & 95.73 $\sim$ & \cellcolor{g}95.74 $\sim$ & \cellcolor{p}94.66 $-$ & 95.46 \\
\texttt{wpb} & \cellcolor{g}93.22 $+$ & \cellcolor{p}86.89 $-$ & 90.44 $+$ & 88.69 & 54.80 $\sim$ & \cellcolor{g}56.16 $\sim$ & \cellcolor{p}53.64 $\sim$ & 55.71 \\
\texttt{yst} & \cellcolor{g}57.47 $+$ & \cellcolor{p}52.15 $-$ & 52.69 $-$ & 54.56 & \cellcolor{g}46.24 $\sim$ & \cellcolor{p}45.18 $\sim$ & 45.36 $\sim$ & 45.75 \\
\bhline{1pt}
Rank & \textit{2.10}$\downarrow$ & \textit{2.90}$\downarrow^{\dag\dag}$ & \cellcolor{p}\textit{2.97}$\downarrow^{\dag\dag}$ & \cellcolor{g}\textit{2.03} & \textit{2.50}$\downarrow$ & \textit{2.47}$\downarrow^\dag$ & \cellcolor{p}\textit{3.07}$\downarrow^{\dag\dag}$ & \cellcolor{g}\textit{1.97} \\
Position & \textit{2} & \textit{3} & \textit{4} & \textit{1} & \textit{3} & \textit{2} & \textit{4} & \textit{1} \\
$+/-/\sim$ & 16/10/4 & 1/20/9 & 5/19/6 & - & 6/10/14 & 2/6/22 & 1/12/17 & - \\
\bhline{1pt}
$p$-value & 0.821 & 0.000471 & 0.000593 & - & 0.221 & 0.0368 & 0.000110 & - \\
$p_\text{Holm}$-value & 0.821 & 0.00141 & 0.00141 & - & 0.221 & 0.0736 & 0.000331 & - \\

\bhline{1pt}

\end{tabular}
}
\end{center}
\end{table*}

\label{r2-5-2}

Tables \ref{tb: f1 score 10 epoch} and \ref{tb: f1 score 50 epoch} present each system's average training and test {macro F1 scores} and average rank across all datasets At the 10th and 50th epochs, respectively.
Green-shaded values denote the best values among all systems, while peach-shaded values indicate the worst values among all systems. The terms ``Rank'' and ``Position'' denote each system's overall average rank obtained by using the Friedman test and its position in the final ranking, respectively. Statistical results of the Wilcoxon signed-rank test are summarized with symbols wherein ``$+$'', ``$-$'', and ``$\sim$'' represent that the {macro F1 score} of a conventional system is significantly better, worse, and competitive compared to that obtained by the proposed Fuzzy-UCS$_\text{DS}$, respectively. The ``$p$-value'' and ``$p_\text{Holm}$-value'' are derived from the Wilcoxon signed-rank test and the Holm-adjusted Wilcoxon signed-rank test, respectively. Arrows denote whether the rank improved or declined compared to Fuzzy-UCS$_\text{DS}$. $\dag$ ($\dag\dag$) indicates statistically significant differences compared to Fuzzy-UCS$_\text{DS}$, i.e., $p$-value ($p_\text{Holm}$-value) $<\alpha=0.05$.

Based on Tables \ref{tb: f1 score 10 epoch} and \ref{tb: f1 score 50 epoch}, {Fuzzy-UCS$_\text{DS}$ recorded significantly higher test macro F1 scores than all other systems at both the 10th epoch ($p_\text{Holm}<0.05$) and 50th epoch when compared to Fuzzy-UCS$_\text{VOTE}$ ($p<0.05$) and Fuzzy-UCS$_\text{SWIN}$ ($p_\text{Holm}<0.05$), as well as higher average ranks than UCS.}  Furthermore, Tables \ref{tb: f1 score 10 epoch} and \ref{tb: f1 score 50 epoch} reveal that Fuzzy-UCS$_\text{DS}$ did not register the lowest {macro F1 scores} for any problem regardless of the number of training epochs, except for the \texttt{smk} and \texttt{tit} problems in training and test {macro F1 scores} at the 10th epoch.  These findings underscore that Fuzzy-UCS$_\text{DS}$ is a robust system, consistently exhibiting high classification performance across a diverse range of problems. 

{Additional experimental results are provided in Appendices \ref{sec: sup Experimental Results on Standard Accuracy}-\ref{sec: sup comparison with state-of-the-art classification methods}. In Appendix \ref{sec: sup Experimental Results on Standard Accuracy}, \ds\ demonstrated significantly higher test classification accuracy compared to UCS, \vote, and \swin. 
            In Appendix \ref{sec: sup impact of dataset selection and ground truth}, \ds\ demonstrated superior performance in artificial noiseless problems compared to \vote\ and \swin. For larger-scale datasets (Appendix \ref{sec: sup scalability considerations}), \ds\ maintains stable performance regardless of dataset size, demonstrating its robustness and scalability. 
            Appendix \ref{sec: sup comparison with a state-of-the-art non-fuzzy lcs} revealed the following findings: (i) \all\ generated significantly fewer rules compared to both UCS and a state-of-the-art nonfuzzy LCS (i.e., scikit-ExSTraCS \cite{urbanowicz2015exstracs}); (ii) while scikit-ExSTraCS achieved higher test accuracy on many datasets, \ds\ offers unique advantages in terms of interpretability and simplicity; and (iii) future work will explore incorporating scikit-ExSTraCS's noise-handling mechanisms into \ds\ to further enhance its capabilities.
            In comparisons with modern classification methods (Appendix \ref{sec: sup comparison with state-of-the-art classification methods}), Random Forest and XGBoost achieved better predictive performance on many datasets than \ds, demonstrating a clear trade-off: while these methods offer superior classification accuracy, they lead to significantly higher model complexity.}
\label{r2-6}

\subsection{Discussion}
\label{ss: discussion}

\subsubsection{Comparison With the UCS Nonfuzzy-Classifier System}
\label{sss: comparison with the ucs nonfuzzy classifier system}
\label{r2-5-3}

Table \ref{tb: f1 score 10 epoch} shows that during the initial learning phase (i.e., after the 10th epoch), UCS recorded the lowest ranks in terms of training and test {macro F1 scores}. 
This lower performance can be attributed to the complexity of optimizing non-linguistic rules in UCS. Each dimension in the UCS rule antecedent is represented by two continuous values defining the matching interval’s lower and upper bounds. In contrast, Fuzzy-UCS$_*$ employs the pre-specified linguistic terms to discretely partition the input space for each dimension of the rule antecedent, thus reducing the complexity of the search space. Consequently, UCS, with its relatively large rule search space, shows weaker performance than Fuzzy-UCS$_*$ in the early stages of learning.

Conversely, with adequate learning time (i.e., 50 epochs), UCS enhances its training {macro F1 scores} and demonstrates performance that significantly competes with Fuzzy-UCS$_\text{DS}$, which recorded the highest rank in training data classification ($p_\text{Holm}={0.821}$, cf. Table \ref{tb: f1 score 50 epoch}). Moreover, as evidenced by the results of the Wilcoxon signed-rank test, UCS recorded significantly higher training {macro F1 scores} compared to Fuzzy-UCS$_\text{DS}$ in {16} problems. This count exceeds the number of problems where Fuzzy-UCS$_\text{DS}$ recorded significantly higher training {macro F1 scores} compared to UCS, which is 10 problems. This enhancement results from the non-linguistic rules’ ability to accurately capture the training data through their capacity to partition the input space arbitrarily. However, the rank of UCS on test data classification remains lower than that of Fuzzy-UCS$_\text{VOTE}$ and Fuzzy-UCS$_\text{DS}$. This finding aligns with reported experimental results in \cite{shiraishi2023fuzzy}, which suggests that nonfuzzy rules tend to overfit training data, thus reducing their effectiveness on unseen test data. Indeed, while UCS shows significantly higher training {macro F1 scores} than Fuzzy-UCS$_\text{DS}$ in high-dimensional problems with greater than or equal to 30 dimensions (e.g., {\texttt{frt}}) and problems with missing attributes (e.g., {\texttt{hpt}}), its test {macro F1 score} is significantly lower. 

Thus, in situations characterized by inherent uncertainty, such as real-world classification problems, fuzzy systems like Fuzzy-UCS$_\text{DS}$ demonstrate greater efficacy than nonfuzzy systems like UCS. However, it is worth noting that linguistic rules might not always be advantageous, especially in cases like the \texttt{pdy} problem where all attributes are continuous, and more specific rules are required for accurate classification \cite{shiraishi2022absumption}.

\subsubsection{Impact of the DS Theory-based Class Inference Scheme}
\label{r2-5-4}
\label{sss: Impact of the DS Theory-based Class Inference Scheme}

As shown in Tables \ref{tb: f1 score 10 epoch} and \ref{tb: f1 score 50 epoch}, Fuzzy-UCS$_\text{DS}$ demonstrates a consistent improvement in training and test {macro F1 scores} from the initial learning phase (10 epochs) to the later learning phase (50 epochs) across many datasets. 
This tendency reveals that the DS theory-based inference scheme becomes more effective as the learning progresses and the system’s fuzzy rules are further optimized. 
While Fuzzy-UCS$_\text{DS}$ already shows effective performance in the initial learning phase with limited rule optimization (i.e., after 10 epochs), its capability is notably enhanced when more epochs for rule adjustment is allowed (i.e., 50 epochs).
The observed slight decrease in the average ranking of Fuzzy-UCS$_\text{DS}$ from 10 to 50 epochs does not indicate a decline in performance. This can be attributed to the improvement of some other systems in the later learning phase. Unlike Fuzzy-UCS$_\text{DS}$, UCS and the other two Fuzzy-UCS$_*$ variants may require a more comprehensive set of well-adjusted rules to achieve high performance. On the other hand, Fuzzy-UCS$_\text{DS}$ can efficiently utilize less-adjusted rules in the initial learning phase and continues to perform robustly as the rules become more adjusted in the later learning phase.

A key strength of Fuzzy-UCS$_\text{DS}$ is its capacity to mitigate ``false overconfidence'' in class inference, especially when it is based on uncertain training data. This is achieved by quantitatively expressing uncertainty (i.e., the ``I don't know'' state) through the complete ignorance hypothesis $\mathbf{\Theta}$, a fundamental concept of the DS theory. At the same time, Fuzzy-UCS$_\text{DS}$ utilizes the strength of fuzzy rules in reducing overfitting, as noted in \cite{shiraishi2023fuzzy}. This contributes to the improvement of the reliability and accuracy of the classification process. The combination of DS theory's robustness with the flexibility of fuzzy rules emphasizes the effectiveness of Fuzzy-UCS$_\text{DS}$. This dual approach is effective in various learning tasks. {Notably, in datasets with pronounced class imbalance (e.g., \texttt{ecl}, \texttt{frt}, \texttt{gls}, \texttt{yst}), no significant performance degradation was observed for Fuzzy-UCS$_\text{DS}$ at either epoch point compared to other systems. This suggests that Fuzzy-UCS$_\text{DS}$ helps mitigate bias introduced by class imbalance, leveraging its ability to capture uncertain (i.e., class-imbalanced) information through the DS theory, as discussed in \cite{tian2022reliability}.}

\subsubsection{Complexity of Each Class Inference Scheme}
\label{sss: relative complexities of classifiers}
{In the context of L(F)CSs, complexity is often defined by the number of rules within the system \cite{liu2021visualizations}. From this perspective, there is no difference in complexity among the Fuzzy-UCS variants—\vote, \swin, and \ds—as they all utilize the same ruleset generated during training, that is, they do not affect the rule generation process.

However, when considering the prediction-level complexity, differences arise:
 \begin{itemize}
      \item \textbf{Voting-based inference (\vote):} This scheme typically results in moderate complexity due to its reliance on aggregating votes from multiple rules. It provides a good balance between computational efficiency and robustness by leveraging collective decision-making but may not handle uncertainty as effectively as more complex schemes. Our results indicate that \vote\ is inferior to \ds\ in terms of average rank in test {macro F1 scores} (cf. Section \ref{ss: results})

      \item \textbf{Single-winner-based inference (\swin):} This scheme generally has lower complexity as it selects a single rule with the highest fitness for decision-making. However, it may sacrifice robustness due to its reliance on individual rule performance. Our results indicate that \swin\ is inferior to \vote\ and \ds\ in terms of average rank in test {macro F1 scores} (cf. Section \ref{ss: results}).

      \item \textbf{DS theory-based inference (\ds):} This scheme tends to have higher complexity due to its comprehensive approach to uncertainty management. It requires calculating and combining belief masses, which adds computational layers but enhances robustness and interpretability by providing confidence measures for decisions. Our results indicate that \ds\ is superior to \vote\ and \swin\ in terms of average rank in test {macro F1 scores} (cf. Section \ref{ss: results}), but takes longer runtime than \vote\ and \swin\ (cf. Section \ref{ss: runtime comparison}).
  \end{itemize}

  In conclusion, while the rule-level complexity remains consistent across these schemes, the choice of inference scheme should be guided by the specific requirements of the application domain. For tasks that demand high interpretability and robustness, such as in healthcare or finance, the \ds\ scheme offers significant advantages despite its higher complexity. The ability to quantify uncertainty and provide confidence measures can be crucial in these sensitive domains. Conversely, applications with strict computational constraints or real-time processing requirements might benefit more from simpler schemes like \swin, which offers faster decision-making at the expense of some robustness.}

\section{Analysis}
\label{sec: analysis}
{This section further provides analytical results: Section \ref{ss: runtime comparison} presents the runtime comparisons; Section \ref{ss: decision boundaries} examines the characteristics of decision boundaries provided by each class inference scheme; Section \ref{ss: visualization belief mass} explains how to interpret the inference results acquired by \ds; and Section \ref{ss: quantification of the i don't know state} quantifies the ``I don't know'' state in \ds.}

\subsection{Runtime Comparisons}
\label{ss: runtime comparison}
\begin{table*}[b]
\begin{center}
\caption{Average Runtime to Complete One Run.}
\label{tb: runtime}
\scalebox{0.8}{
\begin{tabular}{cccc|cccc}
\bhline{1pt}
\multicolumn{8}{c}{\textsc{Runtime (sec.)}}\\
\multicolumn{4}{c|}{10 Epochs} & \multicolumn{4}{c}{50 Epochs}\\
 UCS & Fuzzy-UCS$_\text{VOTE}$ & Fuzzy-UCS$_\text{SWIN}$ & Fuzzy-UCS$_\text{DS}$ & UCS & Fuzzy-UCS$_\text{VOTE}$ & Fuzzy-UCS$_\text{SWIN}$ & Fuzzy-UCS$_\text{DS}$\\
\bhline{1pt}
\cellcolor{g}1.478 & 3.214 & 3.147 & \cellcolor{p}7.174 & \cellcolor{g}9.113 & 49.93 & 49.26 & \cellcolor{p}73.33\\

\bhline{1pt}

\end{tabular}
}
\end{center}
\end{table*}

{We conduct runtime comparisons to assess the computational efficiency of the proposed DS theory-based class inference scheme (Fuzzy-UCS$_\text{DS}$). Specifically, we compare the runtimes of UCS and each variant of Fuzzy-UCS$_*$.
All experiments are conducted using our implementations in Julia, on an Intel\textregistered\ Core\texttrademark\ i7-9700 CPU at 3.00 GHz with 16 GB RAM, following the experimental setup outlined in Section \ref{ss: experimental setup}.

Table \ref{tb: runtime} presents the runtime required to complete one run for UCS and each variant of Fuzzy-UCS$_*$ At the 10th and 50th epochs. Green-shaded values indicate the best (shortest) runtimes, while peach-shaded values denote the worst (longest) runtimes.

As shown in Table \ref{tb: runtime}, Fuzzy-UCS$_\text{DS}$ recorded the longest runtime among all systems at both the 10th and 50th epochs. Specifically, compared to Fuzzy-UCS$_\text{VOTE}$ and Fuzzy-UCS$_\text{SWIN}$, Fuzzy-UCS$_\text{DS}$ required approximately 1.5 to 2 times longer to complete. This increase in runtime is attributable to the computational complexity introduced by the DS theory-based inference mechanism, particularly when the number of hypotheses (i.e., class labels) is large. This complexity represents a notable limitation of our approach.
Furthermore, nonfuzzy systems such as UCS demonstrated significantly shorter runtimes compared to their fuzzy counterparts like Fuzzy-UCS. This difference is primarily due to the necessity in fuzzy systems to calculate membership degrees for each rule every time the match set $[M]$ is formed, which serves as a computational bottleneck.

While Fuzzy-UCS$_\text{DS}$ offers significant improvements in classification performance, its increased runtime is an important consideration. Future work may explore optimization techniques to mitigate this computational cost, thereby enhancing the practicality of the DS theory-based inference scheme in real-world applications. }

\subsection{Visualization of Decision Boundaries}
\label{ss: decision boundaries}

\begin{figure*}[t]
  \begin{minipage}[b]{0.24\linewidth}
    \centering
    \includegraphics[keepaspectratio, scale=0.3]{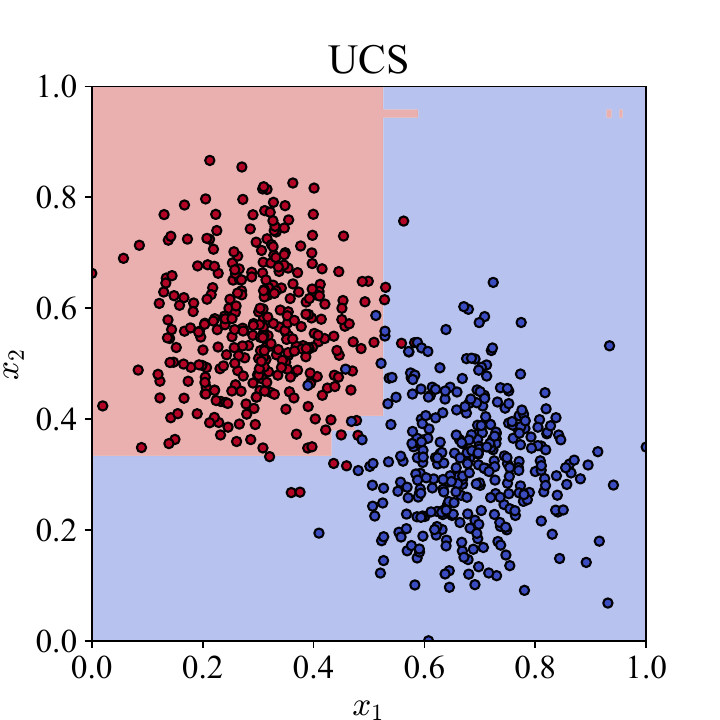}
  \end{minipage}
  \begin{minipage}[b]{0.24\linewidth}
    \centering
    \includegraphics[keepaspectratio, scale=0.3]{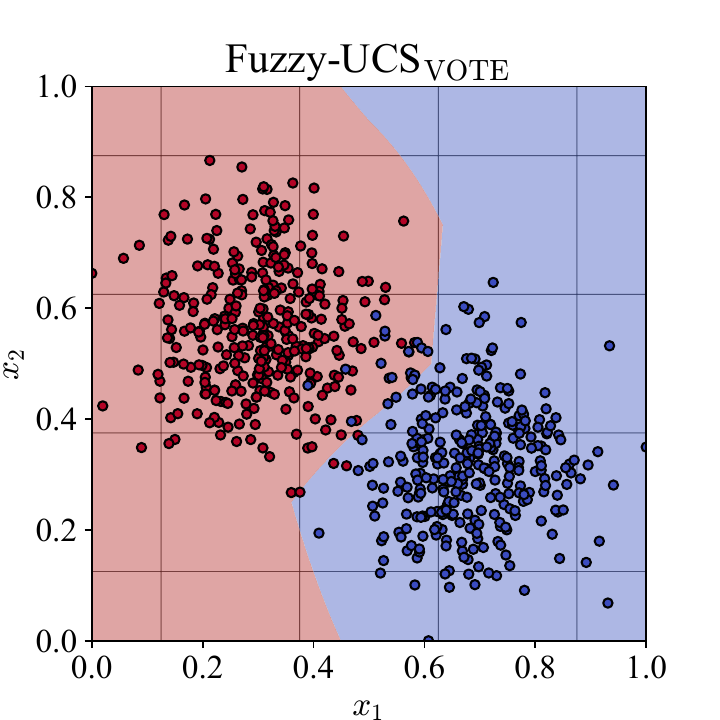}
  \end{minipage}
    \begin{minipage}[b]{0.24\linewidth}
    \centering
    \includegraphics[keepaspectratio, scale=0.3]{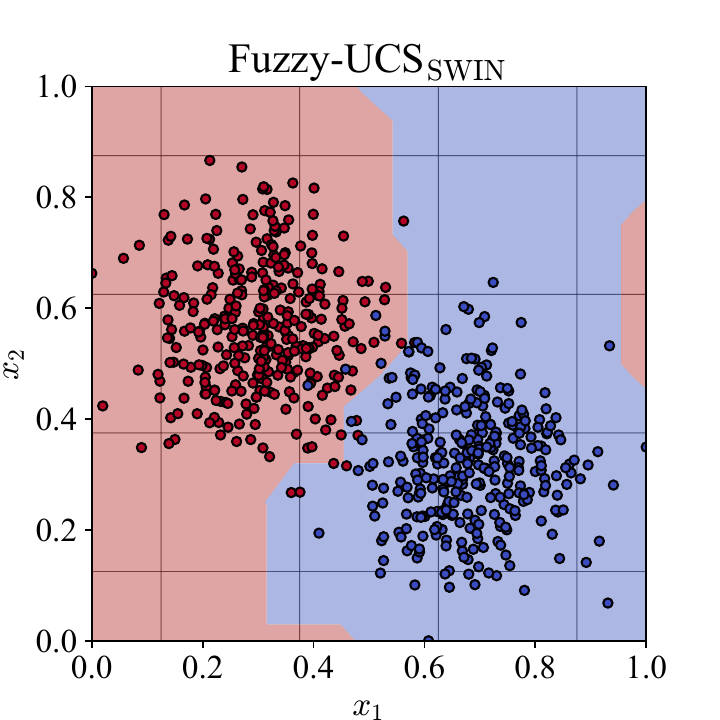}
  \end{minipage}
    \begin{minipage}[b]{0.24\linewidth}
    \centering
    \includegraphics[keepaspectratio, scale=0.3]{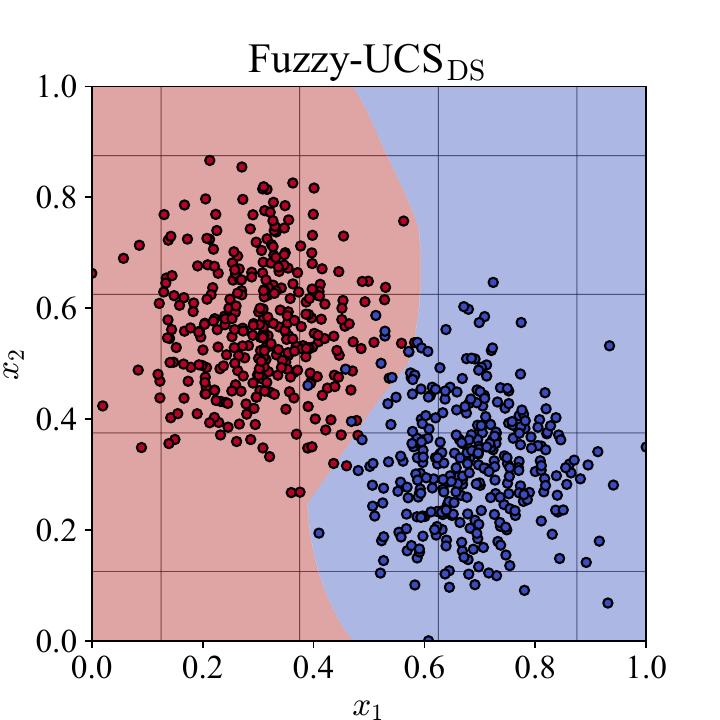}
  \end{minipage}
  \caption{Decision boundaries obtained with UCS and Fuzzy-UCS$_*$. UCS, Fuzzy-UCS$_\text{VOTE}$, Fuzzy-UCS$_\text{SWIN}$, and Fuzzy-UCS$_\text{DS}$ achieved 94.65 \%, 96.84\%,  97.45\%, and 97.74\% average training accuracy, respectively.}
  \label{fig: analysis 1}
  \vspace{-3mm}
\end{figure*}

This subsection aims to visually analyze the decision boundaries formed by each system. We use an artificial dataset generated by \textit{scikit-learn} \cite{scikit-learn} for a comparative experiment. As depicted in Fig. \ref{fig: analysis 1}, the dataset comprises a class-balanced binary classification problem with a total of 600 two-dimensional data points. Data points from class $c_1$ are shown in blue and class $c_2$ in red. This problem is difficult for UCS with interval-based nonfuzzy rules and also for Fuzzy-UCS$_*$ with linguistic fuzzy rules because it involves some noisy data points and requires curved decision boundaries for effective classification. The system configuration aligns with the experiment in Section \ref{sec: experiment}, and the number of training iterations is set to 10 epochs.

Fig. \ref{fig: analysis 1} illustrates the inferred class assignments $\hat{c}\in\{c_1, c_2\}$ by each system in the input space, shown in blue and red, respectively (resolution: 1000 $\times$ 1000). For Fuzzy-UCS$_*$, the grid in each plot indicates the $5 \times 5$ partition of the input space by the intersection points of the adjacent membership functions for the five linguistic terms. Note again that all three Fuzzy-UCS$_*$ variants use the same ${{[P]}}$. The training accuracy achieved is detailed in the figure captions, representing averages across 30 runs with different seeds. A randomly selected run among the 30 runs is selected for visualization in Fig. \ref{fig: analysis 1}.

The decision boundaries offer several insights:
\begin{itemize}
    \item UCS only creates boundaries that are axis-parallel.
    \item Fuzzy-UCS$_\text{VOTE}$ forms a slightly smooth, oblique boundary due to the collective voting of multiple rules.
    \item Fuzzy-UCS$_\text{SWIN}$ presents a steeper boundary than Fuzzy-UCS$_\text{VOTE}$ since it relies on a single rule with the highest $\mu_{\mathbf{A}^k}(\mathbf{x}) \cdot F^k $ for inference. This leads to potentially unintuitive class assignments in subregions lacking training data, such as around (1.0, 0.6). Some parts of the boundary are axis-parallel as in UCS, while others are non-axis-parallel as in Fuzzy-UCS$_\text{VOTE}$.
    \item Fuzzy-UCS$_\text{DS}$ exhibits the highest training accuracy among all four systems. This is attributed to the combination of belief masses from multiple rules across different hypotheses (i.e., $\{\theta_1\}, \{\theta_2\}, \mathbf{\Theta}$) using Dempster's rule of combination. The smoothest boundary is obtained by Fuzzy-UCS$_\text{DS}$ among the four systems in Fig. \ref{fig: analysis 1}.
\end{itemize}

\subsection{Visualization of Belief Masses for Each Hypothesis}
\label{ss: visualization belief mass}

\begin{figure*}[t]
  \begin{minipage}[b]{0.24\linewidth}
    \centering
    \includegraphics[keepaspectratio, scale=0.3]{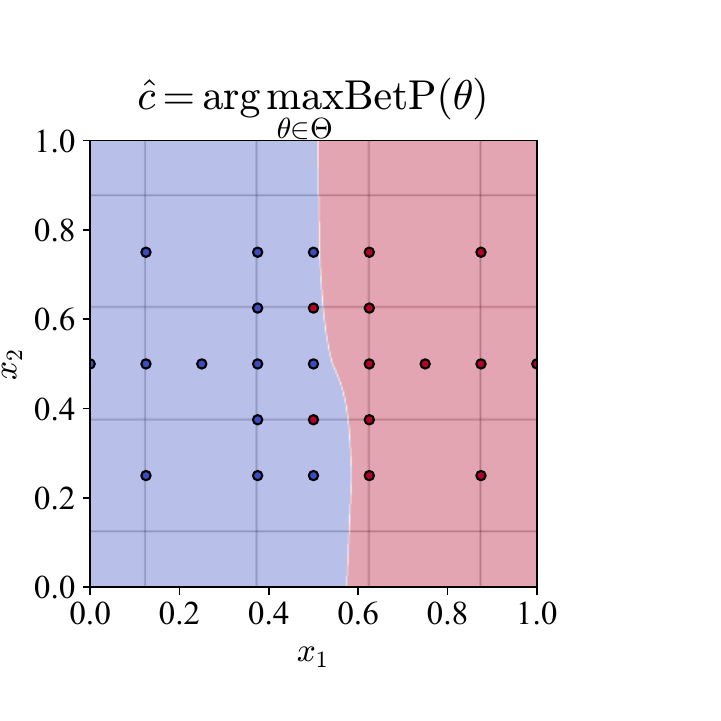}
  \end{minipage}
  \begin{minipage}[b]{0.24\linewidth}
    \centering
    \includegraphics[keepaspectratio, scale=0.3]{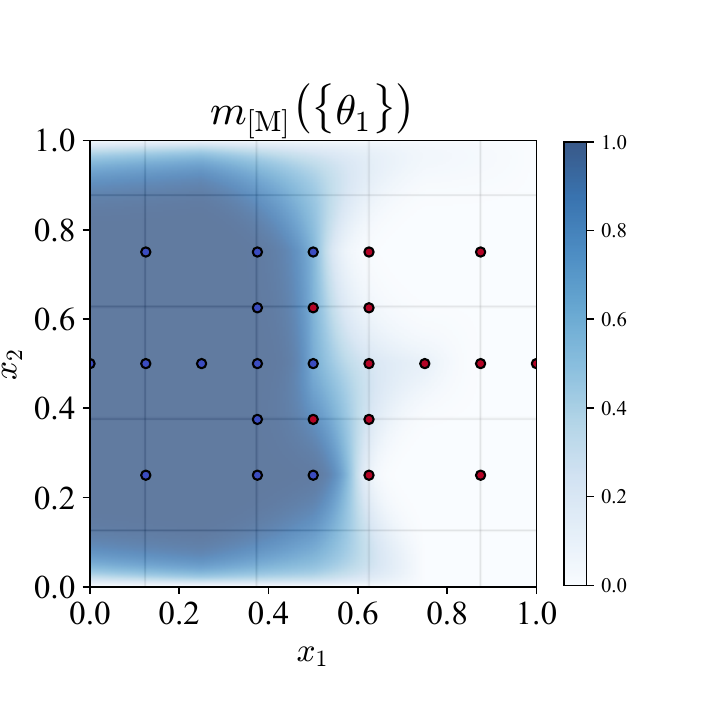}
  \end{minipage}
    \begin{minipage}[b]{0.24\linewidth}
    \centering
    \includegraphics[keepaspectratio, scale=0.3]{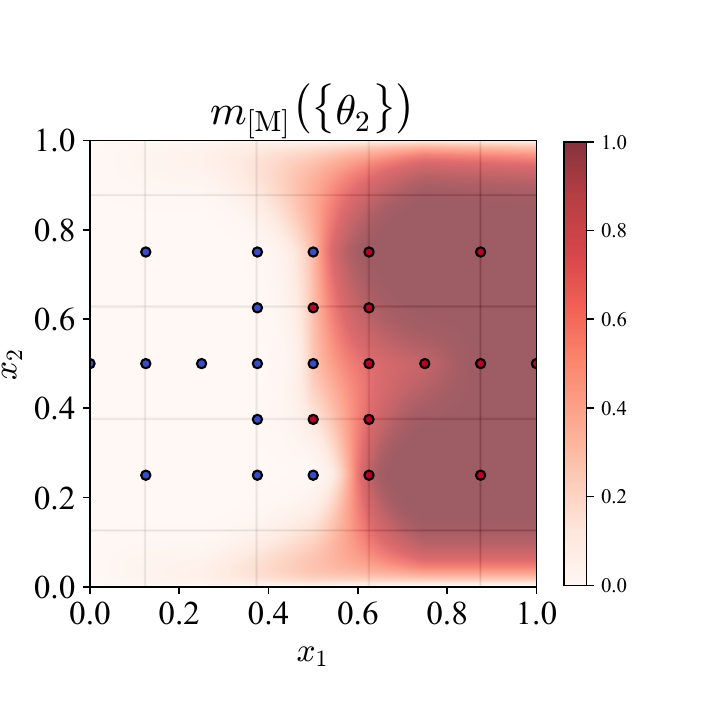}
  \end{minipage}
    \begin{minipage}[b]{0.24\linewidth}
    \centering
    \includegraphics[keepaspectratio, scale=0.3]{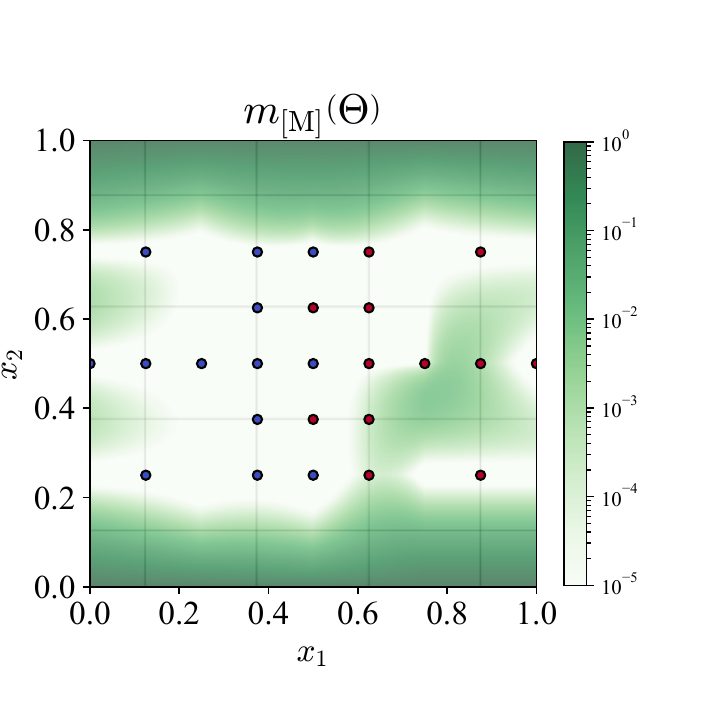}
  \end{minipage}
  \caption{Class assignments $\hat{c}\in\{c_1, c_2\}$ by Fuzzy-UCS$_\text{DS}$ and belief mass landscapes for each hypothesis: $m_{{[M]}}(\{\theta_1\})$: class $c_1$, $m_{{[M]}}(\{\theta_2\})$: class $c_2$, $m_{{[M]}}(\mathbf{\Theta})$: complete ignorance (``I don't know''). All figures are a result of a randomly selected run.}
  \label{fig: analysis 2 random}
  \vspace{-3mm}
\end{figure*}

This subsection aims to visually interpret the inference results by Fuzzy-UCS$_\text{DS}$. We use an artificial dataset in Fig. \ref{fig: analysis 2 random}, which is inspired from \cite{masson2008ecm}. As shown in Fig. \ref{fig: analysis 2 random}, the dataset is a binary classification problem with 25 data points: including 13 blue (class $c_1$) and 12 red (class $c_2$) data points. No data points are given in the top and bottom regions where $x_2\in[0.0, 0.25)$ and $x_2\in(0.75,1.0]$. On the vertical line with $x_1=0.5$, three blue points and two red points are given. The 25 data points show the perfect symmetry with respect to the horizontal line with $x_2=0.5$. The setup for Fuzzy-UCS$_\text{DS}$ and the training epochs remain the same as in Section \ref{ss: decision boundaries}.

Fig. \ref{fig: analysis 2 random} displays class assignments $\hat{c}\in\{c_1, c_2\}$ by Fuzzy-UCS$_\text{DS}$ and belief mass landscapes for each hypothesis ($m_{{[M]}}(\{\theta_1\})$, $m_{{[M]}}(\{\theta_2\})$, $m_{{[M]}}(\mathbf{\Theta})$). The color in each belief mass figure shows the value of the belief mass at the corresponding point in the input space (i.e., the darker color shows the larger value of the belief mass, and the zero belief mass is shown by white color). As explained in Section \ref{sec: ds theory}, the sum of the belief mass values over the three figures is always 1 at any point in the input space. The landscape for $m_{{[M]}}(\mathbf{\Theta})$ is shown on a logarithmic scale for clarity. The average training accuracy by Fuzzy-UCS$_\text{DS}$ is 91.33\% over 30 runs with different seeds. A randomly selected run among the 30 runs is selected for visualization in Fig. \ref{fig: analysis 2 random}.

The results show that Fuzzy-UCS$_\text{DS}$ forms a smooth vertical class boundary, correctly classifying all training data, except for those along the line of $x_1=0.5$. Investigating the belief mass landscape provides several insights:
\begin{itemize}
    \item 
    In subregions close to the training data, the values for the ``I don't know'' belief mass, $m_{{[M]}}(\mathbf{\Theta})$, are close to 0.
    This is because Fuzzy-UCS$_*$ activates the covering operator which ensures that the sum of correct answer rules' membership degrees is always greater than 1 (cf. Alg. \ref{alg: fuzzy-ucs training}, line 8). As a result, total membership degrees in the neighborhood of training data points tend to be high. Consequently, this leads to low values of $m_{{[M]}}(\mathbf{\Theta})$ in these areas.

    \item 
    In the two subregions with no data points (i.e., $x_2 \leq 0.2$ and $x_2 > 0.8$), the function value of $m_{{[M]}}(\mathbf{\Theta})$ is large. This indicates greater uncertainty due to fewer rules covering these subregions.

    \item 
    In the challenging subregion with mixed classes (around $x_1=0.5$), $m_{{[M]}}(\{\theta_1\})$ and $m_{{[M]}}(\{\theta_2\})$ are close to 0.5, and $m_{{[M]}}(\mathbf{\Theta})$ is close to 0. This reflects the presence of rules that equally support conflicting hypotheses.

    \item In simpler subregions (e.g., around $(0.0,0.5)$ and $(1.0,0.5)$), where classification is easy, $m_{{[M]}}(\mathbf{\Theta})$ is close to 0, and a high belief mass is observed for a specific class hypothesis.

\end{itemize}

\begin{figure*}[t]
  \begin{minipage}[b]{0.24\linewidth}
    \centering
    \includegraphics[keepaspectratio, scale=0.3]{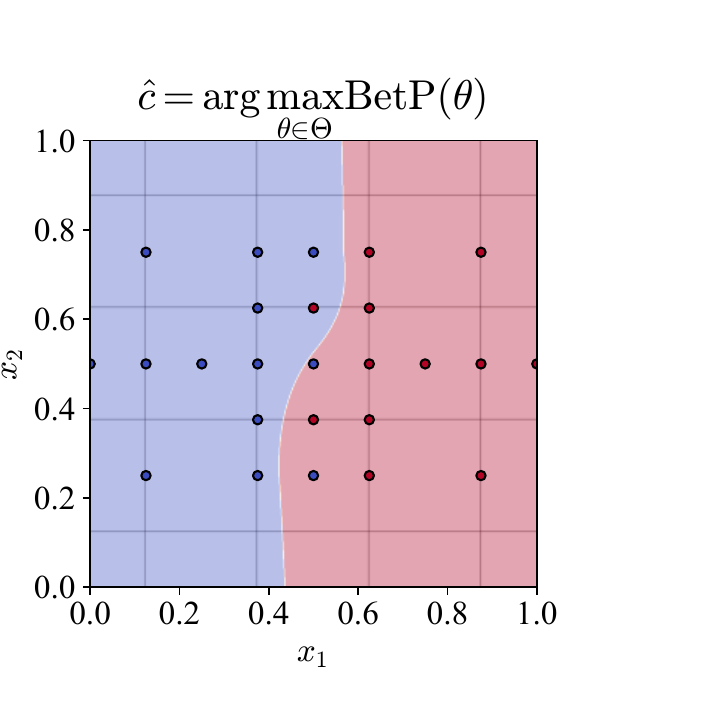}
  \end{minipage}
  \begin{minipage}[b]{0.24\linewidth}
    \centering
    \includegraphics[keepaspectratio, scale=0.3]{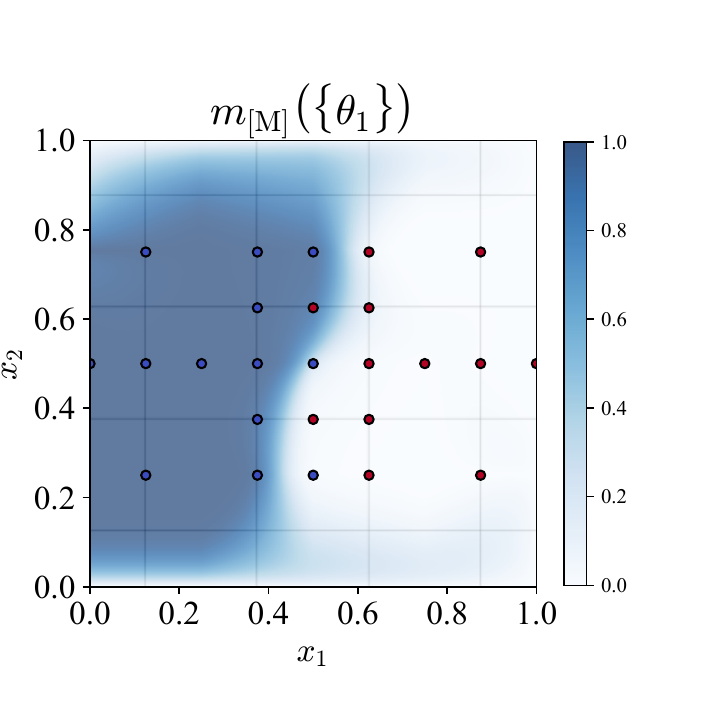}
  \end{minipage}
    \begin{minipage}[b]{0.24\linewidth}
    \centering
    \includegraphics[keepaspectratio, scale=0.3]{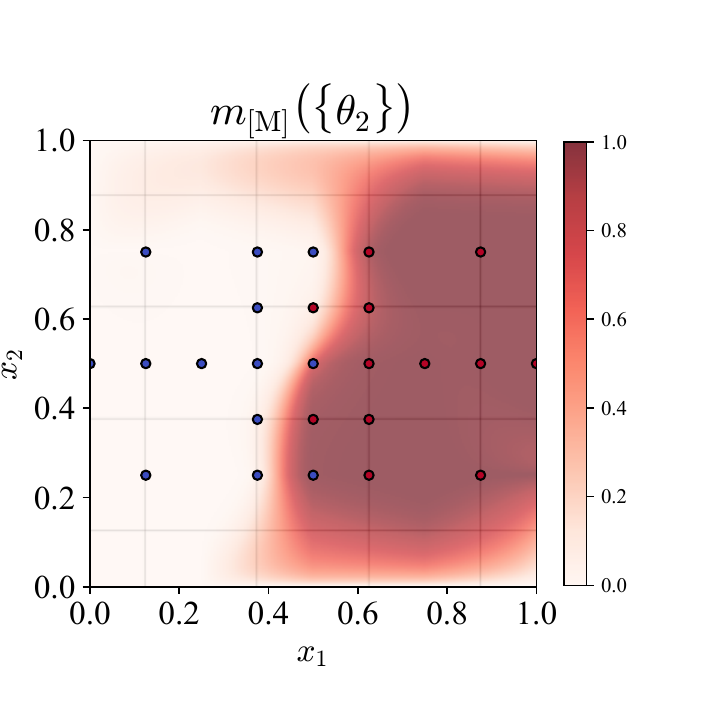}
  \end{minipage}
    \begin{minipage}[b]{0.24\linewidth}
    \centering
    \includegraphics[keepaspectratio, scale=0.3]{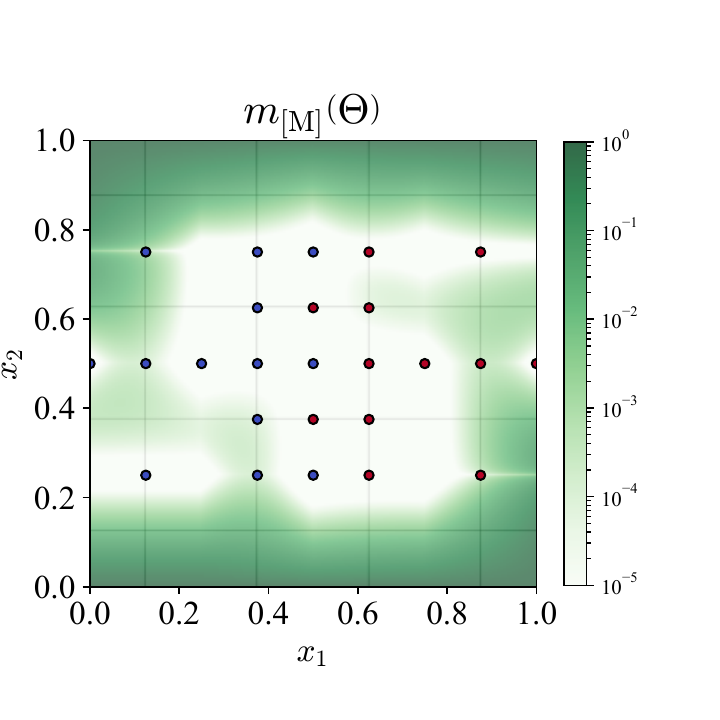}
  \end{minipage}
  \caption{Class assignments by Fuzzy-UCS$_\text{DS}$ and belief mass landscapes for each hypothesis. All figures are a result of a single run that has the worst training accuracy.}
  \label{fig: analysis 2 worst}
  \vspace{-3mm}
\end{figure*}

In a nutshell, the belief masses across all hypotheses help interpret the inference result $\hat{c}$:
\begin{itemize}

\item In subregions where $m_{{[M]}}(\mathbf{\Theta})$ is close to 0, and only one specific class hypothesis has a high belief mass, classification by fuzzy rules is easy. Hence, $\hat{c}$ is likely to be correct.
\item In subregions where $m_{{[M]}}(\mathbf{\Theta})$ is close to 0 and several specific class hypotheses have similar belief masses, classification by fuzzy rules is difficult. Hence, $\hat{c}$ may be unreliable. Those subregions are likely around the class boundary.
\item In subregions where $m_{{[M]}}(\mathbf{\Theta})$ is close to 1, it is unclear whether classification by fuzzy rules is easy or difficult. Hence, the reliability of $\hat{c}$ is uncertain. Those subregions likely have no training data points.
\end{itemize}

This analysis demonstrates how the belief mass function for each hypothesis can provide insights into the confidence and uncertainty associated with these classifications.

In Fig. \ref{fig: analysis 2 random}, we can observe that each result (e.g., classification boundary in the left-most figure) is not symmetric with respect to the horizontal line with $x_1=0.5$. This is because the fuzzy rule generation in Fuzzy-UCS depends on the presentation order of the given training data points. Whereas slightly different results are obtained from different runs with different seeds, we can always obtain the above-mentioned observation. In Fig. \ref{fig: analysis 2 worst}, we show the results of a single run that has the worst training data accuracy 88.00\%. We can see that similar belief mass functions are obtained in Figs. \ref{fig: analysis 2 random} and \ref{fig: analysis 2 worst}.

\subsection{Quantification of the ``I Don't Know'' State}
\label{ss: quantification of the i don't know state}
{The DS theory-based class inference scheme assigns a belief mass to the ``I don't know'' state when high uncertainty is detected during classification. This subsection reports the belief mass associated with the ``I don't know'' state for all data points in both the training and test datasets., providing insights into how our approach manages uncertainty. A higher ``I don't know'' belief mass indicates greater uncertainty in classification. Therefore, it is expected that Fuzzy-UCS$_\text{DS}$ will output a higher ``I don't know'' belief mass on test datasets, which contain unseen data points, compared to training datasets. The experimental setup adheres to the conditions outlined in Section \ref{ss: experimental setup}. 

\begin{table*}[t]
\begin{center}
\caption{Results Displaying Average ``I Don't Know'' Belief Mass Across 30 Runs.}
\label{tb: idk}
\scalebox{0.7}{
\begin{tabular}{c|cc|cc|cc}
\bhline{1pt}
 &\multicolumn{6}{c}{\textsc{$m_{\mathcal{D}}(\mathbf{\Theta})$: ``I Don't Know''}}\\
 & \multicolumn{2}{c|}{5 Epochs} & \multicolumn{2}{c|}{10 Epochs}&\multicolumn{2}{c}{50 Epochs}\\
 & Training & Test & Training & Test & Training & Test \\
 \bhline{1pt}
\texttt{bnk} & \cellcolor{g}0.8017 & 0.8031 $\sim$ & 0.9789 & \cellcolor{g}0.9751 $\sim$ & 0.9803 & \cellcolor{g}0.9797 $\sim$ \\
\texttt{can} & \cellcolor{g}0.04141 & 0.04784 $\sim$ & \cellcolor{g}0.01276 & 0.02871 $-$ & \cellcolor{g}0.004121 & 0.02284 $-$ \\
\texttt{car} & \cellcolor{g}0.008148 & 0.03333 $\sim$ & \cellcolor{g}0.0007407 & 0.01333 $\sim$ & 0.4407 & \cellcolor{g}0.4400 $\sim$ \\
\texttt{col} & \cellcolor{g}0.4514 & 0.4516 $\sim$ & \cellcolor{g}0.7001 & 0.7054 $\sim$ & 0.9781 & \cellcolor{g}0.9731 $\sim$ \\
\texttt{dbt} & \cellcolor{g}0.6629 & 0.6727 $\sim$ & 0.9413 & \cellcolor{g}0.9377 $\sim$ & 0.9253 & \cellcolor{g}0.9238 $\sim$ \\
\texttt{ecl} & 0.2845 & \cellcolor{g}0.2767 $\sim$ & 0.5311 & \cellcolor{g}0.5069 $\sim$ & 0.6077 & \cellcolor{g}0.5912 $\sim$ \\
\texttt{frt} & \cellcolor{g}0.02212 & 0.02273 $\sim$ & \cellcolor{g}0.02679 & 0.02785 $\sim$ & \cellcolor{g}0.01153 & 0.01357 $\sim$ \\
\texttt{gls} & 0.6045 & \cellcolor{g}0.5971 $\sim$ & \cellcolor{g}0.7670 & 0.7857 $\sim$ & \cellcolor{g}0.6359 & 0.6486 $\sim$ \\
\texttt{hrt} & \cellcolor{g}0.01391 & 0.01516 $\sim$ & 0.03634 & \cellcolor{g}0.03578 $\sim$ & \cellcolor{g}0.005305 & 0.01687 $-$ \\
\texttt{hpt} & \cellcolor{g}0.1381 & 0.1910 $-$ & \cellcolor{g}0.05583 & 0.09489 $-$ & \cellcolor{g}0.01826 & 0.09713 $-$ \\
\texttt{hcl} & \cellcolor{g}0.07774 & 0.1428 $-$ & \cellcolor{g}0.06792 & 0.1420 $-$ & \cellcolor{g}0.03079 & 0.1549 $-$ \\
\texttt{irs} & 3.582E-5 & \cellcolor{g}1.785E-5 $+$ & \cellcolor{g}0.006181 & 0.01111 $-$ & 0.5249 & \cellcolor{g}0.5200 $\sim$ \\
\texttt{lnd} & 0.9798 & \cellcolor{g}0.9775 $\sim$ & 0.9969 & \cellcolor{g}0.9961 $\sim$ & \cellcolor{g}0.9332 & 0.9500 $-$ \\
\texttt{mam} & \cellcolor{g}0.7961 & 0.8007 $\sim$ & \cellcolor{g}0.9832 & 0.9845 $\sim$ & 0.9902 & \cellcolor{g}0.9900 $\sim$ \\
\texttt{pdy} & \cellcolor{g}0.9013 & 0.9035 $\sim$ & \cellcolor{g}0.7812 & 0.7861 $-$ & \cellcolor{g}0.5905 & 0.5924 $\sim$ \\
\texttt{pis} & \cellcolor{g}0.6408 & 0.6411 $\sim$ & 0.8007 & \cellcolor{g}0.7964 $\sim$ & 0.8232 & \cellcolor{g}0.8200 $\sim$ \\
\texttt{pha} & \cellcolor{g}0.06045 & 0.06271 $\sim$ & \cellcolor{g}0.01429 & 0.01767 $\sim$ & \cellcolor{g}0.009177 & 0.03297 $-$ \\
\texttt{pre} & \cellcolor{g}0.9790 & 0.9805 $-$ & \cellcolor{g}0.9559 & 0.9566 $\sim$ & 0.9984 & \cellcolor{g}0.9983 $\sim$ \\
\texttt{pmp} & \cellcolor{g}0.7902 & 0.7963 $\sim$ & \cellcolor{g}0.9206 & 0.9235 $\sim$ & \cellcolor{g}0.9626 & 0.9629 $\sim$ \\
\texttt{rsn} & 0.6606 & \cellcolor{g}0.6600 $\sim$ & 0.9004 & \cellcolor{g}0.8978 $\sim$ & \cellcolor{g}0.9807 & 0.9811 $\sim$ \\
\texttt{seg} & \cellcolor{g}0.1223 & 0.1236 $\sim$ & 0.05225 & \cellcolor{g}0.04753 $\sim$ & \cellcolor{g}0.01814 & 0.01998 $\sim$ \\
\texttt{sir} & \cellcolor{g}0.003658 & 0.007363 $\sim$ & \cellcolor{g}0.001045 & 0.002416 $\sim$ & 0.4744 & \cellcolor{g}0.4567 $\sim$ \\
\texttt{smk} & 0.9926 & \cellcolor{g}0.9919 $\sim$ & \cellcolor{g}0.9971 & \cellcolor{g}0.9971 $\sim$ & \cellcolor{g}0.9982 & 0.9987 $\sim$ \\
\texttt{tae} & \cellcolor{g}0.04667 & 0.05208 $\sim$ & 0.4225 & \cellcolor{g}0.4062 $\sim$ & 0.8256 & \cellcolor{g}0.8021 $\sim$ \\
\texttt{tip} & 0.8788 & \cellcolor{g}0.8628 $+$ & 0.8550 & \cellcolor{g}0.8509 $\sim$ & 0.5107 & \cellcolor{g}0.5104 $\sim$ \\
\texttt{tit} & 0.9784 & \cellcolor{g}0.9767 $\sim$ & 0.9994 & \cellcolor{g}0.9978 $\sim$ & 0.9967 & \cellcolor{g}0.9948 $\sim$ \\
\texttt{wne} & \cellcolor{g}0.02619 & 0.03305 $\sim$ & \cellcolor{g}0.007416 & 0.01272 $\sim$ & \cellcolor{g}0.03417 & 0.04105 $\sim$ \\
\texttt{wbc} & 0.01929 & \cellcolor{g}0.01897 $\sim$ & 0.08151 & \cellcolor{g}0.07905 $\sim$ & 0.07059 & \cellcolor{g}0.07000 $\sim$ \\
\texttt{wpb} & \cellcolor{g}0.3321 & 0.4033 $-$ & \cellcolor{g}0.07253 & 0.2628 $-$ & \cellcolor{g}0.04257 & 0.2831 $-$ \\
\texttt{yst} & 0.9981 & \cellcolor{g}0.9978 $\sim$ & \cellcolor{g}0.4002 & 0.4096 $\sim$ & \cellcolor{g}0.1602 & 0.1636 $\sim$ \\
\bhline{1pt}
Rank & \cellcolor{g}\textit{1.33} & \textit{1.67}$\downarrow^{\dag}$ & \cellcolor{g}\textit{1.42} & \textit{1.58}$\downarrow$ & \cellcolor{g}\textit{1.47} & \textit{1.53}$\downarrow$ \\
Position & \textit{1} & \textit{2} & \textit{1} & \textit{2} & \textit{1} & \textit{2} \\
$+/-/\sim$ & - & 2/4/24 & - & 0/6/24 & - & 0/7/23 \\
\bhline{1pt}
$p$-value & - & 0.0285 & - & 0.137 & - & 0.245 \\

\bhline{1pt}

\end{tabular}
}
\end{center}
\end{table*}

Table \ref{tb: idk} presents the average ``I don't know'' belief mass, $m_\mathcal{D}(\mathbf{\Theta})$, at the 5th, 10th, and 50th epochs, where:
\begin{equation}
    m_\mathcal{D}(\mathbf{\Theta}) = \dfrac{\sum_{\mathbf{x}\in\mathcal{D}}m_{[M]}(\mathbf{\Theta}; \mathbf{x})}{|\mathcal{D}|},
\end{equation}
and $\mathcal{D}\in\{\mathcal{D}_\text{tr},\mathcal{D}_\text{te}\}$ represents a training or test dataset. Green-shaded values denote the smallest (best) values between the ``I don't know'' belief mass from a training dataset and that from a test dataset. 
{These values represent the degree of uncertainty in classification, ranging from 0 (complete certainty) to 1 (complete uncertainty), rather than the proportion of “I don’t know” predictions. For example, a value of 0.8017 for the \texttt{bnk} dataset at the 5th epoch during training means that, on average, 80.17\% of the total belief mass was assigned to the “I don’t know” state when classifying data points in this training dataset, indicating high uncertainty in \ds’s predictions. Note that even with high “I don’t know” belief mass, \ds\ still makes class predictions based on the remaining belief mass distributed among specific classes through the pignistic transform (cf. Section \ref{ss: pignistic transform}).}\label{r2-4}
Statistical results from the Wilcoxon signed-rank test are summarized with symbols wherein ``$+$'', ``$-$'', and ``$\sim$'' represent that the ``I don't know'' belief mass from a test dataset is significantly smaller, larger, or competitive compared to that from a training dataset, respectively. The ``$p$-value'' is derived from the Wilcoxon signed-rank test. Arrows denote whether the rank improved or declined compared to the ``I don't know'' belief mass from a training dataset. $\dag$ indicates statistically significant differences compared to the ``I don't know'' belief mass from a training dataset, i.e., $p$-value $<\alpha=0.05$.

From Table \ref{tb: idk}, it is evident that at the 5th, 10th, and 50th epochs, the average rank during testing is lower than during training, meaning that Fuzzy-UCS$_\text{DS}$ tends to output a larger ``I don't know'' belief mass during testing than during training. Particularly at the 5th epoch, there was a statistically significant difference in the ``I don't know'' belief mass between training and testing ($p<0.05$). At the 10th and 50th epochs, as expected, there were no datasets where the ``I don't know'' belief mass during testing was significantly smaller than during training (denoted by the ``$+$'' symbol). An interesting insight is that the seven datasets showing a significantly larger ``I don't know'' belief mass during testing at the 50th epoch (i.e., \texttt{can}, \texttt{hrt}, \texttt{hpt}, \texttt{hcl}, \texttt{lnd}, \texttt{pha}, and \texttt{wpb}, denoted by the ``$-$'' symbol) were all characterized by high uncertainty, i.e., either having very small instance counts (676 or fewer) and/or containing missing values. These results, along with analyses from Section \ref{ss: visualization belief mass}, conclude that the ``I don't know'' belief mass output by Fuzzy-UCS$_\text{DS}$ adequately quantifies classification uncertainty.}

\section{Concluding Remarks}
\label{sec: concluding remarks}
This article introduced a novel class inference scheme for LFCSs, using the DS theory to enhance LFCS performance, especially in uncertain real-world problem domains. Our scheme transforms the membership degrees and weight vectors of multiple fuzzy rules into belief mass functions. These belief mass functions are then integrated using Dempster's rule of combination and the pignistic transform during the class inference process.
The application of the proposed inference scheme within the Fuzzy-UCS framework yielded significant improvement in classification performance on real-world problems. 
A key advantage of our scheme is its ability to provide both inference results and a measure of confidence for each inference result. This offers valuable insights into the certainty of classification.

{Future research directions include:}
\begin{itemize}
    \item Applying our DS theory-based class inference scheme to recently studied XCSR \cite{wagner2022mechanisms} and ACS2 \cite{smierzchala2023anticipatory} classifier systems for reinforcement learning tasks with discrete action labels (e.g., an action label set is \{up, down, left, right\}). 

    \item {Extending our scheme to various evolutionary learning algorithms that generate sets of solutions, each suggesting different classifications for a given input (e.g., multi-tree Genetic Programming \cite{qinglan2024multi}).}

    \item Extending our scheme to LFCSs which can deal with multi-label classification problems (e.g., \cite{omozaki2022evolutionary}) by developing a methodology that accounts for belief mass assignment to meta-classes.

    \item {Exploring the combined application of our scheme and \textit{absumption} techniques \cite{shiraishi2022absumption,liu2023absumption,liu2024phenotypic}. Investigating whether these two techniques complement each other or exhibit redundancy will provide deeper insights into optimizing LFCS architectures for enhanced generalization and robustness against overfitting. }

    \item {Addressing limitations of the use of predefined fuzzy sets (e.g., our homogeneous linguistic discretization of domain intervals), such as the potential imposition of assumptions about data distribution \cite{antonelli2009multi} and the loss of specificity in some domains where precision is essential \cite{orriols2008fuzzy}. This could involve integrating adaptive discretization methods \cite{murata1995adjusting} that adjust linguistic term partitions to feature distributions and/or mixed discrete-continuous attribute list knowledge representation \cite{urbanowicz2015exstracs} that handles both discrete and continuous attributes effectively.}

    \item {Investigating the application of the DS theory in Pittsburgh-style LFCSs, which often result in more compact rule sets and may enhance interpretability and efficiency. Comparisons with existing work (e.g., \cite{bishop2021genetic}) could provide valuable insights.}

    \item {Developing benchmark datasets to evaluate LFCS capabilities by incorporating characteristics such as inherent linguistic uncertainty, interpretability-focused problems, and varying degrees of feature overlap.}\label{r2-2}
\end{itemize}

\begin{acks}
This work was supported by the Japan Society for the Promotion of Science KAKENHI (Grant Nos. JP23KJ0993, JP23K20388), the National Natural Science Foundation of China (Grant No. 62376115), and the Guangdong Provincial Key Laboratory (Grant No. 2020B121201001). 
\end{acks}


\bibliographystyle{ACM-Reference-Format}
\bibliography{telo}

\appendix
\clearpage
\section{Analysis of Fitness Range Impact}
\label{sec: sup analysis of fitness range impact}
{A notable feature of Fuzzy-UCS is the use of a fitness range of (-1, 1], which differs from the traditional (0, 1] range used in many classifier systems. This extended range was originally introduced by Ishibuchi and Yamamoto \shortcite{ishibuchi2005rule}, who demonstrated its effectiveness in improving the performance of LFCSs. Following their findings, the original Fuzzy-UCS paper \cite{orriols2008fuzzy} adopted this extended range as the default setting.

To validate the effectiveness of the (-1, 1] fitness range in our Fuzzy-UCS variants, we conduct comparative experiments, where the same problems and experimental settings as in Section \ref{ss: experimental setup} are used. The number of training iterations is set to 50 epochs. Here, fitness values within the (0, 1] range have been calculated according to Ishibuchi and Yamamoto’s original formulation: $F_{t+1}^k=\max\left(\mathbf{v}_{t+1}^k\right)$.}

\begin{table*}[b]
\begin{center}
\caption{Results When Changing Fitness Ranges, Displaying Average Classification Accuracy Across 30 Runs.}
\label{tb: sup compare fitness range}

\scalebox{0.7}{
\begin{tabular}{c|cc|cc|cc|cc|cc|cc}
\bhline{1pt}
\multicolumn{1}{c|}{\multirow{2}{*}{}} &\multicolumn{6}{c|}{\textsc{Training Accuracy (\%)}} & \multicolumn{6}{c}{\textsc{Test Accuracy (\%)}}\\
& \multicolumn{2}{c|}{Fuzzy-UCS$_\text{VOTE}$} & \multicolumn{2}{c|}{Fuzzy-UCS$_\text{SWIN}$} & \multicolumn{2}{c|}{Fuzzy-UCS$_\text{DS}$} & \multicolumn{2}{c|}{Fuzzy-UCS$_\text{VOTE}$} & \multicolumn{2}{c|}{Fuzzy-UCS$_\text{SWIN}$} & \multicolumn{2}{c}{Fuzzy-UCS$_\text{DS}$} \\
$F^k\in$&$(-1,1]$ & $(0,1]$ & $(-1,1]$ & $(0,1]$ & $(-1,1]$ & $(0,1]$ & $(-1,1]$ & $(0,1]$ & $(-1,1]$ & $(0,1]$ & $(-1,1]$ & $(0,1]$ \\
\bhline{1pt}
\texttt{bnk} & \cellcolor{g}92.63 & 91.86 $-$ & \cellcolor{g}93.65 & 91.92 $-$ & \cellcolor{g}92.95 & 91.94 $-$ & \cellcolor{g}92.34 & 91.40 $-$ & \cellcolor{g}93.00 & 91.86 $-$ & \cellcolor{g}92.63 & 91.43 $-$ \\
\texttt{can} & \cellcolor{g}95.29 & \cellcolor{g}95.29 $\sim$ & 94.83 & \cellcolor{g}94.94 $\sim$ & \cellcolor{g}95.39 & 95.12 $-$ & \cellcolor{g}94.74 & 94.44 $\sim$ & \cellcolor{g}94.44 & 94.09 $\sim$ & \cellcolor{g}94.80 & 94.68 $\sim$ \\
\texttt{car} & \cellcolor{g}87.93 & 85.04 $-$ & 88.07 & \cellcolor{g}88.37 $\sim$ & 88.22 & \cellcolor{g}88.44 $\sim$ & \cellcolor{g}76.00 & 74.67 $\sim$ & \cellcolor{g}76.67 & 75.33 $\sim$ & 74.00 & \cellcolor{g}77.33 $\sim$ \\
\texttt{col} & \cellcolor{g}71.85 & 68.58 $-$ & \cellcolor{g}68.24 & 65.90 $-$ & \cellcolor{g}71.37 & 67.31 $-$ & \cellcolor{g}69.35 & 66.77 $\sim$ & \cellcolor{g}65.81 & 64.84 $\sim$ & \cellcolor{g}69.03 & 65.59 $-$ \\
\texttt{dbt} & 76.36 & \cellcolor{g}77.07 $+$ & \cellcolor{g}73.57 & 70.62 $-$ & \cellcolor{g}76.48 & 75.05 $-$ & 73.07 & \cellcolor{g}74.29 $+$ & \cellcolor{g}70.04 & 66.97 $-$ & \cellcolor{g}72.94 & 72.21 $\sim$ \\
\texttt{ecl} & 85.79 & \cellcolor{g}87.34 $+$ & \cellcolor{g}80.39 & 78.32 $-$ & \cellcolor{g}85.64 & 83.63 $-$ & 84.02 & \cellcolor{g}86.08 $+$ & \cellcolor{g}79.80 & 77.94 $\sim$ & \cellcolor{g}83.92 & 83.43 $\sim$ \\
\texttt{frt} & 87.12 & \cellcolor{g}88.14 $+$ & 85.17 & \cellcolor{g}85.80 $+$ & \cellcolor{g}88.43 & 88.26 $\sim$ & 84.74 & \cellcolor{g}86.30 $+$ & 82.33 & \cellcolor{g}83.26 $\sim$ & \cellcolor{g}86.44 & 86.11 $\sim$ \\
\texttt{gls} & \cellcolor{g}72.01 & 68.63 $-$ & \cellcolor{g}71.75 & 69.25 $-$ & \cellcolor{g}72.14 & 66.55 $-$ & \cellcolor{g}62.73 & 59.09 $\sim$ & \cellcolor{g}64.09 & 61.67 $\sim$ & \cellcolor{g}64.39 & 55.45 $-$ \\
\texttt{hrt} & \cellcolor{g}94.66 & 93.27 $-$ & \cellcolor{g}94.35 & 93.97 $\sim$ & \cellcolor{g}95.05 & 94.08 $-$ & 81.83 & \cellcolor{g}82.69 $\sim$ & \cellcolor{g}82.69 & \cellcolor{g}82.69 $\sim$ & 81.83 & \cellcolor{g}82.58 $\sim$ \\
\texttt{hpt} & \cellcolor{g}91.46 & 90.22 $-$ & \cellcolor{g}89.88 & 88.75 $-$ & \cellcolor{g}92.33 & 90.86 $-$ & \cellcolor{g}64.58 & 63.33 $\sim$ & 62.08 & \cellcolor{g}63.75 $\sim$ & \cellcolor{g}65.62 & 63.96 $\sim$ \\
\texttt{hcl} & 87.25 & \cellcolor{g}87.69 $\sim$ & \cellcolor{g}84.19 & 83.72 $\sim$ & \cellcolor{g}87.76 & 86.58 $-$ & \cellcolor{g}71.80 & 70.00 $\sim$ & 69.37 & \cellcolor{g}71.17 $\sim$ & \cellcolor{g}71.62 & 68.83 $-$ \\
\texttt{irs} & \cellcolor{g}94.81 & 94.02 $\sim$ & 95.51 & \cellcolor{g}95.65 $\sim$ & \cellcolor{g}95.26 & 94.84 $\sim$ & \cellcolor{g}94.00 & 93.11 $\sim$ & \cellcolor{g}94.89 & 94.44 $\sim$ & \cellcolor{g}94.22 & 92.89 $\sim$ \\
\texttt{lnd} & 32.05 & \cellcolor{g}45.01 $+$ & 39.18 & \cellcolor{g}42.03 $+$ & 46.27 & \cellcolor{g}49.41 $+$ & \cellcolor{g}38.24 & 32.35 $-$ & 40.10 & \cellcolor{g}40.59 $\sim$ & \cellcolor{g}40.20 & 34.12 $-$ \\
\texttt{mam} & \cellcolor{g}80.89 & 80.61 $-$ & \cellcolor{g}77.89 & 77.87 $\sim$ & \cellcolor{g}81.07 & 80.94 $\sim$ & 79.73 & \cellcolor{g}80.10 $\sim$ & \cellcolor{g}76.77 & 76.63 $\sim$ & \cellcolor{g}80.03 & 79.90 $\sim$ \\
\texttt{pdy} & 50.94 & \cellcolor{g}68.48 $+$ & 50.70 & \cellcolor{g}51.93 $+$ & 53.94 & \cellcolor{g}72.40 $+$ & 51.46 & \cellcolor{g}68.56 $+$ & 51.36 & \cellcolor{g}52.54 $+$ & 54.63 & \cellcolor{g}71.83 $+$ \\
\texttt{pis} & \cellcolor{g}86.89 & 86.64 $-$ & \cellcolor{g}85.53 & 84.54 $-$ & \cellcolor{g}86.91 & 86.73 $-$ & 85.88 & \cellcolor{g}86.08 $\sim$ & \cellcolor{g}84.67 & 83.29 $-$ & \cellcolor{g}85.92 & 85.84 $\sim$ \\
\texttt{pha} & \cellcolor{g}82.10 & 77.95 $-$ & \cellcolor{g}86.80 & 83.95 $-$ & \cellcolor{g}85.64 & 83.22 $-$ & 66.96 & \cellcolor{g}67.45 $\sim$ & \cellcolor{g}64.41 & 63.73 $\sim$ & 66.13 & \cellcolor{g}67.01 $\sim$ \\
\texttt{pre} & 98.01 & \cellcolor{g}98.69 $+$ & \cellcolor{g}100.0 & \cellcolor{g}100.0 $\sim$ & 99.76 & \cellcolor{g}99.98 $+$ & 98.03 & \cellcolor{g}98.87 $+$ & \cellcolor{g}100.0 & \cellcolor{g}100.0 $\sim$ & 99.79 & \cellcolor{g}100.0 $+$ \\
\texttt{pmp} & \cellcolor{g}86.99 & 86.97 $\sim$ & \cellcolor{g}86.89 & 86.10 $-$ & \cellcolor{g}87.03 & 86.93 $-$ & \cellcolor{g}86.17 & 85.97 $\sim$ & \cellcolor{g}86.05 & 85.40 $-$ & \cellcolor{g}86.21 & 85.95 $\sim$ \\
\texttt{rsn} & 84.33 & \cellcolor{g}84.43 $\sim$ & 85.51 & \cellcolor{g}85.56 $\sim$ & \cellcolor{g}85.13 & 84.79 $\sim$ & 84.19 & \cellcolor{g}84.44 $\sim$ & \cellcolor{g}85.30 & \cellcolor{g}85.30 $\sim$ & \cellcolor{g}85.07 & 84.81 $\sim$ \\
\texttt{seg} & \cellcolor{g}90.10 & 89.45 $-$ & 90.52 & \cellcolor{g}91.01 $+$ & \cellcolor{g}90.42 & 90.19 $\sim$ & \cellcolor{g}89.88 & 89.25 $-$ & 90.01 & \cellcolor{g}90.85 $+$ & \cellcolor{g}90.30 & 89.73 $\sim$ \\
\texttt{sir} & \cellcolor{g}84.78 & 83.52 $-$ & \cellcolor{g}85.00 & 82.93 $-$ & \cellcolor{g}84.78 & 83.89 $-$ & \cellcolor{g}81.67 & 79.00 $\sim$ & \cellcolor{g}81.33 & 80.00 $\sim$ & \cellcolor{g}82.00 & 81.00 $\sim$ \\
\texttt{smk} & \cellcolor{g}95.39 & 93.15 $-$ & \cellcolor{g}93.49 & 62.81 $-$ & \cellcolor{g}95.71 & 86.73 $-$ & \cellcolor{g}95.08 & 93.11 $-$ & \cellcolor{g}93.14 & 63.01 $-$ & \cellcolor{g}95.11 & 86.76 $-$ \\
\texttt{tae} & \cellcolor{g}63.63 & 59.70 $-$ & \cellcolor{g}64.15 & 62.77 $\sim$ & \cellcolor{g}64.35 & 60.59 $-$ & 54.58 & \cellcolor{g}55.00 $\sim$ & 53.75 & \cellcolor{g}55.00 $\sim$ & \cellcolor{g}56.46 & 55.42 $\sim$ \\
\texttt{tip} & \cellcolor{g}81.09 & 80.77 $-$ & \cellcolor{g}80.24 & 79.33 $-$ & \cellcolor{g}81.20 & 80.87 $-$ & \cellcolor{g}80.00 & 79.87 $\sim$ & \cellcolor{g}78.88 & 78.21 $-$ & \cellcolor{g}80.23 & 80.00 $\sim$ \\
\texttt{tit} & \cellcolor{g}72.28 & 71.64 $-$ & \cellcolor{g}70.57 & 69.21 $-$ & \cellcolor{g}72.38 & 71.86 $-$ & \cellcolor{g}71.30 & 70.85 $\sim$ & \cellcolor{g}68.93 & 67.96 $\sim$ & \cellcolor{g}71.52 & 71.07 $\sim$ \\
\texttt{wne} & \cellcolor{g}98.31 & 97.65 $-$ & \cellcolor{g}98.58 & 98.23 $\sim$ & \cellcolor{g}98.79 & 98.00 $-$ & 96.48 & \cellcolor{g}96.67 $\sim$ & \cellcolor{g}96.11 & 94.63 $\sim$ & 96.67 & \cellcolor{g}97.41 $\sim$ \\
\texttt{wbc} & 97.12 & \cellcolor{g}97.23 $\sim$ & 96.42 & \cellcolor{g}96.50 $\sim$ & 97.07 & \cellcolor{g}97.15 $\sim$ & 96.19 & \cellcolor{g}96.38 $\sim$ & \cellcolor{g}95.29 & 95.10 $\sim$ & 95.95 & \cellcolor{g}96.14 $\sim$ \\
\texttt{wpb} & \cellcolor{g}91.57 & 91.46 $\sim$ & 93.60 & \cellcolor{g}93.73 $\sim$ & \cellcolor{g}92.58 & 91.93 $-$ & 72.17 & \cellcolor{g}74.33 $\sim$ & 71.67 & \cellcolor{g}71.83 $\sim$ & 72.00 & \cellcolor{g}73.50 $\sim$ \\
\texttt{yst} & \cellcolor{g}53.26 & 49.62 $-$ & \cellcolor{g}56.22 & 49.18 $-$ & \cellcolor{g}58.78 & 45.86 $-$ & \cellcolor{g}52.04 & 49.40 $\sim$ & \cellcolor{g}53.65 & 48.43 $-$ & \cellcolor{g}57.61 & 45.86 $-$ \\
\bhline{1pt}
Rank & \cellcolor{g}\textit{1.32} & \textit{1.68}$\downarrow$ & \cellcolor{g}\textit{1.35} & \textit{1.65}$\downarrow^{\dag}$ & \cellcolor{g}\textit{1.17} & \textit{1.83}$\downarrow^{\dag}$ & \cellcolor{g}\textit{1.47} & \textit{1.53}$\downarrow$ & \cellcolor{g}\textit{1.32} & \textit{1.68}$\downarrow^{\dag}$ & \cellcolor{g}\textit{1.27} & \textit{1.73}$\downarrow^{\dag}$ \\
Position & \textit{1} & \textit{2} & \textit{1} & \textit{2} & \textit{1} & \textit{2} & \textit{1} & \textit{2} & \textit{1} & \textit{2} & \textit{1} & \textit{2} \\
$+/-/\sim$ & - & 6/17/7 & - & 4/14/12 & - & 3/20/7 & - & 5/4/21 & - & 2/7/21 & - & 2/7/21 \\
\bhline{1pt}
$p$-value & - & 0.0645 & - & 0.00599 & - & 0.000743 & - & 0.329 & - & 0.0385 & - & 0.0341 \\

\bhline{1pt}

\end{tabular}
}
\end{center}
\end{table*}

{Table \ref{tb: sup compare fitness range} presents the training and test classification accuracy for each variant of Fuzzy-UCS$_*$ at the 50th epoch. Green-shaded values denote the best values between Fuzzy-UCS$_*$ with a fitness range of (-1, 1] and Fuzzy-UCS$_*$ with a fitness range of (0, 1]. Statistical results of the Wilcoxon signed-rank test are summarized with symbols wherein ``$+$'', ``$-$'', and ``$\sim$'' represent that the classification accuracy of Fuzzy-UCS$_*$ with a fitness range of (0, 1] is significantly better, worse, and competitive compared to that obtained by Fuzzy-UCS$_*$ with a fitness range of (-1, 1], respectively. The ``$p$-value'' is derived from the Wilcoxon signed-rank test. Arrows denote whether the rank improved or declined compared to Fuzzy-UCS$_*$ with a fitness range of (-1, 1]. $\dag$ indicates statistically significant differences compared to Fuzzy-UCS$_*$ with a fitness range of (-1, 1], i.e., $p$-value $<\alpha=0.05$. 

The experimental results show that the use of (-1, 1] fitness range improved the test accuracy of \all\ with the (0, 1] fitness range on many datasets, as evidenced by lower average ranks in the table. Negative fitness values allow the system to assign penalties to rules that have a high probability of misclassifying instances. In LFCSs including Fuzzy-UCS, rules with negative fitness values are excluded from parent selection in the GA (refer to Section \ref{sss: training phase Fuzzy-UCS}). This mechanism enables the GA to guide the system in systematically eliminating inaccurate rules, thereby enhancing the performance of the classifier.}

\clearpage

\section{Analysis of Experience Update Methods}
\label{sec: sup analysis of experience update methods}
{In traditional LCSs such as UCS, the experience of a rule is typically updated incrementally as the match set is formed, using the formula ${\rm exp}_{t+1}^k = {\rm exp}_{t}^k + 1$. This approach uniformly increases the experience of all matched rules by one at each time step, regardless of their individual contributions to the classification process. In contrast, Fuzzy-UCS introduces a membership degree-based experience update method: ${\rm exp}_{t+1}^k = {\rm exp}_{t}^k + \mu_{\mathbf{A}^k}(\mathbf{x})$. This approach updates experience based on membership degrees, allowing rules that are more relevant to the input data to receive greater experience increments. This differentiation enables the GA to prioritize and evolve rules that have higher contributions to accurate classification.

To examine the impact of these different methods of updating experience on the classification performance of Fuzzy-UCS, we conduct comparative experiments, where the same problems and experimental settings as in Section \ref{ss: experimental setup} are used. The number of training iterations is set to 50 epochs.}

  \begin{table*}[b]
\begin{center}
\caption{Results When Changing Experience Update Method, Displaying Average Classification Accuracy Across 30 Runs.}
\label{tb: sup changing experience update method}

\scalebox{0.7}{
\begin{tabular}{c|cc|cc|cc|cc|cc|cc}
\bhline{1pt}
\multicolumn{1}{c|}{\multirow{2}{*}{}} &\multicolumn{6}{c|}{\textsc{Training Accuracy (\%)}} & \multicolumn{6}{c}{\textsc{Test Accuracy (\%)}}\\
& \multicolumn{2}{c|}{Fuzzy-UCS$_\text{VOTE}$} & \multicolumn{2}{c|}{Fuzzy-UCS$_\text{SWIN}$} & \multicolumn{2}{c|}{Fuzzy-UCS$_\text{DS}$} & \multicolumn{2}{c|}{Fuzzy-UCS$_\text{VOTE}$} & \multicolumn{2}{c|}{Fuzzy-UCS$_\text{SWIN}$} & \multicolumn{2}{c}{Fuzzy-UCS$_\text{DS}$} \\
$\text{exp}_{t+1}^k=\text{exp}_{t}^k+$&$\mu_{\mathbf{A}^k}(\mathbf{x})$ & 1 & $\mu_{\mathbf{A}^k}(\mathbf{x})$ & 1 & $\mu_{\mathbf{A}^k}(\mathbf{x})$ & 1 & $\mu_{\mathbf{A}^k}(\mathbf{x})$ & 1 & $\mu_{\mathbf{A}^k}(\mathbf{x})$ & 1 & $\mu_{\mathbf{A}^k}(\mathbf{x})$ & 1 \\
\bhline{1pt}
\texttt{bnk} & \cellcolor{g}92.63 & 55.53 $-$ & \cellcolor{g}93.65 & 52.96 $-$ & \cellcolor{g}92.95 & 55.53 $-$ & \cellcolor{g}92.34 & 55.63 $-$ & \cellcolor{g}93.00 & 52.87 $-$ & \cellcolor{g}92.63 & 55.63 $-$ \\
\texttt{can} & \cellcolor{g}95.29 & 62.74 $-$ & \cellcolor{g}94.83 & 61.92 $-$ & \cellcolor{g}95.39 & 62.74 $-$ & \cellcolor{g}94.74 & 62.75 $-$ & \cellcolor{g}94.44 & 61.64 $-$ & \cellcolor{g}94.80 & 62.75 $-$ \\
\texttt{car} & \cellcolor{g}87.93 & 68.22 $-$ & \cellcolor{g}88.07 & 68.22 $-$ & \cellcolor{g}88.22 & 68.22 $-$ & \cellcolor{g}76.00 & 66.00 $-$ & \cellcolor{g}76.67 & 66.00 $-$ & \cellcolor{g}74.00 & 66.00 $-$ \\
\texttt{col} & \cellcolor{g}71.85 & 23.94 $-$ & \cellcolor{g}68.24 & 47.84 $-$ & \cellcolor{g}71.37 & 48.40 $-$ & \cellcolor{g}69.35 & 23.23 $-$ & \cellcolor{g}65.81 & 47.96 $-$ & \cellcolor{g}69.03 & 48.28 $-$ \\
\texttt{dbt} & \cellcolor{g}76.36 & 65.24 $-$ & \cellcolor{g}73.57 & 65.24 $-$ & \cellcolor{g}76.48 & 65.24 $-$ & \cellcolor{g}73.07 & 63.90 $-$ & \cellcolor{g}70.04 & 63.90 $-$ & \cellcolor{g}72.94 & 63.90 $-$ \\
\texttt{ecl} & \cellcolor{g}85.79 & 20.52 $-$ & \cellcolor{g}80.39 & 41.61 $-$ & \cellcolor{g}85.64 & 42.24 $-$ & \cellcolor{g}84.02 & 20.59 $-$ & \cellcolor{g}79.80 & 44.51 $-$ & \cellcolor{g}83.92 & 45.39 $-$ \\
\texttt{frt} & \cellcolor{g}87.12 & 7.550 $-$ & \cellcolor{g}85.17 & 20.59 $-$ & \cellcolor{g}88.43 & 22.35 $-$ & \cellcolor{g}84.74 & 7.148 $-$ & \cellcolor{g}82.33 & 22.22 $-$ & \cellcolor{g}86.44 & 19.78 $-$ \\
\texttt{gls} & \cellcolor{g}72.01 & 7.726 $-$ & \cellcolor{g}71.75 & 34.51 $-$ & \cellcolor{g}72.14 & 35.75 $-$ & \cellcolor{g}62.73 & 9.848 $-$ & \cellcolor{g}64.09 & 32.42 $-$ & \cellcolor{g}64.39 & 31.67 $-$ \\
\texttt{hrt} & \cellcolor{g}94.66 & 54.52 $-$ & \cellcolor{g}94.35 & 51.73 $-$ & \cellcolor{g}95.05 & 54.52 $-$ & \cellcolor{g}81.83 & 53.87 $-$ & \cellcolor{g}82.69 & 49.35 $-$ & \cellcolor{g}81.83 & 53.87 $-$ \\
\texttt{hpt} & \cellcolor{g}91.46 & 54.87 $-$ & \cellcolor{g}89.88 & 52.33 $-$ & \cellcolor{g}92.33 & 54.87 $-$ & \cellcolor{g}64.58 & 54.58 $-$ & \cellcolor{g}62.08 & 51.67 $-$ & \cellcolor{g}65.62 & 54.58 $-$ \\
\texttt{hcl} & \cellcolor{g}87.25 & 66.07 $-$ & \cellcolor{g}84.19 & 65.18 $-$ & \cellcolor{g}87.76 & 66.07 $-$ & \cellcolor{g}71.80 & 68.38 $-$ & \cellcolor{g}69.37 & 65.59 $\sim$ & \cellcolor{g}71.62 & 68.38 $-$ \\
\texttt{irs} & \cellcolor{g}94.81 & 32.15 $-$ & \cellcolor{g}95.51 & 33.70 $-$ & \cellcolor{g}95.26 & 34.57 $-$ & \cellcolor{g}94.00 & 44.00 $-$ & \cellcolor{g}94.89 & 30.00 $-$ & \cellcolor{g}94.22 & 22.22 $-$ \\
\texttt{lnd} & \cellcolor{g}32.05 & 19.41 $-$ & \cellcolor{g}39.18 & 19.85 $-$ & \cellcolor{g}46.27 & 21.17 $-$ & \cellcolor{g}38.24 & 21.37 $-$ & \cellcolor{g}40.10 & 19.61 $-$ & \cellcolor{g}40.20 & 16.47 $-$ \\
\texttt{mam} & \cellcolor{g}80.89 & 53.73 $-$ & \cellcolor{g}77.89 & 51.13 $-$ & \cellcolor{g}81.07 & 53.73 $-$ & \cellcolor{g}79.73 & 53.33 $-$ & \cellcolor{g}76.77 & 49.73 $-$ & \cellcolor{g}80.03 & 53.33 $-$ \\
\texttt{pdy} & \cellcolor{g}50.94 & 24.83 $-$ & \cellcolor{g}50.70 & 24.94 $-$ & \cellcolor{g}53.94 & 25.17 $-$ & \cellcolor{g}51.46 & 26.51 $-$ & \cellcolor{g}51.36 & 25.54 $-$ & \cellcolor{g}54.63 & 23.44 $-$ \\
\texttt{pis} & \cellcolor{g}86.89 & 57.39 $-$ & \cellcolor{g}85.53 & 55.44 $-$ & \cellcolor{g}86.91 & 57.39 $-$ & \cellcolor{g}85.88 & 57.07 $-$ & \cellcolor{g}84.67 & 54.96 $-$ & \cellcolor{g}85.92 & 57.07 $-$ \\
\texttt{pha} & \cellcolor{g}82.10 & 55.09 $-$ & \cellcolor{g}86.80 & 51.60 $-$ & \cellcolor{g}85.64 & 55.09 $-$ & \cellcolor{g}66.96 & 53.04 $-$ & \cellcolor{g}64.41 & 51.86 $-$ & \cellcolor{g}66.13 & 53.04 $-$ \\
\texttt{pre} & \cellcolor{g}98.01 & 33.22 $-$ & \cellcolor{g}100.0 & 33.22 $-$ & \cellcolor{g}99.76 & 33.49 $-$ & \cellcolor{g}98.03 & 34.37 $-$ & \cellcolor{g}100.0 & 34.36 $-$ & \cellcolor{g}99.79 & 31.88 $-$ \\
\texttt{pmp} & \cellcolor{g}86.99 & 51.91 $-$ & \cellcolor{g}86.89 & 50.08 $-$ & \cellcolor{g}87.03 & 51.91 $-$ & \cellcolor{g}86.17 & 52.80 $-$ & \cellcolor{g}86.05 & 50.59 $-$ & \cellcolor{g}86.21 & 52.80 $-$ \\
\texttt{rsn} & \cellcolor{g}84.33 & 50.54 $-$ & \cellcolor{g}85.51 & 49.86 $-$ & \cellcolor{g}85.13 & 50.54 $-$ & \cellcolor{g}84.19 & 45.15 $-$ & \cellcolor{g}85.30 & 51.30 $-$ & \cellcolor{g}85.07 & 45.15 $-$ \\
\texttt{seg} & \cellcolor{g}90.10 & 14.10 $-$ & \cellcolor{g}90.52 & 14.23 $-$ & \cellcolor{g}90.42 & 14.49 $-$ & \cellcolor{g}89.88 & 15.99 $-$ & \cellcolor{g}90.01 & 14.76 $-$ & \cellcolor{g}90.30 & 12.42 $-$ \\
\texttt{sir} & \cellcolor{g}84.78 & 51.56 $-$ & \cellcolor{g}85.00 & 50.81 $-$ & \cellcolor{g}84.78 & 51.56 $-$ & \cellcolor{g}81.67 & 36.00 $-$ & \cellcolor{g}81.33 & 42.67 $-$ & \cellcolor{g}82.00 & 36.00 $-$ \\
\texttt{smk} & \cellcolor{g}95.39 & 60.46 $-$ & \cellcolor{g}93.49 & 58.30 $-$ & \cellcolor{g}95.71 & 60.46 $-$ & \cellcolor{g}95.08 & 60.91 $-$ & \cellcolor{g}93.14 & 59.32 $-$ & \cellcolor{g}95.11 & 60.91 $-$ \\
\texttt{tae} & \cellcolor{g}63.63 & 31.98 $-$ & \cellcolor{g}64.15 & 33.04 $-$ & \cellcolor{g}64.35 & 35.04 $-$ & \cellcolor{g}54.58 & 42.92 $-$ & \cellcolor{g}53.75 & 33.12 $-$ & \cellcolor{g}56.46 & 25.00 $-$ \\
\texttt{tip} & \cellcolor{g}81.09 & 64.18 $-$ & \cellcolor{g}80.24 & 64.18 $-$ & \cellcolor{g}81.20 & 64.18 $-$ & \cellcolor{g}80.00 & 65.09 $-$ & \cellcolor{g}78.88 & 65.09 $-$ & \cellcolor{g}80.23 & 65.09 $-$ \\
\texttt{tit} & \cellcolor{g}72.28 & 61.77 $-$ & \cellcolor{g}70.57 & 61.77 $-$ & \cellcolor{g}72.38 & 61.77 $-$ & \cellcolor{g}71.30 & 60.22 $-$ & \cellcolor{g}68.93 & 60.22 $-$ & \cellcolor{g}71.52 & 60.22 $-$ \\
\texttt{wne} & \cellcolor{g}98.31 & 29.35 $-$ & \cellcolor{g}98.58 & 35.94 $-$ & \cellcolor{g}98.79 & 40.08 $-$ & \cellcolor{g}96.48 & 32.22 $-$ & \cellcolor{g}96.11 & 31.67 $-$ & \cellcolor{g}96.67 & 38.15 $-$ \\
\texttt{wbc} & \cellcolor{g}97.12 & 65.57 $-$ & \cellcolor{g}96.42 & 65.57 $-$ & \cellcolor{g}97.07 & 65.57 $-$ & \cellcolor{g}96.19 & 65.05 $-$ & \cellcolor{g}95.29 & 65.05 $-$ & \cellcolor{g}95.95 & 65.05 $-$ \\
\texttt{wpb} & \cellcolor{g}91.57 & 76.29 $-$ & \cellcolor{g}93.60 & 76.29 $-$ & \cellcolor{g}92.58 & 76.29 $-$ & 72.17 & \cellcolor{g}76.00 $+$ & 71.67 & \cellcolor{g}76.00 $+$ & 72.00 & \cellcolor{g}76.00 $+$ \\
\texttt{yst} & \cellcolor{g}53.26 & 15.53 $-$ & \cellcolor{g}56.22 & 28.80 $-$ & \cellcolor{g}58.78 & 31.15 $-$ & \cellcolor{g}52.04 & 16.06 $-$ & \cellcolor{g}53.65 & 28.10 $-$ & \cellcolor{g}57.61 & 31.66 $-$ \\
\bhline{1pt}
Rank & \cellcolor{g}\textit{1.00} & \textit{2.00}$\downarrow^{\dag}$ & \cellcolor{g}\textit{1.00} & \textit{2.00}$\downarrow^{\dag}$ & \cellcolor{g}\textit{1.00} & \textit{2.00}$\downarrow^{\dag}$ & \cellcolor{g}\textit{1.03} & \textit{1.97}$\downarrow^{\dag}$ & \cellcolor{g}\textit{1.03} & \textit{1.97}$\downarrow^{\dag}$ & \cellcolor{g}\textit{1.03} & \textit{1.97}$\downarrow^{\dag}$ \\
Position & \textit{1} & \textit{2} & \textit{1} & \textit{2} & \textit{1} & \textit{2} & \textit{1} & \textit{2} & \textit{1} & \textit{2} & \textit{1} & \textit{2} \\
$+/-/\sim$ & - & 0/30/0 & - & 0/30/0 & - & 0/30/0 & - & 1/29/0 & - & 1/28/1 & - & 1/29/0 \\
\bhline{1pt}
$p$-value & - & 1.86E-09 & - & 1.86E-09 & - & 1.86E-09 & - & 2.24E-06 & - & 5.59E-09 & - & 5.59E-09 \\

\bhline{1pt}

\end{tabular}
}
\end{center}
\end{table*}

{Table \ref{tb: sup changing experience update method} presents the training and test classification accuracy for each variant of Fuzzy-UCS$_*$ at the 50th epoch. Green-shaded values denote the best values between Fuzzy-UCS$_*$ with the ${\rm exp}_{t+1}^k={\rm exp}_{t}^k+\mu_{\mathbf{A}^k}(\mathbf{x})$ update method (denoted as Fuzzy-UCS$_*$-Memb) and Fuzzy-UCS$_*$ with the ${\rm exp}_{t+1}^k={\rm exp}_{t}^k+1$ update method (denoted as Fuzzy-UCS$_*$-NoMemb). Statistical results of the Wilcoxon signed-rank test are summarized with symbols wherein ``$+$'', ``$-$'', and ``$\sim$'' represent that the classification accuracy of Fuzzy-UCS$_*$-NoMemb is significantly better, worse, and competitive compared to that obtained by Fuzzy-UCS$_*$-Memb, respectively. The ``$p$-value'' is derived from the Wilcoxon signed-rank test. Arrows denote whether the rank improved or declined compared to Fuzzy-UCS$_*$-Memb. $\dag$ indicates statistically significant differences compared to Fuzzy-UCS$_*$-Memb, i.e., $p$-value $<\alpha=0.05$. 

The experimental results indicate that Fuzzy-UCS$_*$-Memb achieved higher training and test accuracy compared to Fuzzy-UCS$_*$-NoMemb, as evidenced by lower average ranks in the table. The superior performance observed with the membership degree-based update method can be attributed to differentiated rule contribution in Fuzzy-UCS. }

\clearpage
\section{Experimental Results on Standard Accuracy}
\label{sec: sup Experimental Results on Standard Accuracy}
\label{r2-5-1}
\begin{table*}[b]
\begin{center}
\caption{Results When the Number of Training Iterations {{is}} 10 Epochs, Displaying Average Classification Accuracy Across 30 Runs.}
\label{tb: result 10}
\scalebox{0.7}{
\begin{tabular}{c|c c c c|cccc}
\bhline{1pt}
\multicolumn{1}{c|}{\multirow{2}{*}{10 Epochs}} &\multicolumn{4}{c|}{\textsc{Training Accuracy (\%)}} & \multicolumn{4}{c}{\textsc{Test Accuracy (\%)}}\\
& UCS & Fuzzy-UCS$_\text{VOTE}$ & Fuzzy-UCS$_\text{SWIN}$ & Fuzzy-UCS$_\text{DS}$ & UCS & Fuzzy-UCS$_\text{VOTE}$ & Fuzzy-UCS$_\text{SWIN}$ & Fuzzy-UCS$_\text{DS}$\\
\bhline{1pt}
\texttt{bnk} & \cellcolor{p}73.63 $-$ & 89.60 $-$ & \cellcolor{g}91.93 $+$ & 90.04 & \cellcolor{p}72.87 $-$ & 89.20 $-$ & \cellcolor{g}91.43 $+$ & 89.73 \\
\texttt{can} & \cellcolor{g}94.06 $\sim$ & \cellcolor{p}93.91 $-$ & 94.00 $\sim$ & 94.03 & \cellcolor{p}90.23 $-$ & 93.51 $\sim$ & \cellcolor{g}93.63 $\sim$ & 93.51 \\
\texttt{car} & \cellcolor{p}79.04 $-$ & 83.33 $-$ & 83.70 $\sim$ & \cellcolor{g}84.37 & \cellcolor{p}70.00 $\sim$ & 74.00 $\sim$ & \cellcolor{g}77.33 $\sim$ & 75.33 \\
\texttt{col} & \cellcolor{p}64.87 $-$ & \cellcolor{g}69.33 $+$ & 67.14 $-$ & 68.06 & \cellcolor{p}61.61 $\sim$ & \cellcolor{g}66.56 $\sim$ & 64.30 $\sim$ & 65.27 \\
\texttt{dbt} & \cellcolor{p}69.81 $-$ & 72.06 $-$ & 70.97 $-$ & \cellcolor{g}72.33 & \cellcolor{p}67.92 $\sim$ & 69.09 $\sim$ & 68.48 $\sim$ & \cellcolor{g}69.18 \\
\texttt{ecl} & \cellcolor{p}65.14 $-$ & 79.19 $-$ & 75.93 $-$ & \cellcolor{g}80.72 & \cellcolor{p}64.90 $-$ & 79.71 $-$ & 76.76 $-$ & \cellcolor{g}80.98 \\
\texttt{frt} & \cellcolor{g}86.06 $\sim$ & 83.02 $-$ & \cellcolor{p}81.29 $-$ & 85.78 & 81.00 $-$ & 81.63 $-$ & \cellcolor{p}79.85 $-$ & \cellcolor{g}83.70 \\
\texttt{gls} & 54.32 $-$ & \cellcolor{p}40.47 $-$ & \cellcolor{g}61.60 $\sim$ & 61.46 & 48.94 $\sim$ & \cellcolor{p}36.36 $-$ & \cellcolor{g}52.88 $\sim$ & \cellcolor{g}52.88 \\
\texttt{hrt} & \cellcolor{p}83.24 $-$ & 88.25 $-$ & \cellcolor{g}89.58 $\sim$ & 89.51 & \cellcolor{p}77.10 $-$ & 81.94 $\sim$ & 80.75 $\sim$ & \cellcolor{g}82.47 \\
\texttt{hpt} & \cellcolor{g}85.11 $+$ & \cellcolor{p}76.24 $-$ & 77.51 $\sim$ & 77.10 & \cellcolor{p}57.92 $-$ & 65.21 $\sim$ & 64.38 $\sim$ & \cellcolor{g}65.62 \\
\texttt{hcl} & \cellcolor{g}78.16 $+$ & 74.52 $-$ & \cellcolor{p}73.95 $-$ & 75.35 & \cellcolor{p}62.70 $-$ & 68.29 $\sim$ & 67.30 $\sim$ & \cellcolor{g}68.74 \\
\texttt{irs} & \cellcolor{p}77.56 $-$ & 93.68 $-$ & 94.07 $\sim$ & \cellcolor{g}94.35 & \cellcolor{p}74.67 $-$ & 93.56 $\sim$ & \cellcolor{g}94.22 $\sim$ & 93.78 \\
\texttt{lnd} & 37.11 $-$ & \cellcolor{p}27.71 $-$ & 35.35 $-$ & \cellcolor{g}43.34 & \cellcolor{p}32.16 $-$ & 36.57 $\sim$ & 35.69 $\sim$ & \cellcolor{g}37.25 \\
\texttt{mam} & 79.79 $-$ & 80.37 $-$ & \cellcolor{p}77.59 $-$ & \cellcolor{g}80.52 & \cellcolor{g}79.18 $\sim$ & 78.76 $\sim$ & \cellcolor{p}76.56 $-$ & 79.04 \\
\texttt{pdy} & \cellcolor{g}72.47 $+$ & 50.43 $-$ & \cellcolor{p}50.04 $-$ & 52.40 & \cellcolor{g}71.57 $+$ & 50.93 $-$ & \cellcolor{p}50.68 $-$ & 53.06 \\
\texttt{pis} & 85.56 $-$ & \cellcolor{g}86.66 $\sim$ & \cellcolor{p}85.37 $-$ & 86.64 & 84.42 $-$ & \cellcolor{g}85.89 $\sim$ & \cellcolor{p}84.26 $-$ & 85.71 \\
\texttt{pha} & 76.45 $\sim$ & \cellcolor{p}76.19 $-$ & \cellcolor{g}78.60 $+$ & 77.40 & \cellcolor{p}59.75 $-$ & \cellcolor{g}65.44 $\sim$ & 62.89 $\sim$ & 64.85 \\
\texttt{pre} & \cellcolor{p}99.33 $\sim$ & 99.64 $\sim$ & \cellcolor{g}100.0 $+$ & 99.61 & \cellcolor{p}99.34 $\sim$ & 99.67 $\sim$ & \cellcolor{g}100.0 $+$ & 99.60 \\
\texttt{pmp} & \cellcolor{p}86.45 $\sim$ & \cellcolor{g}86.88 $\sim$ & \cellcolor{p}86.45 $-$ & 86.84 & \cellcolor{p}85.35 $\sim$ & 86.19 $\sim$ & 85.88 $-$ & \cellcolor{g}86.24 \\
\texttt{rsn} & \cellcolor{p}80.33 $-$ & 83.71 $-$ & \cellcolor{g}85.16 $+$ & 84.38 & \cellcolor{p}81.74 $-$ & 83.37 $-$ & \cellcolor{g}85.11 $+$ & 84.37 \\
\texttt{seg} & \cellcolor{p}87.61 $-$ & 88.93 $-$ & \cellcolor{g}89.42 $\sim$ & 89.26 & \cellcolor{p}87.30 $-$ & 88.63 $-$ & \cellcolor{g}89.16 $\sim$ & 89.02 \\
\texttt{sir} & \cellcolor{p}71.81 $-$ & 83.59 $\sim$ & \cellcolor{g}83.96 $\sim$ & 83.59 & \cellcolor{p}69.67 $-$ & \cellcolor{g}81.00 $\sim$ & 79.33 $\sim$ & \cellcolor{g}81.00 \\
\texttt{smk} & \cellcolor{g}87.02 $+$ & 79.20 $+$ & 86.05 $+$ & \cellcolor{p}76.38 & \cellcolor{g}86.05 $+$ & 78.32 $+$ & 85.47 $+$ & \cellcolor{p}75.76 \\
\texttt{tae} & \cellcolor{p}53.75 $-$ & 57.43 $-$ & 60.64 $\sim$ & \cellcolor{g}61.48 & \cellcolor{p}45.62 $-$ & 49.79 $\sim$ & 52.08 $\sim$ & \cellcolor{g}53.75 \\
\texttt{tip} & 80.01 $\sim$ & 80.34 $\sim$ & \cellcolor{p}79.11 $-$ & \cellcolor{g}80.40 & 78.69 $-$ & \cellcolor{g}79.66 $\sim$ & \cellcolor{p}77.97 $-$ & 79.61 \\
\texttt{tit} & 70.05 $+$ & 69.28 $\sim$ & \cellcolor{g}70.22 $+$ & \cellcolor{p}69.22 & \cellcolor{g}70.30 $\sim$ & 68.89 $\sim$ & 69.07 $\sim$ & \cellcolor{p}68.56 \\
\texttt{wne} & \cellcolor{p}87.65 $-$ & 97.12 $-$ & 97.48 $\sim$ & \cellcolor{g}97.56 & \cellcolor{p}84.44 $-$ & 95.56 $\sim$ & 94.63 $\sim$ & \cellcolor{g}95.93 \\
\texttt{wbc} & 95.79 $-$ & 96.47 $-$ & \cellcolor{p}95.73 $-$ & \cellcolor{g}96.63 & 94.90 $\sim$ & \cellcolor{g}96.24 $\sim$ & \cellcolor{p}94.38 $-$ & 95.95 \\
\texttt{wpb} & \cellcolor{p}86.03 $-$ & 86.89 $-$ & \cellcolor{g}91.03 $+$ & 89.33 & \cellcolor{p}68.33 $-$ & \cellcolor{g}75.17 $\sim$ & 73.17 $\sim$ & \cellcolor{g}75.17 \\
\texttt{yst} & \cellcolor{p}45.25 $-$ & 45.91 $-$ & 51.66 $-$ & \cellcolor{g}54.14 & \cellcolor{p}42.10 $-$ & 44.38 $-$ & 50.74 $-$ & \cellcolor{g}53.58 \\
\bhline{1pt}
Rank & \cellcolor{p}\textit{3.05}$\downarrow^{\dag\dag}$ & \textit{2.75}$\downarrow^{\dag\dag}$ & \textit{2.35}$\downarrow$ & \cellcolor{g}\textit{1.85} & \cellcolor{p}\textit{3.43}$\downarrow^{\dag\dag}$ & \textit{2.25}$\downarrow^{\dag\dag}$ & \textit{2.55}$\downarrow^{\dag\dag}$ & \cellcolor{g}\textit{1.77} \\
Position & \textit{4} & \textit{3} & \textit{2} & \textit{1} & \textit{4} & \textit{2} & \textit{3} & \textit{1} \\
$+/-/\sim$ & 5/19/6 & 2/22/6 & 7/13/10 & - & 2/19/9 & 1/8/21 & 4/9/17 & - \\
\bhline{1pt}
$p$-value & 0.00683 & 0.000479 & 0.120 & - & 0.000209 & 0.0143 & 0.00917 & - \\
$p_\text{Holm}$-value & 0.0137 & 0.00144 & 0.120 & - & 0.000627 & 0.0183 & 0.0183 & - \\

\bhline{1pt}

\end{tabular}
}
\end{center}

\end{table*}

\begin{table*}[t]
\begin{center}
\caption{Results When the Number of Training Iterations {{is}} 50 Epochs, Displaying Average Classification Accuracy Across 30 Runs.}
\label{tb: result 50}
\scalebox{0.7}{
\begin{tabular}{c |cccc|cccc}
\bhline{1pt}
\multicolumn{1}{c|}{\multirow{2}{*}{50 Epochs}} &\multicolumn{4}{c|}{\textsc{Training Accuracy (\%)}} & \multicolumn{4}{c}{\textsc{Test Accuracy (\%)}}\\
& UCS & Fuzzy-UCS$_\text{VOTE}$ & Fuzzy-UCS$_\text{SWIN}$ & Fuzzy-UCS$_\text{DS}$ & UCS & Fuzzy-UCS$_\text{VOTE}$ & Fuzzy-UCS$_\text{SWIN}$ & Fuzzy-UCS$_\text{DS}$\\
\bhline{1pt}
\texttt{bnk} & \cellcolor{g}94.93 $+$ & \cellcolor{p}92.63 $-$ & 93.65 $+$ & 92.95 & \cellcolor{g}94.47 $+$ & \cellcolor{p}92.34 $-$ & 93.00 $\sim$ & 92.63 \\
\texttt{can} & \cellcolor{g}97.19 $+$ & 95.29 $-$ & \cellcolor{p}94.83 $-$ & 95.39 & \cellcolor{p}91.75 $-$ & 94.74 $\sim$ & 94.44 $\sim$ & \cellcolor{g}94.80 \\
\texttt{car} & \cellcolor{p}81.56 $-$ & 87.93 $\sim$ & 88.07 $\sim$ & \cellcolor{g}88.22 & \cellcolor{p}71.33 $\sim$ & 76.00 $\sim$ & \cellcolor{g}76.67 $\sim$ & 74.00 \\
\texttt{col} & \cellcolor{g}76.01 $+$ & 71.85 $+$ & \cellcolor{p}68.24 $-$ & 71.37 & \cellcolor{g}70.00 $\sim$ & 69.35 $\sim$ & \cellcolor{p}65.81 $-$ & 69.03 \\
\texttt{dbt} & \cellcolor{g}78.12 $+$ & 76.36 $\sim$ & \cellcolor{p}73.57 $-$ & 76.48 & 72.90 $\sim$ & \cellcolor{g}73.07 $\sim$ & \cellcolor{p}70.04 $-$ & 72.94 \\
\texttt{ecl} & \cellcolor{p}78.55 $-$ & \cellcolor{g}85.79 $\sim$ & 80.39 $-$ & 85.64 & \cellcolor{p}76.18 $-$ & \cellcolor{g}84.02 $\sim$ & 79.80 $-$ & 83.92 \\
\texttt{frt} & \cellcolor{g}93.67 $+$ & 87.12 $-$ & \cellcolor{p}85.17 $-$ & 88.43 & 83.41 $-$ & 84.74 $-$ & \cellcolor{p}82.33 $-$ & \cellcolor{g}86.44 \\
\texttt{gls} & \cellcolor{p}65.19 $-$ & 72.01 $\sim$ & 71.75 $\sim$ & \cellcolor{g}72.14 & \cellcolor{p}53.33 $-$ & 62.73 $\sim$ & 64.09 $\sim$ & \cellcolor{g}64.39 \\
\texttt{hrt} & \cellcolor{p}90.53 $-$ & 94.66 $-$ & 94.35 $-$ & \cellcolor{g}95.05 & \cellcolor{p}81.08 $\sim$ & 81.83 $\sim$ & \cellcolor{g}82.69 $\sim$ & 81.83 \\
\texttt{hpt} & 90.10 $-$ & 91.46 $-$ & \cellcolor{p}89.88 $-$ & \cellcolor{g}92.33 & \cellcolor{p}57.08 $-$ & 64.58 $\sim$ & 62.08 $\sim$ & \cellcolor{g}65.62 \\
\texttt{hcl} & \cellcolor{g}89.86 $+$ & 87.25 $-$ & \cellcolor{p}84.19 $-$ & 87.76 & \cellcolor{p}63.69 $-$ & \cellcolor{g}71.80 $\sim$ & 69.37 $\sim$ & 71.62 \\
\texttt{irs} & \cellcolor{p}84.17 $-$ & 94.81 $-$ & \cellcolor{g}95.51 $\sim$ & 95.26 & \cellcolor{p}81.78 $-$ & 94.00 $\sim$ & \cellcolor{g}94.89 $\sim$ & 94.22 \\
\texttt{lnd} & \cellcolor{g}55.29 $+$ & \cellcolor{p}32.05 $-$ & 39.18 $-$ & 46.27 & \cellcolor{g}42.06 $\sim$ & \cellcolor{p}38.24 $\sim$ & 40.10 $\sim$ & 40.20 \\
\texttt{mam} & \cellcolor{g}83.10 $+$ & 80.89 $-$ & \cellcolor{p}77.89 $-$ & 81.07 & \cellcolor{g}80.72 $\sim$ & 79.73 $\sim$ & \cellcolor{p}76.77 $-$ & 80.03 \\
\texttt{pdy} & \cellcolor{g}89.16 $+$ & 50.94 $-$ & \cellcolor{p}50.70 $-$ & 53.94 & \cellcolor{g}88.24 $+$ & 51.46 $-$ & \cellcolor{p}51.36 $-$ & 54.63 \\
\texttt{pis} & \cellcolor{g}87.54 $+$ & 86.89 $\sim$ & \cellcolor{p}85.53 $-$ & 86.91 & \cellcolor{g}86.12 $\sim$ & 85.88 $\sim$ & \cellcolor{p}84.67 $-$ & 85.92 \\
\texttt{pha} & \cellcolor{p}79.17 $-$ & 82.10 $-$ & \cellcolor{g}86.80 $+$ & 85.64 & \cellcolor{p}62.55 $-$ & \cellcolor{g}66.96 $\sim$ & 64.41 $\sim$ & 66.13 \\
\texttt{pre} & 99.63 $\sim$ & \cellcolor{p}98.01 $-$ & \cellcolor{g}100.0 $+$ & 99.76 & 99.64 $\sim$ & \cellcolor{p}98.03 $-$ & \cellcolor{g}100.0 $+$ & 99.79 \\
\texttt{pmp} & \cellcolor{g}87.24 $\sim$ & 86.99 $\sim$ & \cellcolor{p}86.89 $-$ & 87.03 & \cellcolor{p}86.01 $\sim$ & 86.17 $\sim$ & 86.05 $\sim$ & \cellcolor{g}86.21 \\
\texttt{rsn} & 85.02 $\sim$ & \cellcolor{p}84.33 $-$ & \cellcolor{g}85.51 $+$ & 85.13 & 84.63 $\sim$ & \cellcolor{p}84.19 $-$ & \cellcolor{g}85.30 $\sim$ & 85.07 \\
\texttt{seg} & \cellcolor{g}95.40 $+$ & \cellcolor{p}90.10 $-$ & 90.52 $\sim$ & 90.42 & \cellcolor{g}93.02 $+$ & \cellcolor{p}89.88 $-$ & 90.01 $\sim$ & 90.30 \\
\texttt{sir} & \cellcolor{p}76.96 $-$ & 84.78 $\sim$ & \cellcolor{g}85.00 $\sim$ & 84.78 & \cellcolor{p}70.67 $-$ & 81.67 $\sim$ & 81.33 $\sim$ & \cellcolor{g}82.00 \\
\texttt{smk} & \cellcolor{g}97.66 $+$ & 95.39 $-$ & \cellcolor{p}93.49 $-$ & 95.71 & \cellcolor{g}96.50 $+$ & 95.08 $\sim$ & \cellcolor{p}93.14 $-$ & 95.11 \\
\texttt{tae} & \cellcolor{p}59.23 $-$ & 63.63 $-$ & 64.15 $\sim$ & \cellcolor{g}64.35 & \cellcolor{p}49.17 $-$ & 54.58 $\sim$ & 53.75 $\sim$ & \cellcolor{g}56.46 \\
\texttt{tip} & \cellcolor{g}82.71 $+$ & 81.09 $-$ & \cellcolor{p}80.24 $-$ & 81.20 & \cellcolor{g}80.47 $\sim$ & 80.00 $-$ & \cellcolor{p}78.88 $-$ & 80.23 \\
\texttt{tit} & \cellcolor{g}73.25 $+$ & 72.28 $\sim$ & \cellcolor{p}70.57 $-$ & 72.38 & \cellcolor{g}72.30 $\sim$ & 71.30 $\sim$ & \cellcolor{p}68.93 $-$ & 71.52 \\
\texttt{wne} & \cellcolor{p}93.62 $-$ & 98.31 $-$ & 98.58 $\sim$ & \cellcolor{g}98.79 & \cellcolor{p}88.52 $-$ & 96.48 $\sim$ & 96.11 $\sim$ & \cellcolor{g}96.67 \\
\texttt{wbc} & \cellcolor{g}97.59 $+$ & 97.12 $\sim$ & \cellcolor{p}96.42 $-$ & 97.07 & \cellcolor{g}96.19 $\sim$ & \cellcolor{g}96.19 $\sim$ & \cellcolor{p}95.29 $-$ & 95.95 \\
\texttt{wpb} & \cellcolor{g}95.21 $+$ & \cellcolor{p}91.57 $-$ & 93.60 $+$ & 92.58 & \cellcolor{p}70.00 $\sim$ & \cellcolor{g}72.17 $\sim$ & 71.67 $\sim$ & 72.00 \\
\texttt{yst} & \cellcolor{g}60.73 $+$ & \cellcolor{p}53.26 $-$ & 56.22 $-$ & 58.78 & 56.89 $\sim$ & \cellcolor{p}52.04 $-$ & 53.65 $-$ & \cellcolor{g}57.61 \\
\bhline{1pt}
Rank & \textit{2.10}$\downarrow$ & \cellcolor{p}\textit{2.98}$\downarrow^{\dag\dag}$ & \textit{2.93}$\downarrow^{\dag\dag}$ & \cellcolor{g}\textit{1.98} & \textit{2.72}$\downarrow^{\dag\dag}$ & \textit{2.50}$\downarrow^{\dag\dag}$ & \cellcolor{p}\textit{2.93}$\downarrow^{\dag\dag}$ & \cellcolor{g}\textit{1.85} \\
Position & \textit{2} & \textit{4} & \textit{3} & \textit{1} & \textit{3} & \textit{2} & \textit{4} & \textit{1} \\
$+/-/\sim$ & 17/10/3 & 1/20/9 & 5/18/7 & - & 4/11/15 & 0/8/22 & 1/12/17 & - \\
\bhline{1pt}
$p$-value & 0.968 & 4.80E-05 & 0.00148 & - & 0.0483 & 0.0121 & 0.000529 & - \\
$p_\text{Holm}$-value & 0.968 & 0.000144 & 0.00297 & - & 0.0483 & 0.0243 & 0.00159 & - \\

\bhline{1pt}

\end{tabular}
}
\end{center}

\end{table*}

{This section presents the standard accuracy results for all systems. While macro F1 scores (presented in Section \ref{ss: results}) provide a balanced measure accounting for both precision and recall, accuracy results are included here for comprehensive evaluation.}

{Tables \ref{tb: result 10} and \ref{tb: result 50} present each system's average training and test classification accuracy and average rank across all datasets At the 10th and 50th epochs, respectively.
Green-shaded values denote the best values among all systems, while peach-shaded values indicate the worst values among all systems. The terms ``Rank'' and ``Position'' denote each system's overall average rank obtained by using the Friedman test and its position in the final ranking, respectively. Statistical results of the Wilcoxon signed-rank test are summarized with symbols wherein ``$+$'', ``$-$'', and ``$\sim$'' represent that the classification accuracy of a conventional system is significantly better, worse, and competitive compared to that obtained by the proposed Fuzzy-UCS$_\text{DS}$, respectively. The ``$p$-value'' and ``$p_\text{Holm}$-value'' are derived from the Wilcoxon signed-rank test and the Holm-adjusted Wilcoxon signed-rank test, respectively. Arrows denote whether the rank improved or declined compared to Fuzzy-UCS$_\text{DS}$. $\dag$ ($\dag\dag$) indicates statistically significant differences compared to Fuzzy-UCS$_\text{DS}$, i.e., $p$-value ($p_\text{Holm}$-value) $<\alpha=0.05$.

Based on Tables \ref{tb: result 10} and \ref{tb: result 50}, Fuzzy-UCS$_\text{DS}$ significantly outperformed all other systems ($p_\text{Holm}<0.05$) in test accuracy, securing the highest rank irrespective of the number of training epochs. In terms of training accuracy, Fuzzy-UCS$_\text{DS}$ also significantly outperformed all other systems ($p_\text{Holm}<0.05$), with the exception of Fuzzy-UCS$_\text{SWIN}$ at the 10th epoch and UCS at the 50th epoch, thereby achieving the highest rank. Furthermore, Tables \ref{tb: result 10} and \ref{tb: result 50} reveal that Fuzzy-UCS$_\text{DS}$ did not register the lowest accuracy for any problem regardless of the number of training epochs, except for the \texttt{smk} and \texttt{tit} problems in training and test accuracy at the 10th epoch.  These findings underscore that Fuzzy-UCS$_\text{DS}$ is a robust system, consistently exhibiting high classification performance across a diverse range of problems.}

\clearpage
\section{Experiments on Problems with Known Ground Truth}
\label{sec: sup impact of dataset selection and ground truth}
{While UCI and Kaggle datasets used in our experiment are invaluable for their diversity and ability to facilitate comparisons with existing work, they present limitations in terms of knowing inherent complexities and feature relevancies.

To mitigate the limitation of unknown ground truths in real-world datasets, we conduct supplementary experiments using artificial datasets where the underlying relationships and feature relevancies are predefined. 
Specifically, we utilize three artificial noiseless benchmark problems commonly employed in the LCS literature:

\textit{(1) Real-Valued Multiplexer Problem.} The $n$-bit multiplexer problem ($n$-\textit{MUX}) was initially introduced as a binary-valued classification problem to assess the generalization capability of XCS \cite{wilson1995xcs}. It was later extended for real-valued classification tasks \cite{wilson1999xcsr,nakata2020learning,aenugu2019lexicase}. The $n$-\textit{MUX} is tailored for binary strings defined by $n=k+2^k$, where the initial $k$ bits, when converted to decimal, correspond to one of the subsequent $2^k$ bit positions. The correct class corresponds to the bit the first $k$ bits indicate. In the real-valued version, $n$-\textit{RMUX}, upon receiving an input $\mathbf{x}\in[0,1]^n$, each attribute $x_i$ is binarized using a threshold of 0.5 (i.e., 0 if $x_i<0.5$, and 1 otherwise). The correct class for this real-valued input is ascertained using the same procedure applied to $n$-\textit{MUX}. We use \textit{11-RMUX}, denoted as \texttt{mux}.

\textit{(2) Real-Valued Majority-On Problem.}
The $n$-bit majority-on problem ($n$-\textit{MAJ}) was originally designed for binary-valued inputs \cite{iqbal2013reusing}. In $n$-\textit{MAJ}, the correct class is determined as 1 if the count of ``1''s surpasses that of ``0''s, and 0 otherwise. This article utilizes the recent extended version, $n$-\textit{RMAJ}, tailored for real-valued classification \cite{hamasaki2021minimum}. Similar to $n$-\textit{RMUX}, each attribute $x_i$ is binarized using the 0.5 threshold. The $n$-\textit{RMAJ} is a highly overlapping problem and is known to suffer significantly from performance improvements in the hyperrectangular representation used in UCS. This article uses \textit{11-RMAJ}, denoted as \texttt{maj}.

\textit{(3) Real-Valued Carry Problem.}
The $n$-bit carry problem ($n$-\textit{CAR}) was initially developed for binary-valued inputs \cite{iqbal2013reusing}. In $n$-\textit{CAR}, equal-length binary numbers are summed. Consequently, the output is 1 if there is a carry-over and 0 otherwise. For instance, with 3-bit numbers 101 and 001, the output stands at 0, while for 100 and 100, it is 1. $n$-\textit{CAR} is an overlapping problem. This article extends $n$-\textit{CAR} to a real-valued classification framework termed $n$-\textit{RCAR}. Each attribute $x_i$ is binarized using the 0.5 threshold, the same as in $n$-\textit{RMUX} and $n$-\textit{RMAJ}. This article uses \textit{12-RCAR}, denoted as \texttt{car}.

For all the artificial noiseless problems, each dataset consists of 6,000 data points uniformly sampled at random from the input space. The details of the datasets are shown in Table \ref{tb: sup artifical problem dataset}. The settings used in our experiments adhere to those outlined in Section \ref{ss: experimental setup}. The number of training iterations is set to 10 and 50 epochs. 
To the best of our knowledge, there has been no comprehensive evaluation of the performance of LFCSs with linguistic rules, including Fuzzy-UCS, across all these problem domains. It is worth noting that these benchmark problems were originally designed for real-valued LCSs, such as XCSR and UCS, where the antecedent part of each rule is represented by a combination of lower and upper bounds for each dimension. }

\begin{table}[t]
\begin{center}
\caption{Properties of the three Artificial Noiseless Datasets. The Columns Describe: the Identifier (ID.), the Name (Name), the Number of Instances ($\#$\textsc{Inst.}), the Total Number of Features ($\#$\textsc{Fea.}), the Number of Classes ($\#$\textsc{Cl.}), the Percentage of Missing Attributes ($\%$\textsc{Mis.}), {the Percentage of Instances of the Majority Class ($\%$\textsc{Maj.}), the Percentage of Instances of the Minority Class ($\%$\textsc{Min.})}, and the Source (Ref.).}
\label{tb: sup artifical problem dataset}
\footnotesize
\scalebox{0.9}{
\begin{tabular}{c l c c c c c c c}
\bhline{1pt}
ID. & Name  & $\#$\textsc{Inst.}  & $\#$\textsc{Fea.} &$\#$\textsc{Cl.} & $\%$\textsc{Mis.} & {$\%$\textsc{Maj.}} & {$\%$\textsc{Min.}} & Ref.
\\
\bhline{1pt}
\texttt{car} &12 bit real-valued carry & 6000 & 12 & 2 & 0 & 50.85 & {49.15} & \cite{iqbal2013reusing} \\
\texttt{maj} &11 bit real-valued majority-on & 6000 & 11 & 2 & 0 & {50.22} & {49.78} & \cite{hamasaki2021minimum}\\
\texttt{mux} &11 bit real-valued multiplexer & 6000 & 11 & 2 & 0 & 50.50 & 49.50 & \cite{wilson1999xcsr}\\
\bhline{1pt}
\end{tabular}
}
\end{center}
\end{table}

    \begin{table}[t]
\begin{center}
\caption{Results When the Number of Training Iterations is 10 Epochs, Displaying Average Classification Accuracy Across 30 Runs.}

\label{tb: sup artificial problem 10epoch}
\scalebox{0.7}{
\begin{tabular}{c|c c c c|cccc}
\bhline{1pt}
\multicolumn{1}{c|}{\multirow{2}{*}{10 Epochs}} &\multicolumn{4}{c|}{\textsc{Training Accuracy (\%)}} & \multicolumn{4}{c}{\textsc{Test Accuracy (\%)}}\\
& UCS & Fuzzy-UCS$_\text{VOTE}$ & Fuzzy-UCS$_\text{SWIN}$ & Fuzzy-UCS$_\text{DS}$ & UCS & Fuzzy-UCS$_\text{VOTE}$ & Fuzzy-UCS$_\text{SWIN}$ & Fuzzy-UCS$_\text{DS}$\\
\bhline{1pt}
\texttt{car} & \cellcolor{g}87.57 $+$ & \cellcolor{p}85.57 $-$ & 86.50 $+$ & 85.92 & 86.45 $\sim$ & \cellcolor{p}85.74 $\sim$ & \cellcolor{g}86.64 $+$ & 85.95 \\
\texttt{maj} & \cellcolor{p}83.64 $-$ & 85.72 $-$ & 86.04 $-$ & \cellcolor{g}86.34 & \cellcolor{p}81.84 $-$ & 85.54 $-$ & 85.31 $-$ & \cellcolor{g}86.23 \\
\texttt{mux} & \cellcolor{p}72.32 $-$ & 74.53 $-$ & \cellcolor{g}76.42 $+$ & 75.00 & \cellcolor{p}69.54 $-$ & 73.26 $-$ & \cellcolor{g}74.73 $+$ & 73.66 \\
\bhline{1pt}
Rank & \textit{3.00}$\downarrow$ & \cellcolor{p}\textit{3.33}$\downarrow$ & \cellcolor{g}\textit{1.67}$\uparrow$ & \textit{2.00} & \cellcolor{p}\textit{3.33}$\downarrow$ & \textit{3.00}$\downarrow$ & \cellcolor{g}\textit{1.67}$\uparrow$ & \textit{2.00} \\
Position & \textit{3} & \textit{4} & \textit{1} & \textit{2} & \textit{4} & \textit{3} & \textit{1} & \textit{2} \\
$+/-/\sim$ & 1/2/0 & 0/3/0 & 2/1/0 & - & 0/2/1 & 0/2/1 & 2/1/0 & - \\


\bhline{1pt}

\end{tabular}
}
\end{center}

\end{table}

    \begin{table}[t]
\begin{center}
\caption{Results When the Number of Training Iterations is 50 Epochs, Displaying Average Classification Accuracy Across 30 Runs.}

\label{tb: sup artificial problem 50epoch}
\scalebox{0.7}{
\begin{tabular}{c|c c c c|cccc}
\bhline{1pt}
\multicolumn{1}{c|}{\multirow{2}{*}{50 Epochs}} &\multicolumn{4}{c|}{\textsc{Training Accuracy (\%)}} & \multicolumn{4}{c}{\textsc{Test Accuracy (\%)}}\\
& UCS & Fuzzy-UCS$_\text{VOTE}$ & Fuzzy-UCS$_\text{SWIN}$ & Fuzzy-UCS$_\text{DS}$ & UCS & Fuzzy-UCS$_\text{VOTE}$ & Fuzzy-UCS$_\text{SWIN}$ & Fuzzy-UCS$_\text{DS}$\\
\bhline{1pt}
\texttt{car} & \cellcolor{g}93.89 $+$ & \cellcolor{p}86.50 $-$ & 88.22 $+$ & 86.99 & \cellcolor{g}92.28 $+$ & \cellcolor{p}86.43 $-$ & 88.08 $+$ & 86.87 \\
\texttt{maj} & \cellcolor{g}90.19 $+$ & \cellcolor{p}86.22 $-$ & 87.07 $\sim$ & 86.90 & \cellcolor{g}87.23 $\sim$ & 85.81 $-$ & \cellcolor{p}85.68 $-$ & 86.44 \\
\texttt{mux} & \cellcolor{g}93.02 $+$ & 86.50 $\sim$ & \cellcolor{p}86.08 $-$ & 86.51 & \cellcolor{g}92.17 $+$ & 85.16 $\sim$ & \cellcolor{p}84.87 $\sim$ & 85.26 \\
\bhline{1pt}
Rank & \cellcolor{g}\textit{1.00}$\uparrow$ & \cellcolor{p}\textit{3.67}$\downarrow$ & \textit{2.67} & \textit{2.67} & \cellcolor{g}\textit{1.00}$\uparrow$ & \cellcolor{p}\textit{3.33}$\downarrow$ & \cellcolor{p}\textit{3.33}$\downarrow$ & \textit{2.33} \\
Position & \textit{1} & \textit{4} & \textit{2.5} & \textit{2.5} & \textit{1} & \textit{3.5} & \textit{3.5} & \textit{2} \\
$+/-/\sim$ & 3/0/0 & 0/2/1 & 1/1/1 & - & 2/0/1 & 0/2/1 & 1/1/1 & - \\

\bhline{1pt}

\end{tabular}
}
\end{center}
\end{table}

{
Tables \ref{tb: sup artificial problem 10epoch} and \ref{tb: sup artificial problem 50epoch} present each system's average training and test classification accuracy and average rank across all datasets at the 10th and 50th epochs, respectively.
Green-shaded values denote the best values among all systems, while peach-shaded values indicate the worst values among all systems. The terms ``Rank'' and ``Position'' denote each system's overall average rank obtained by using the Friedman test and its position in the final ranking, respectively. Statistical results of the Wilcoxon signed-rank test are summarized with symbols wherein ``$+$'', ``$-$'', and ``$\sim$'' represent that the classification accuracy of a conventional system is significantly better, worse, and competitive compared to that obtained by the proposed Fuzzy-UCS$_\text{DS}$, respectively. The ``$p$-value'' is derived from the Wilcoxon signed-rank test. Arrows denote whether the rank improved or declined compared to Fuzzy-UCS$_\text{DS}$.

Based on Table \ref{tb: sup artificial problem 10epoch}, at the 10th epoch, \swin\ achieved the highest rank in both training and testing. This suggests that single winner-based inference, which relies on the rule with the highest fitness, was effective for noiseless problems when training iterations were limited. Conversely, UCS recorded the lowest rank in testing at the 10th epoch. As discussed in Section \ref{sss: comparison with the ucs nonfuzzy classifier system}, this is likely because the crisp rules used by UCS were not sufficiently optimized with limited training iterations. Indeed, when given sufficient training time (i.e., at the 50th epoch, cf. Table \ref{tb: sup artificial problem 50epoch}), UCS achieved the highest rank in both training and testing. This can be attributed to UCS's rule antecedent representation (i.e., unordered bound hyperrectangular representation) being more flexible than \all 's linguistic label representation, coupled with the noiseless nature of the problems solved.

It is noteworthy that in situations where rules were sufficiently optimized (i.e., at the 50th epoch), the proposed \ds\ outperformed \vote\ and \swin\ in terms of test accuracy, as evidenced by the average rankings. Furthermore, \ds\ never recorded the worst performance among all systems at both the 10th and 50th epochs (i.e., no peach-colored cells), leading to the conclusion that \ds\ demonstrates stable performance regardless of the presence or absence of noise. Particularly, as shown in Appendix \ref{sec: sup Experimental Results on Standard Accuracy}, \ds\ exhibits high classification accuracy for real-world problems with high uncertainty, such as those involving noise.

In summary, the insights gained from experiments with artificial noiseless datasets can be summarized as follows:
\begin{itemize}
    \item UCS, which uses non-linguistic crisp rules, records superior test accuracy compared to \all\ in noiseless problems. However, when training iterations are limited, its test accuracy is inferior to \all\ due to the higher difficulty of optimizing crisp rules.
    \item The single-winner-based inference (\swin), which relies on the rule with the highest fitness, is often effective in noiseless problems.
    \item \ds\ demonstrates stable performance regardless of the presence or absence of noise. Notably, it achieves high classification accuracy for real-world problems characterized by high uncertainty, including those involving noise.
\end{itemize}
}

\clearpage
\section{Scalability and Performance on Larger-Scale Datasets}
\label{sec: sup scalability considerations}
{

To further validate our algorithm's effectiveness and scalability, we extend our experiments to include larger and more complex datasets. These datasets, characterized by either a high number of features or instances, are detailed in Table \ref{tb: sup large scale dataset}. This expansion allows us to assess how well our proposed algorithm adapts to increased data volume and feature complexity, mirroring real-world challenges across various domains.

The settings used in our experiments adhere to those outlined in Section \ref{ss: experimental setup}. The number of training iterations is set to 50 epochs. For datasets with a large number of features (60 or more), specifically \texttt{con}, \texttt{lib}, and \texttt{mcd}, the hyperparameter $P_\#$ for both UCS and \all\ is set to 0.8 to mitigate the risk of cover-delete cycles \cite{butz2004toward}, a common occurrence in high-dimensional input problems.

\begin{table}[b]
\begin{center}
\caption{Properties of the Four {Larger-Scale} Real-World Datasets. ``ID.'', ``Name'', ``$\#$\textsc{Inst.}'', ``$\#$\textsc{Fea.}'', ``$\#$\textsc{Cl.}'', ``$\%$\textsc{Mis.}'', ``$\%$\textsc{Maj.}'', ``$\%$\textsc{Min.}'', and ``Ref.'' Should be Interpreted as in Table \ref{tb: sup artifical problem dataset}. Numerical Values Characterizing ``{Larger-Scale}'' are Shown in Bold.}
\label{tb: sup large scale dataset}
\footnotesize
\scalebox{0.9}{
\begin{tabular}{c l c c c c c c c}
\bhline{1pt}
ID. & Name  & $\#$\textsc{Inst.}  & $\#$\textsc{Fea.} &$\#$\textsc{Cl.} & $\%$\textsc{Mis.} & {$\%$\textsc{Maj.}} & {$\%$\textsc{Min.}} & Ref.
\\
\bhline{1pt}
\texttt{con} & Connectionist bench (sonar, mines vs. rocks) & 208 & \textbf{60} & 2 & 0 & 53.37 & 46.63 & \cite{dua2019uci}\\
\texttt{lib} & Libras movement & 360 & \textbf{90} & 15 & 0 & 6.67 & 6.67 & \cite{dua2019uci}\\
\texttt{mcd} & Myocardial\_infarction & 1700 & \textbf{111} & 8 & 8.47 & 84.06 & 0.71 & \tablefootnote{{\url{https://www.kaggle.com/datasets/shilpiiic/myocardial-infarction/data} (\today)}}\\
\texttt{phi} & Phishing Websites & \textbf{11055} & 30 & 2 & 0 & 55.69 & 44.31 & \cite{dua2019uci}\\
\bhline{1pt}
\end{tabular}
}
\end{center}
\end{table}

\begin{table}[b]
\begin{center}
\caption{Results When the Number of Training Iterations is 10 Epochs, Displaying Average Classification Accuracy Across 30 Runs.}

\label{tb: sup large scale 10epoch}
\scalebox{0.7}{
\begin{tabular}{c|c c c c|cccc}
\bhline{1pt}
\multicolumn{1}{c|}{\multirow{2}{*}{10 Epochs}} &\multicolumn{4}{c|}{\textsc{Training Accuracy (\%)}} & \multicolumn{4}{c}{\textsc{Test Accuracy (\%)}}\\
& UCS & Fuzzy-UCS$_\text{VOTE}$ & Fuzzy-UCS$_\text{SWIN}$ & Fuzzy-UCS$_\text{DS}$ & UCS & Fuzzy-UCS$_\text{VOTE}$ & Fuzzy-UCS$_\text{SWIN}$ & Fuzzy-UCS$_\text{DS}$\\
\bhline{1pt}
\texttt{con} & \cellcolor{p}78.07 $-$ & 89.13 $-$ & \cellcolor{g}93.24 $+$ & 91.19 & \cellcolor{p}28.61 $-$ & 68.15 $-$ & \cellcolor{g}72.13 $\sim$ & 70.09 \\
\texttt{lib} & \cellcolor{g}99.73 $+$ & \cellcolor{p}85.48 $-$ & 87.93 $\sim$ & 88.72 & \cellcolor{p}83.24 $-$ & 84.94 $-$ & 84.84 $-$ & \cellcolor{g}85.35 \\
\texttt{mcd} & \cellcolor{g}93.76 $+$ & \cellcolor{p}85.18 $-$ & 85.61 $-$ & 87.06 & \cellcolor{p}76.99 $-$ & 92.99 $-$ & 93.09 $\sim$ & \cellcolor{g}93.21 \\
\texttt{phi} & \cellcolor{p}80.19 $-$ & 93.58 $-$ & 93.81 $\sim$ & \cellcolor{g}93.87 & \cellcolor{p}60.32 $-$ & 74.13 $\sim$ & 72.54 $\sim$ & \cellcolor{g}74.60 \\
\bhline{1pt}
Rank & \textit{2.50}$\downarrow$ & \cellcolor{p}\textit{3.50}$\downarrow$ & \textit{2.25}$\downarrow$ & \cellcolor{g}\textit{1.75} & \cellcolor{p}\textit{4.00}$\downarrow$ & \textit{2.50}$\downarrow$ & \textit{2.25}$\downarrow$ & \cellcolor{g}\textit{1.25} \\
Position & \textit{3} & \textit{4} & \textit{2} & \textit{1} & \textit{4} & \textit{3} & \textit{2} & \textit{1} \\
$+/-/\sim$ & 2/2/0 & 0/4/0 & 1/1/2 & - & 0/4/0 & 0/3/1 & 0/1/3 & - \\
\bhline{1pt}

\end{tabular}
}
\end{center}

\end{table}

\begin{table}[b]
\begin{center}
\caption{Results When the Number of Training Iterations is 50 Epochs, Displaying Average Classification Accuracy Across 30 Runs.}

\label{tb: sup large scale 50epoch}
\scalebox{0.7}{
\begin{tabular}{c|c c c c|cccc}
\bhline{1pt}
\multicolumn{1}{c|}{\multirow{2}{*}{50 Epochs}} &\multicolumn{4}{c|}{\textsc{Training Accuracy (\%)}} & \multicolumn{4}{c}{\textsc{Test Accuracy (\%)}}\\
& UCS & Fuzzy-UCS$_\text{VOTE}$ & Fuzzy-UCS$_\text{SWIN}$ & Fuzzy-UCS$_\text{DS}$ & UCS & Fuzzy-UCS$_\text{VOTE}$ & Fuzzy-UCS$_\text{SWIN}$ & Fuzzy-UCS$_\text{DS}$\\
\bhline{1pt}
\texttt{con} & \cellcolor{p}91.80 $\sim$ & 91.84 $-$ & \cellcolor{g}94.39 $+$ & 92.80 & \cellcolor{p}66.19 $-$ & 76.67 $\sim$ & 74.92 $\sim$ & \cellcolor{g}77.14 \\
\texttt{lib} & \cellcolor{g}99.89 $+$ & \cellcolor{p}83.24 $-$ & 85.29 $\sim$ & 85.28 & \cellcolor{p}28.43 $-$ & 65.65 $-$ & 66.39 $\sim$ & \cellcolor{g}67.13 \\
\texttt{mcd} & \cellcolor{g}93.80 $+$ & \cellcolor{p}85.68 $-$ & 87.25 $\sim$ & 87.67 & \cellcolor{p}83.25 $-$ & 85.27 $\sim$ & 85.00 $\sim$ & \cellcolor{g}85.43 \\
\texttt{phi} & \cellcolor{p}80.34 $-$ & 94.11 $-$ & \cellcolor{g}95.00 $\sim$ & 94.91 & \cellcolor{p}76.91 $-$ & 93.70 $-$ & 93.98 $-$ & \cellcolor{g}94.21 \\
\bhline{1pt}
Rank & \textit{2.50}$\downarrow$ & \cellcolor{p}\textit{3.50}$\downarrow$ & \cellcolor{g}\textit{1.75}$\uparrow$ & \textit{2.25} & \cellcolor{p}\textit{4.00}$\downarrow$ & \textit{2.50}$\downarrow$ & \textit{2.50}$\downarrow$ & \cellcolor{g}\textit{1.00} \\
Position & \textit{3} & \textit{4} & \textit{1} & \textit{2} & \textit{4} & \textit{2.5} & \textit{2.5} & \textit{1} \\
$+/-/\sim$ & 2/1/1 & 0/4/0 & 1/0/3 & - & 0/4/0 & 0/2/2 & 0/1/3 & - \\

\bhline{1pt}

\end{tabular}
}
\end{center}

\end{table}
Tables \ref{tb: sup large scale 10epoch} and \ref{tb: sup large scale 50epoch} present each system's average training and test classification accuracy and average rank across all datasets at the 10th and 50th epochs, respectively.
``$+$'', ``$-$'', ``$\sim$'', ``Rank'', ``Position'', arrows, and green- and peach-shaded values should be interpreted as in Tables \ref{tb: sup artificial problem 10epoch} and \ref{tb: sup artificial problem 50epoch}.

From Tables \ref{tb: sup large scale 10epoch} and \ref{tb: sup large scale 50epoch}, \ds\ recorded the highest rank in test accuracy at both 10 and 50 epochs. Furthermore, across all datasets, no instances were observed where \ds\ had significantly worse test accuracy (denoted by ``$+$''). Therefore, \ds\ demonstrates good performance even on {larger-scale} datasets. On the other hand, UCS using crisp rules and \swin\ relying solely on the rule with the highest fitness showed significant superiority during training but not during testing. This indicates a tendency to overfit training data as the dataset scale increases.

These results suggest that while UCS and \swin\ may excel during training, they are prone to overfitting on {larger-scale} datasets. In contrast, \ds\ maintains stable performance regardless of dataset size, demonstrating its robustness and scalability.}

\clearpage
\section{Comparison With a State-of-the-Art Nonfuzzy LCS}
\label{sec: sup comparison with a state-of-the-art non-fuzzy lcs}
{
To further evaluate our proposed algorithm's performance against a state-of-the-art nonfuzzy LCS designed for noisy datasets, we conduct additional experiments comparing it with scikit-ExSTraCS \cite{urbanowicz2015exstracs}\footnote{\url{https://github.com/UrbsLab/scikit-ExSTraCS/tree/master} (\today)}. This system is chosen due to its open-source availability and reputation as a robust nonfuzzy LCS capable of handling complex noisy datasets.

The problems and settings used in our experiments adhere to those outlined in Section \ref{ss: experimental setup}. The number of training iterations is set to 10 and 50 epochs. For scikit-ExSTraCS, we use the default hyperparameter values. However, to ensure a fair comparison, we only adjust the hyperparameters common to both UCS and Fuzzy-UCS as follows: $\text{learning\_iterations}=\text{10 or 50 Epochs}$, $N=2000$, $\text{theta\_GA}=50$, $\text{theta\_del}=50$, $\text{theta\_sub}=50$, $\text{theta\_sel}=0.4$, $\text{do\_correct\_set\_subsumption}=\text{True}$, and $\text{rule\_compaction}=\text{None}$.}

\begin{table*}[b]
\begin{center}
\caption{Results When the Number of Training Iterations is 10 Epochs, Displaying Average Classification Accuracy Across 30 Runs.}

\label{tb: sup exstracs 10epoch}
\scalebox{0.55}{
\begin{tabular}{c|c c c cc|ccccc}
\bhline{1pt}
\multicolumn{1}{c|}{\multirow{2}{*}{10 Epochs}} &\multicolumn{5}{c|}{\textsc{Training Accuracy (\%)}} & \multicolumn{5}{c}{\textsc{Test Accuracy (\%)}}\\
& scikit-ExSTraCS & UCS & Fuzzy-UCS$_\text{VOTE}$ & Fuzzy-UCS$_\text{SWIN}$ & Fuzzy-UCS$_\text{DS}$ & scikit-ExSTraCS & UCS & Fuzzy-UCS$_\text{VOTE}$ & Fuzzy-UCS$_\text{SWIN}$ & Fuzzy-UCS$_\text{DS}$\\
\bhline{1pt}
\texttt{bnk} & \cellcolor{g}95.44 $+$ & \cellcolor{p}73.63 $-$ & 89.60 $-$ & 91.93 $+$ & 90.04 & \cellcolor{g}94.93 $+$ & \cellcolor{p}72.87 $-$ & 89.20 $-$ & 91.43 $+$ & 89.73 \\
\texttt{can} & \cellcolor{g}94.30 $\sim$ & 94.06 $\sim$ & \cellcolor{p}93.91 $-$ & 94.00 $\sim$ & 94.03 & 93.51 $\sim$ & \cellcolor{p}90.23 $-$ & 93.51 $\sim$ & \cellcolor{g}93.63 $\sim$ & 93.51 \\
\texttt{car} & 81.11 $-$ & \cellcolor{p}79.04 $-$ & 83.33 $-$ & 83.70 $\sim$ & \cellcolor{g}84.37 & \cellcolor{p}69.33 $\sim$ & 70.00 $\sim$ & 74.00 $\sim$ & \cellcolor{g}77.33 $\sim$ & 75.33 \\
\texttt{col} & \cellcolor{g}76.73 $+$ & \cellcolor{p}64.87 $-$ & 69.33 $+$ & 67.14 $\sim$ & 68.06 & \cellcolor{g}73.55 $+$ & \cellcolor{p}61.61 $\sim$ & 66.56 $\sim$ & 64.30 $\sim$ & 65.27 \\
\texttt{dbt} & \cellcolor{g}75.77 $+$ & \cellcolor{p}69.81 $-$ & 72.06 $-$ & 70.97 $-$ & 72.33 & \cellcolor{g}73.38 $+$ & \cellcolor{p}67.92 $\sim$ & 69.09 $\sim$ & 68.48 $\sim$ & 69.18 \\
\texttt{ecl} & \cellcolor{g}84.61 $+$ & \cellcolor{p}65.14 $-$ & 79.19 $-$ & 75.93 $-$ & 80.72 & \cellcolor{g}83.04 $\sim$ & \cellcolor{p}64.90 $-$ & 79.71 $-$ & 76.76 $-$ & 80.98 \\
\texttt{frt} & 84.09 $-$ & \cellcolor{g}86.06 $\sim$ & 83.02 $-$ & \cellcolor{p}81.29 $-$ & 85.78 & 81.67 $\sim$ & 81.00 $-$ & 81.63 $-$ & \cellcolor{p}79.85 $-$ & \cellcolor{g}83.70 \\
\texttt{gls} & \cellcolor{g}64.93 $+$ & 54.32 $-$ & \cellcolor{p}40.47 $-$ & 61.60 $\sim$ & 61.46 & \cellcolor{g}60.30 $+$ & 48.94 $\sim$ & \cellcolor{p}36.36 $-$ & 52.88 $\sim$ & 52.88 \\
\texttt{hrt} & 84.77 $-$ & \cellcolor{p}83.24 $-$ & 88.25 $-$ & \cellcolor{g}89.58 $\sim$ & 89.51 & \cellcolor{g}83.12 $\sim$ & \cellcolor{p}77.10 $-$ & 81.94 $\sim$ & 80.75 $\sim$ & 82.47 \\
\texttt{hpt} & \cellcolor{p}71.39 $-$ & \cellcolor{g}85.11 $+$ & 76.24 $-$ & 77.51 $\sim$ & 77.10 & \cellcolor{g}67.08 $\sim$ & \cellcolor{p}57.92 $-$ & 65.21 $\sim$ & 64.38 $\sim$ & 65.62 \\
\texttt{hcl} & 75.77 $\sim$ & \cellcolor{g}78.16 $+$ & 74.52 $-$ & \cellcolor{p}73.95 $-$ & 75.35 & \cellcolor{g}73.24 $\sim$ & \cellcolor{p}62.7 $-$ & 68.29 $\sim$ & 67.30 $\sim$ & 68.74 \\
\texttt{irs} & \cellcolor{g}94.72 $\sim$ & \cellcolor{p}77.56 $-$ & 93.68 $-$ & 94.07 $\sim$ & 94.35 & \cellcolor{g}94.22 $\sim$ & \cellcolor{p}74.67 $-$ & 93.56 $\sim$ & \cellcolor{g}94.22 $\sim$ & 93.78 \\
\texttt{lnd} & \cellcolor{g}49.76 $+$ & 37.11 $-$ & \cellcolor{p}27.71 $-$ & 35.35 $-$ & 43.34 & \cellcolor{g}40.78 $+$ & \cellcolor{p}32.16 $-$ & 36.57 $\sim$ & 35.69 $\sim$ & 37.25 \\
\texttt{mam} & \cellcolor{g}82.71 $+$ & 79.79 $-$ & 80.37 $-$ & \cellcolor{p}77.59 $-$ & 80.52 & \cellcolor{g}81.75 $+$ & 79.18 $\sim$ & 78.76 $\sim$ & \cellcolor{p}76.56 $-$ & 79.04 \\
\texttt{pdy} & \cellcolor{g}78.52 $+$ & 72.47 $+$ & 50.43 $-$ & \cellcolor{p}50.04 $-$ & 52.40 & \cellcolor{g}77.91 $+$ & 71.57 $+$ & 50.93 $-$ & \cellcolor{p}50.68 $-$ & 53.06 \\
\texttt{pis} & 85.99 $-$ & 85.56 $-$ & \cellcolor{g}86.66 $\sim$ & \cellcolor{p}85.37 $-$ & 86.64 & 85.46 $\sim$ & 84.42 $-$ & \cellcolor{g}85.89 $\sim$ & \cellcolor{p}84.26 $-$ & 85.71 \\
\texttt{pha} & \cellcolor{p}69.79 $-$ & 76.45 $\sim$ & 76.19 $-$ & \cellcolor{g}78.60 $+$ & 77.40 & \cellcolor{g}65.93 $\sim$ & \cellcolor{p}59.75 $-$ & 65.44 $\sim$ & 62.89 $\sim$ & 64.85 \\
\texttt{pre} & \cellcolor{p}88.70 $-$ & 99.33 $\sim$ & 99.64 $\sim$ & \cellcolor{g}100.0 $+$ & 99.61 & \cellcolor{p}88.87 $-$ & 99.34 $\sim$ & 99.67 $\sim$ & \cellcolor{g}100.0 $+$ & 99.60 \\
\texttt{pmp} & \cellcolor{g}87.12 $+$ & \cellcolor{p}86.45 $\sim$ & 86.88 $\sim$ & \cellcolor{p}86.45 $-$ & 86.84 & \cellcolor{g}86.29 $\sim$ & \cellcolor{p}85.35 $\sim$ & 86.19 $\sim$ & 85.88 $-$ & 86.24 \\
\texttt{rsn} & \cellcolor{g}85.44 $+$ & \cellcolor{p}80.33 $-$ & 83.71 $-$ & 85.16 $+$ & 84.38 & \cellcolor{g}85.11 $\sim$ & \cellcolor{p}81.74 $-$ & 83.37 $-$ & \cellcolor{g}85.11 $+$ & 84.37 \\
\texttt{seg} & \cellcolor{g}91.12 $+$ & \cellcolor{p}87.61 $-$ & 88.93 $-$ & 89.42 $\sim$ & 89.26 & \cellcolor{g}90.81 $+$ & \cellcolor{p}87.30 $-$ & 88.63 $-$ & 89.16 $\sim$ & 89.02 \\
\texttt{sir} & 82.07 $\sim$ & \cellcolor{p}71.81 $-$ & 83.59 $\sim$ & \cellcolor{g}83.96 $\sim$ & 83.59 & 80.67 $\sim$ & \cellcolor{p}69.67 $-$ & \cellcolor{g}81.00 $\sim$ & 79.33 $\sim$ & \cellcolor{g}81.00 \\
\texttt{smk} & \cellcolor{g}96.14 $+$ & 87.02 $+$ & 79.20 $+$ & 86.05 $+$ & \cellcolor{p}76.38 & \cellcolor{g}95.99 $+$ & 86.05 $+$ & 78.32 $+$ & 85.47 $+$ & \cellcolor{p}75.76 \\
\texttt{tae} & 59.93 $\sim$ & \cellcolor{p}53.75 $-$ & 57.43 $-$ & 60.64 $\sim$ & \cellcolor{g}61.48 & 51.25 $\sim$ & \cellcolor{p}45.62 $-$ & 49.79 $\sim$ & 52.08 $\sim$ & \cellcolor{g}53.75 \\
\texttt{tip} & \cellcolor{p}76.23 $-$ & 80.01 $\sim$ & 80.34 $\sim$ & 79.11 $-$ & \cellcolor{g}80.40 & \cellcolor{p}74.86 $-$ & 78.69 $-$ & \cellcolor{g}79.66 $\sim$ & 77.97 $-$ & 79.61 \\
\texttt{tit} & \cellcolor{g}72.05 $+$ & 70.05 $+$ & 69.28 $\sim$ & 70.22 $+$ & \cellcolor{p}69.22 & \cellcolor{g}71.30 $+$ & 70.30 $\sim$ & 68.89 $\sim$ & 69.07 $\sim$ & \cellcolor{p}68.56 \\
\texttt{wne} & 95.77 $-$ & \cellcolor{p}87.65 $-$ & 97.12 $-$ & 97.48 $\sim$ & \cellcolor{g}97.56 & 93.33 $-$ & \cellcolor{p}84.44 $-$ & 95.56 $\sim$ & 94.63 $\sim$ & \cellcolor{g}95.93 \\
\texttt{wbc} & \cellcolor{g}97.63 $+$ & 95.79 $-$ & 96.47 $-$ & \cellcolor{p}95.73 $-$ & 96.63 & \cellcolor{g}96.95 $+$ & 94.90 $-$ & 96.24 $\sim$ & \cellcolor{p}94.38 $-$ & 95.95 \\
\texttt{wpb} & \cellcolor{p}71.44 $-$ & 86.03 $-$ & 86.89 $-$ & \cellcolor{g}91.03 $+$ & 89.33 & \cellcolor{p}61.83 $-$ & 68.33 $-$ & \cellcolor{g}75.17 $\sim$ & 73.17 $\sim$ & \cellcolor{g}75.17 \\
\texttt{yst} & \cellcolor{g}59.91 $+$ & \cellcolor{p}45.25 $-$ & 45.91 $-$ & 51.66 $-$ & 54.14 & \cellcolor{g}58.84 $+$ & \cellcolor{p}42.10 $-$ & 44.38 $-$ & 50.74 $-$ & 53.58 \\
\bhline{1pt}
Rank & \cellcolor{g}\textit{2.30}$\uparrow$ & \cellcolor{p}\textit{3.82}$\downarrow^{\dag\dag}$ & \textit{3.42}$\downarrow^{\dag\dag}$ & \textit{3.02}$\downarrow$ & \textit{2.45} & \cellcolor{g}\textit{1.97}$\uparrow$ & \cellcolor{p}\textit{4.30}$\downarrow^{\dag\dag}$ & \textit{3.00}$\downarrow^{\dag\dag}$ & \textit{3.28}$\downarrow^{\dag\dag}$ & \textit{2.45} \\
Position & \textit{1} & \textit{5} & \textit{4} & \textit{3} & \textit{2} & \textit{1} & \textit{5} & \textit{3} & \textit{4} & \textit{2} \\
$+/-/\sim$ & 15/10/5 & 5/19/6 & 2/22/6 & 7/12/11 & - & 12/4/14 & 2/20/8 & 1/8/21 & 4/9/17 & - \\
\bhline{1pt}
$p$-value & 0.503 & 0.00683 & 0.000479 & 0.120 & - & 0.0920 & 0.000209 & 0.0143 & 0.00917 & - \\
$p_\text{Holm}$-value & 0.503 & 0.0205 & 0.00192 & 0.241 & - & 0.0920 & 0.000837 & 0.0285 & 0.0275 & - \\

\bhline{1pt}

\end{tabular}
}
\end{center}
\end{table*}

\begin{table*}[t]
\begin{center}
\caption{Results When the Number of Training Iterations is 50 Epochs, Displaying Average Classification Accuracy Across 30 Runs.}

\label{tb: sup exstracs 50epoch}
\scalebox{0.55}{
\begin{tabular}{c|c c c cc|ccccc}
\bhline{1pt}
\multicolumn{1}{c|}{\multirow{2}{*}{50 Epochs}} &\multicolumn{5}{c|}{\textsc{Training Accuracy (\%)}} & \multicolumn{5}{c}{\textsc{Test Accuracy (\%)}}\\
& scikit-ExSTraCS & UCS & Fuzzy-UCS$_\text{VOTE}$ & Fuzzy-UCS$_\text{SWIN}$ & Fuzzy-UCS$_\text{DS}$ & scikit-ExSTraCS & UCS & Fuzzy-UCS$_\text{VOTE}$ & Fuzzy-UCS$_\text{SWIN}$ & Fuzzy-UCS$_\text{DS}$\\
\bhline{1pt}
\texttt{bnk} & \cellcolor{g}97.26 $+$ & 94.93 $+$ & \cellcolor{p}92.63 $-$ & 93.65 $+$ & 92.95 & \cellcolor{g}97.08 $+$ & 94.47 $+$ & \cellcolor{p}92.34 $-$ & 93.00 $\sim$ & 92.63 \\
\texttt{can} & 95.48 $\sim$ & \cellcolor{g}97.19 $+$ & 95.29 $-$ & \cellcolor{p}94.83 $-$ & 95.39 & 94.68 $\sim$ & \cellcolor{p}91.75 $-$ & 94.74 $\sim$ & 94.44 $\sim$ & \cellcolor{g}94.80 \\
\texttt{car} & 86.15 $-$ & \cellcolor{p}81.56 $-$ & 87.93 $\sim$ & 88.07 $\sim$ & \cellcolor{g}88.22 & 75.33 $\sim$ & \cellcolor{p}71.33 $\sim$ & 76.00 $\sim$ & \cellcolor{g}76.67 $\sim$ & 74.00 \\
\texttt{col} & \cellcolor{g}80.91 $+$ & 76.01 $+$ & 71.85 $+$ & \cellcolor{p}68.24 $-$ & 71.37 & \cellcolor{g}77.31 $+$ & 70.00 $\sim$ & 69.35 $\sim$ & \cellcolor{p}65.81 $-$ & 69.03 \\
\texttt{dbt} & 78.09 $+$ & \cellcolor{g}78.12 $+$ & 76.36 $\sim$ & \cellcolor{p}73.57 $-$ & 76.48 & \cellcolor{g}74.68 $\sim$ & 72.90 $\sim$ & 73.07 $\sim$ & \cellcolor{p}70.04 $-$ & 72.94 \\
\texttt{ecl} & \cellcolor{g}89.58 $+$ & \cellcolor{p}78.55 $-$ & 85.79 $\sim$ & 80.39 $-$ & 85.64 & \cellcolor{g}86.47 $+$ & \cellcolor{p}76.18 $-$ & 84.02 $\sim$ & 79.80 $-$ & 83.92 \\
\texttt{frt} & 90.58 $+$ & \cellcolor{g}93.67 $+$ & 87.12 $-$ & \cellcolor{p}85.17 $-$ & 88.43 & \cellcolor{g}87.52 $\sim$ & 83.41 $-$ & 84.74 $-$ & \cellcolor{p}82.33 $-$ & 86.44 \\
\texttt{gls} & \cellcolor{g}76.96 $+$ & \cellcolor{p}65.19 $-$ & 72.01 $\sim$ & 71.75 $\sim$ & 72.14 & \cellcolor{g}65.30 $\sim$ & \cellcolor{p}53.33 $-$ & 62.73 $\sim$ & 64.09 $\sim$ & 64.39 \\
\texttt{hrt} & \cellcolor{p}87.98 $-$ & 90.53 $-$ & 94.66 $-$ & 94.35 $-$ & \cellcolor{g}95.05 & \cellcolor{g}83.98 $\sim$ & \cellcolor{p}81.08 $\sim$ & 81.83 $\sim$ & 82.69 $\sim$ & 81.83 \\
\texttt{hpt} & \cellcolor{p}75.56 $-$ & 90.10 $-$ & 91.46 $-$ & 89.88 $-$ & \cellcolor{g}92.33 & \cellcolor{g}68.12 $\sim$ & \cellcolor{p}57.08 $-$ & 64.58 $\sim$ & 62.08 $\sim$ & 65.62 \\
\texttt{hcl} & \cellcolor{p}81.85 $-$ & \cellcolor{g}89.86 $+$ & 87.25 $-$ & 84.19 $-$ & 87.76 & \cellcolor{g}76.22 $+$ & \cellcolor{p}63.69 $-$ & 71.80 $\sim$ & 69.37 $\sim$ & 71.62 \\
\texttt{irs} & \cellcolor{g}96.57 $+$ & \cellcolor{p}84.17 $-$ & 94.81 $-$ & 95.51 $\sim$ & 95.26 & 94.00 $\sim$ & \cellcolor{p}81.78 $-$ & 94.00 $\sim$ & \cellcolor{g}94.89 $\sim$ & 94.22 \\
\texttt{lnd} & \cellcolor{g}61.85 $+$ & 55.29 $+$ & \cellcolor{p}32.05 $-$ & 39.18 $-$ & 46.27 & \cellcolor{g}47.94 $+$ & 42.06 $\sim$ & \cellcolor{p}38.24 $\sim$ & 40.10 $\sim$ & 40.20 \\
\texttt{mam} & 83.00 $+$ & \cellcolor{g}83.10 $+$ & 80.89 $-$ & \cellcolor{p}77.89 $-$ & 81.07 & \cellcolor{g}81.75 $+$ & 80.72 $\sim$ & 79.73 $\sim$ & \cellcolor{p}76.77 $-$ & 80.03 \\
\texttt{pdy} & 80.14 $+$ & \cellcolor{g}89.16 $+$ & 50.94 $-$ & \cellcolor{p}50.70 $-$ & 53.94 & 79.51 $+$ & \cellcolor{g}88.24 $+$ & 51.46 $-$ & \cellcolor{p}51.36 $-$ & 54.63 \\
\texttt{pis} & 86.69 $\sim$ & \cellcolor{g}87.54 $+$ & 86.89 $\sim$ & \cellcolor{p}85.53 $-$ & 86.91 & 85.91 $\sim$ & \cellcolor{g}86.12 $\sim$ & 85.88 $\sim$ & \cellcolor{p}84.67 $-$ & 85.92 \\
\texttt{pha} & \cellcolor{p}76.79 $-$ & 79.17 $-$ & 82.10 $-$ & \cellcolor{g}86.80 $+$ & 85.64 & 65.83 $\sim$ & \cellcolor{p}62.55 $-$ & \cellcolor{g}66.96 $+$ & 64.41 $\sim$ & 66.13 \\
\texttt{pre} & \cellcolor{p}88.61 $-$ & 99.63 $\sim$ & 98.01 $-$ & \cellcolor{g}100.0 $+$ & 99.76 & \cellcolor{p}89.00 $-$ & 99.64 $\sim$ & 98.03 $-$ & \cellcolor{g}100.0 $+$ & 99.79 \\
\texttt{pmp} & \cellcolor{g}87.69 $+$ & 87.24 $\sim$ & 86.99 $\sim$ & \cellcolor{p}86.89 $-$ & 87.03 & \cellcolor{g}86.84 $+$ & \cellcolor{p}86.01 $\sim$ & 86.17 $\sim$ & 86.05 $\sim$ & 86.21 \\
\texttt{rsn} & \cellcolor{g}86.15 $+$ & 85.02 $\sim$ & \cellcolor{p}84.33 $-$ & 85.51 $+$ & 85.13 & \cellcolor{g}86.00 $+$ & 84.63 $\sim$ & \cellcolor{p}84.19 $-$ & 85.30 $\sim$ & 85.07 \\
\texttt{seg} & 94.34 $+$ & \cellcolor{g}95.40 $+$ & \cellcolor{p}90.10$-$ & 90.52 $\sim$ & 90.42 & \cellcolor{g}93.82 $+$ & 93.02 $+$ & \cellcolor{p}89.88 $-$ & 90.01 $\sim$ & 90.30 \\
\texttt{sir} & \cellcolor{g}87.22 $+$ & \cellcolor{p}76.96 $-$ & 84.78 $\sim$ & 85.00 $\sim$ & 84.78 & \cellcolor{g}82.33 $\sim$ & \cellcolor{p}70.67 $-$ & 81.67 $\sim$ & 81.33 $\sim$ & 82.00 \\
\texttt{smk} & \cellcolor{g}98.67 $+$ & 97.66 $+$ & 95.39 $-$ & \cellcolor{p}93.49 $-$ & 95.71 & \cellcolor{g}98.87 $+$ & 96.5 $+$ & 95.08 $\sim$ & \cellcolor{p}93.14 $-$ & 95.11 \\
\texttt{tae} & \cellcolor{g}67.83 $+$ & \cellcolor{p}59.23 $-$ & 63.63 $-$ & 64.15 $\sim$ & 64.35 & 52.29 $\sim$ & \cellcolor{p}49.17 $-$ & 54.58 $\sim$ & 53.75 $\sim$ & \cellcolor{g}56.46 \\
\texttt{tip} & 80.39 $-$ & \cellcolor{g}82.71 $+$ & 81.09 $-$ & \cellcolor{p}80.24 $-$ & 81.20 & \cellcolor{p}77.87 $-$ & \cellcolor{g}80.47 $\sim$ & 80.00 $-$ & 78.88 $-$ & 80.23 \\
\texttt{tit} & \cellcolor{g}74.04 $+$ & 73.25 $+$ & 72.28 $\sim$ & \cellcolor{p}70.57 $-$ & 72.38 & \cellcolor{g}72.56 $\sim$ & 72.30 $\sim$ & 71.30 $\sim$ & \cellcolor{p}68.93 $-$ & 71.52 \\
\texttt{wne} & 98.40 $-$ & \cellcolor{p}93.62 $-$ & 98.31 $-$ & 98.58 $\sim$ & \cellcolor{g}98.79 & 95.19 $\sim$ & \cellcolor{p}88.52 $-$ & 96.48 $\sim$ & 96.11 $\sim$ & \cellcolor{g}96.67 \\
\texttt{wbc} & \cellcolor{g}97.90 $+$ & 97.59 $+$ & 97.12 $\sim$ & \cellcolor{p}96.42 $-$ & 97.07 & \cellcolor{g}96.57 $\sim$ & 96.19 $\sim$ & 96.19 $\sim$ & \cellcolor{p}95.29 $\sim$ & 95.95 \\
\texttt{wpb} & \cellcolor{p}77.13 $-$ & \cellcolor{g}95.21 $+$ & 91.57 $-$ & 93.60 $+$ & 92.58 & \cellcolor{p}64.83 $-$ & 70.00 $\sim$ & \cellcolor{g}72.17 $\sim$ & 71.67 $\sim$ & 72.00 \\
\texttt{yst} & \cellcolor{g}63.19 $+$ & 60.73 $+$ & \cellcolor{p}53.26 $-$ & 56.22 $-$ & 58.78 & \cellcolor{g}59.66 $+$ & 56.89 $\sim$ & \cellcolor{p}52.04 $-$ & 53.65 $-$ & 57.61 \\
\bhline{1pt}
Rank & \cellcolor{g}\textit{2.37}$\uparrow$ & \textit{2.63}$\uparrow$ & \cellcolor{p}\textit{3.68}$\downarrow^{\dag\dag}$ & \textit{3.67}$\downarrow^{\dag\dag}$ & \textit{2.65} & \cellcolor{g}\textit{1.98}$\uparrow^{\dag\dag}$ & \textit{3.55}$\downarrow^{\dag\dag}$ & \textit{3.22}$\downarrow^{\dag\dag}$ & \cellcolor{p}\textit{3.70}$\downarrow^{\dag\dag}$ & \textit{2.55} \\
Position & \textit{1} & \textit{2} & \textit{5} & \textit{4} & \textit{3} & \textit{1} & \textit{4} & \textit{3} & \textit{5} & \textit{2} \\
$+/-/\sim$ & 19/9/2 & 17/10/3 & 1/20/9 & 5/18/7 & - & 12/3/15 & 4/11/15 & 1/8/21 & 1/11/18 & - \\
\bhline{1pt}
$p$-value & 0.229 & 0.968 & 4.80E-05 & 0.00148 & - & 0.0185 & 0.0483 & 0.0121 & 0.000529 & - \\
$p_\text{Holm}$-value & 0.457 & 0.968 & 0.000192 & 0.00445 & - & 0.0371 & 0.0483 & 0.0364 & 0.00211 & - \\

\bhline{1pt}

\end{tabular}
}
\end{center}
\end{table*}

         \begin{table*}[t]
\begin{center}
\caption{Results Displaying Average Macro-Ruleset Size Across 30 Runs.}
\vspace{2mm}
\label{tb: sup vs exstracs}
\scalebox{0.7}{
\begin{tabular}{c|ccc|ccc}
\bhline{1pt}
&\multicolumn{6}{c}{\textsc{Macro-Ruleset Size}} \\
&\multicolumn{3}{c|}{10 Epochs} & \multicolumn{3}{c}{50 Epochs} \\
& scikit-ExSTraCS & UCS & \all & scikit-ExSTraCS & UCS & \all\\
\bhline{1pt}
\texttt{bnk} & \cellcolor{p}1335 $-$ & 811.0 $-$ & \cellcolor{g}642.9 & \cellcolor{p}1536 $-$ & 1285 $-$ & \cellcolor{g}1244 \\
\texttt{can} & \cellcolor{g}579.3 $+$ & \cellcolor{p}1609 $-$ & 1081 & \cellcolor{g}1702 $+$ & \cellcolor{p}1750 $\sim$ & 1744 \\
\texttt{car} & 47.87 $+$ & \cellcolor{g}44.57 $+$ & \cellcolor{p}70.83 & \cellcolor{g}94.60 $+$ & 218.8 $+$ & \cellcolor{p}257.7 \\
\texttt{col} & \cellcolor{p}432.3 $-$ & \cellcolor{g}256.9 $\sim$ & 274.4 & \cellcolor{p}1670 $-$ & 1275 $-$ & \cellcolor{g}1084 \\
\texttt{dbt} & \cellcolor{p}1247 $-$ & 942.4 $\sim$ & \cellcolor{g}894.2 & \cellcolor{p}1790 $-$ & 1584 $-$ & \cellcolor{g}1546 \\
\texttt{ecl} & \cellcolor{p}746.3 $-$ & \cellcolor{g}461.8 $\sim$ & 485.7 & \cellcolor{p}1694 $-$ & 1467 $-$ & \cellcolor{g}1357 \\
\texttt{frt} & \cellcolor{g}1796 $+$ & \cellcolor{p}1817 $\sim$ & 1811 & \cellcolor{g}1824 $+$ & 1826 $+$ & \cellcolor{p}1858 \\
\texttt{gls} & \cellcolor{p}589.4 $-$ & \cellcolor{g}374.5 $+$ & 543.0 & \cellcolor{p}1826 $-$ & 1430 $-$ & \cellcolor{g}1313 \\
\texttt{hrt} & \cellcolor{g}453.6 $+$ & 968.5 $+$ & \cellcolor{p}1429 & \cellcolor{p}1667 $-$ & 1622 $\sim$ & \cellcolor{g}1620 \\
\texttt{hpt} & \cellcolor{g}213.9 $+$ & 920.2 $+$ & \cellcolor{p}1019 & \cellcolor{g}1228 $+$ & 1683 $\sim$ & \cellcolor{p}1690 \\
\texttt{hcl} & \cellcolor{g}392.2 $+$ & \cellcolor{p}1648 $\sim$ & 1646 & \cellcolor{g}1561 $+$ & 1777 $\sim$ & \cellcolor{p}1786 \\
\texttt{irs} & \cellcolor{p}152.9 $-$ & \cellcolor{g}78.50 $+$ & 88.50 & \cellcolor{p}574.1 $-$ & 477.2 $-$ & \cellcolor{g}357.4 \\
\texttt{lnd} & \cellcolor{p}941.9 $-$ & 411.1 $-$ & \cellcolor{g}349.9 & \cellcolor{p}1579 $-$ & 1374 $-$ & \cellcolor{g}723.0 \\
\texttt{mam} & \cellcolor{p}1117 $-$ & 832.0 $-$ & \cellcolor{g}705.5 & \cellcolor{g}1294 $+$ & \cellcolor{p}1478 $-$ & 1412 \\
\texttt{pdy} & \cellcolor{p}1634 $-$ & 1420 $-$ & \cellcolor{g}846.8 & \cellcolor{p}1581 $-$ & 1403 $-$ & \cellcolor{g}912.4 \\
\texttt{pis} & \cellcolor{p}1777 $-$ & 1653 $-$ & \cellcolor{g}1629 & \cellcolor{p}1727 $-$ & \cellcolor{g}1668 $+$ & 1704 \\
\texttt{pha} & \cellcolor{g}1114 $+$ & 1604 $+$ & \cellcolor{p}1678 & \cellcolor{p}1772 $-$ & \cellcolor{g}1692 $+$ & 1725 \\
\texttt{pre} & \cellcolor{p}1687 $-$ & 1194 $-$ & \cellcolor{g}870.1 & \cellcolor{p}1723 $-$ & \cellcolor{g}958.1 $\sim$ & 990.4 \\
\texttt{pmp} & \cellcolor{p}1751 $-$ & 1608 $-$ & \cellcolor{g}1544 & \cellcolor{p}1682 $\sim$ & \cellcolor{g}1630 $+$ & 1669 \\
\texttt{rsn} & \cellcolor{p}883.0 $-$ & 722.5 $-$ & \cellcolor{g}540.3 & \cellcolor{p}1601 $-$ & 1512 $-$ & \cellcolor{g}1464 \\
\texttt{seg} & \cellcolor{p}1783 $-$ & \cellcolor{g}1693 $+$ & 1712 & \cellcolor{p}1736 $-$ & \cellcolor{g}1701 $\sim$ & 1709 \\
\texttt{sir} & \cellcolor{p}148.6 $-$ & \cellcolor{g}77.37 $+$ & 120.9 & \cellcolor{p}586.3 $-$ & 444.0 $\sim$ & \cellcolor{g}441.1 \\
\texttt{smk} & \cellcolor{g}503.7 $+$ & \cellcolor{p}748.4 $-$ & 618.0 & 1465 $\sim$ & \cellcolor{g}1425 $+$ & \cellcolor{p}1482 \\
\texttt{tae} & \cellcolor{p}302.5 $-$ & \cellcolor{g}188.6 $+$ & 264.2 & \cellcolor{p}1521 $-$ & 1073 $-$ & \cellcolor{g}907.5 \\
\texttt{tip} & \cellcolor{p}1679 $-$ & 1633 $\sim$ & \cellcolor{g}1621 & \cellcolor{p}1665 $-$ & \cellcolor{g}1569 $+$ & 1615 \\
\texttt{tit} & 1112 $+$ & \cellcolor{g}826.8 $+$ & \cellcolor{p}1233 & \cellcolor{g}1521 $\sim$ & \cellcolor{p}1538 $\sim$ & \cellcolor{g}1521 \\
\texttt{wne} & \cellcolor{g}323.1 $+$ & \cellcolor{p}355.7 $\sim$ & 349.8 & \cellcolor{p}1623 $-$ & 1396 $-$ & \cellcolor{g}1340 \\
\texttt{wbc} & \cellcolor{g}738.7 $+$ & \cellcolor{p}1223 $-$ & 1171 & \cellcolor{g}1036 $+$ & \cellcolor{p}1538 $-$ & 1527 \\
\texttt{wpb} & \cellcolor{g}271.9 $+$ & \cellcolor{p}1502 $-$ & 1021 & \cellcolor{g}1655 $+$ & \cellcolor{p}1826 $-$ & 1782 \\
\texttt{yst} & \cellcolor{p}1873 $-$ & 1586 $-$ & \cellcolor{g}1264 & \cellcolor{p}1832 $-$ & 1632 $-$ & \cellcolor{g}1548 \\
\bhline{1pt}
Rank & \cellcolor{p}\textit{2.27}$\downarrow$ & \textit{1.93}$\downarrow$ & \cellcolor{g}\textit{1.80} & \cellcolor{p}\textit{2.38}$\downarrow^{\dag\dag}$ & \textit{1.93}$\downarrow^{\dag\dag}$ & \cellcolor{g}\textit{1.68} \\
Position & \textit{3} & \textit{2} & \textit{1} & \textit{3} & \textit{2} & \textit{1} \\
$+/-/\sim$ & 12/18/0 & 10/13/7 & - & 8/19/3 & 7/15/8 & - \\
\bhline{1pt}
$p$-value & 0.490 & 0.221 & - & 0.0195 & 0.0262 & - \\
$p_\text{Holm}$-value & 0.490 & 0.441 & - & 0.0390 & 0.0390 & - \\

\bhline{1pt}

\end{tabular}
}
\end{center}
\end{table*}

{Tables \ref{tb: sup exstracs 10epoch} and \ref{tb: sup exstracs 50epoch} present each system's average training and test classification accuracy and average rank across all datasets at the 10th and 50th epochs, respectively.
Green-shaded values denote the best values among all systems, while peach-shaded values indicate the worst values among all systems. The terms ``Rank'' and ``Position'' denote each system's overall average rank obtained by using the Friedman test and its position in the final ranking, respectively. Statistical results of the Wilcoxon signed-rank test are summarized with symbols wherein ``$+$'', ``$-$'', and ``$\sim$'' represent that the classification accuracy of a conventional system is significantly better, worse, and competitive compared to that obtained by the proposed Fuzzy-UCS$_\text{DS}$, respectively. The ``$p$-value'' and ``$p_\text{Holm}$-value'' are derived from the Wilcoxon signed-rank test and the Holm-adjusted Wilcoxon signed-rank test, respectively. Arrows denote whether the rank improved or declined compared to Fuzzy-UCS$_\text{DS}$. $\dag$ ($\dag\dag$) indicates statistically significant differences compared to Fuzzy-UCS$_\text{DS}$, i.e., $p$-value ($p_\text{Holm}$-value) $<\alpha=0.05$. 

From Tables \ref{tb: sup exstracs 10epoch} and \ref{tb: sup exstracs 50epoch}, scikit-ExSTraCS recorded the highest rank in test accuracy at both the 10th and 50th epochs. Particularly at the 50th epoch, \ds\ performed significantly worse than scikit-ExSTraCS ($p_{\text{Holm}}=0.0371$). However, it is crucial to interpret these results in the context of the fundamental differences between the two systems:

\begin{itemize}
\item Fuzzy-UCS is a ``fuzzified'' version of UCS, designed as a simple and fundamental system.
\item In contrast, scikit-ExSTraCS incorporates several specialized mechanisms specifically designed for noisy datasets, including:
\begin{itemize}
\item Attribute Feedback: Guides rule generalization based on attribute tracking scores.
\item Attribute Tracking: Tracks attribute importance for each training instance.
\item Mixed Discrete-Continuous Attribute List Knowledge Representation: Efficiently handles both discrete and continuous attributes compared to linguistic terms.
\end{itemize}
\end{itemize}

Given these architectural differences, the performance gap is not unexpected and highlights the trade-offs between simplicity and specialized optimization. 

\label{r1-2-1}
{Additionally, Table \ref{tb: sup vs exstracs} shows UCS, \all, and scikit-ExSTraCS's average macro-ruleset size and average rank across all datasets at the 10th and 50th epochs, respectively.
Green-shaded values denote the best (smallest) values among all systems, while peach-shaded values indicate the worst (largest) values among all systems. Statistical results of the Wilcoxon signed-rank test are summarized with symbols wherein ``$+$'', ``$-$'', and ``$\sim$'' represent that the macro-ruleset size of a conventional system is significantly smaller, larger, and competitive compared to that obtained by \all, respectively. Arrows denote whether the rank improved or declined compared to \all. $\dag$ ($\dag\dag$) indicates statistically significant differences compared to \all, i.e., $p$-value ($p_\text{Holm}$-value) $<\alpha=0.05$. ``Rank'' and ``Position'' should be interpreted as in Tables \ref{tb: sup exstracs 10epoch} and \ref{tb: sup exstracs 50epoch}.

Table \ref{tb: sup vs exstracs} reveals that \all\ achieved the highest average rank (best among all methods) in terms of ruleset size at both the 10th and 50th epochs, while scikit-ExSTraCS recorded the third-best rank (worst among all methods). Notably, at the 50th epoch, \all\ generated significantly fewer rules compared to both UCS and scikit-ExSTraCS ($p_\text{Holm}=0.0390$). This finding aligns with previous research \cite{orriols2008fuzzy}, which suggests that the linguistic rule representation inherent in fuzzy rules (used in \all) promotes better rule generalization compared to non-linguistic rules (used in UCS and scikit-ExSTraCS). Specifically, the use of linguistic terms in fuzzy rules enhances the system's ability to capture and represent knowledge in a more general form.

Given that scikit-ExSTraCS achieved higher test accuracy than \all\ at both the 10th and 50th epochs, these findings demonstrate an important trade-off: while scikit-ExSTraCS achieves higher test accuracy, it does so with more complex non-linguistic rule representations and a larger number of rules compared to \all's interpretable linguistic rules. These results highlight the trade-off between complexity and classification performance.
}

For future research directions, we are considering several ways:
\begin{enumerate}
    \item Enhancing \ds\ by selectively incorporating some of scikit-ExSTraCS's noise-handling mechanisms while maintaining its core simplicity and interoperability.
    \item Exploring the integration of fuzzy rule representations into scikit-ExSTraCS or adapting its inference scheme to incorporate the DS theory, potentially leading to a powerful hybrid approach that combines the strengths of both \ds\ and scikit-ExSTraCS.
    \item A comprehensive review and improvement of the entire Fuzzy-UCS framework. While our current research focused primarily on the decision-making mechanism (i.e., test phase), we recognize that there is significant room for improvement in the rule production mechanism (i.e., training phase).
\end{enumerate}

In conclusion, while our current approach demonstrates the effectiveness of incorporating DS theory into the decision-making process of Fuzzy-UCS, there is still considerable room for improvement in other aspects of the system. Future work will aim to address these challenges, potentially leading to a more robust and versatile LFCS framework capable of handling complex, noisy datasets while maintaining high interpretability.

}

\clearpage
\section{Comparison With State-of-the-Art Classification Methods}
\label{sec: sup comparison with state-of-the-art classification methods}
{
To further evaluate our proposed algorithm's performance against more advanced classification methods, we conduct additional experiments using Random Forest \cite{breiman2001random} and XGBoost \cite{chen2016xgboost}. These systems are chosen due to their widespread recognition as state-of-the-art methods for handling complex datasets effectively.

The experimental setup follows the conditions outlined in Section \ref{sec: experiment}. We use standard implementations of Random Forest\footnote{\url{https://scikit-learn.org/dev/modules/generated/sklearn.ensemble.RandomForestClassifier.html} (\today)} and XGBoost\footnote{\url{https://xgboost.readthedocs.io/en/stable/python/index.html} (\today)}, ensuring consistent dataset splits for fair comparison. For Random Forest and XGBoost, we use default hyperparameter settings. For UCS and \all, the number of training iterations is set to 50 epochs.}

\begin{table*}[b] 
\begin{center}
\caption{Results Displaying Average Classification Accuracy Across 30 Runs.}
\label{tb: sup sota}
\scalebox{0.49}{
\begin{tabular}{c|cccccc|cccccc}
\bhline{1pt}
\multicolumn{1}{c|}{\multirow{2}{*}{}} &\multicolumn{6}{c}{\textsc{Training Accuracy (\%)}} & \multicolumn{6}{c}{\textsc{Test Accuracy (\%)}}\\
& Random Forest & XGBoost & UCS & Fuzzy-UCS$_\text{VOTE}$ & Fuzzy-UCS$_\text{SWIN}$ & Fuzzy-UCS$_\text{DS}$ & Random Forest & XGBoost & UCS & Fuzzy-UCS$_\text{VOTE}$ & Fuzzy-UCS$_\text{SWIN}$ & Fuzzy-UCS$_\text{DS}$\\
\bhline{1pt}
\texttt{bnk} & \cellcolor{g}100.0 $+$ & \cellcolor{g}100.0 $+$ & 94.93 $+$ & \cellcolor{p}92.63 $-$ & 93.65 $+$ & 92.95 & \cellcolor{g}99.47 $+$ & 99.28 $+$ & 94.47 $+$ & \cellcolor{p}92.34 $-$ & 93.00 $\sim$ & 92.63 \\
\texttt{can} & \cellcolor{g}100.0 $+$ & \cellcolor{g}100.0 $+$ & 97.19 $+$ & 95.29 $-$ & \cellcolor{p}94.83 $-$ & 95.39 & 95.85 $\sim$ & \cellcolor{g}96.67 $+$ & \cellcolor{p}91.75 $-$ & 94.74 $\sim$ & 94.44 $\sim$ & 94.80 \\
\texttt{car} & \cellcolor{g}90.22 $+$ & 90.15 $+$ & \cellcolor{p}81.56 $-$ & 87.93 $\sim$ & 88.07 $\sim$ & 88.22 & \cellcolor{g}77.33 $\sim$ & 75.33 $\sim$ & \cellcolor{p}71.33 $\sim$ & 76.00 $\sim$ & 76.67 $\sim$ & 74.00 \\
\texttt{col} & \cellcolor{g}100.0 $+$ & \cellcolor{g}100.0 $+$ & 76.01 $+$ & 71.85 $+$ & \cellcolor{p}68.24 $-$ & 71.37 & \cellcolor{g}84.62 $+$ & \cellcolor{g}84.62 $+$ & 70.00 $\sim$ & 69.35 $\sim$ & \cellcolor{p}65.81 $-$ & 69.03 \\
\texttt{dbt} & \cellcolor{g}100.0 $+$ & \cellcolor{g}100.0 $+$ & 78.12 $+$ & 76.36 $\sim$ & \cellcolor{p}73.57 $-$ & 76.48 & \cellcolor{g}75.71 $+$ & 72.73 $\sim$ & 72.90 $\sim$ & 73.07 $\sim$ & \cellcolor{p}70.04 $-$ & 72.94 \\
\texttt{ecl} & \cellcolor{g}100.0 $+$ & \cellcolor{g}100.0 $+$ & \cellcolor{p}78.55 $-$ & 85.79 $\sim$ & 80.39 $-$ & 85.64 & \cellcolor{g}89.22 $+$ & 86.47 $+$ & \cellcolor{p}76.18 $-$ & 84.02 $\sim$ & 79.80 $-$ & 83.92 \\
\texttt{frt} & \cellcolor{g}100.0 $+$ & \cellcolor{g}100.0 $+$ & 93.67 $+$ & 87.12 $-$ & \cellcolor{p}85.17 $-$ & 88.43 & 90.04 $+$ & \cellcolor{g}91.37 $+$ & 83.41 $-$ & 84.74 $-$ & \cellcolor{p}82.33 $-$ & 86.44 \\
\texttt{gls} & \cellcolor{g}100.0 $+$ & \cellcolor{g}100.0 $+$ & \cellcolor{p}65.19 $-$ & 72.01 $\sim$ & 71.75 $\sim$ & 72.14 & \cellcolor{g}79.55 $+$ & 77.88 $+$ & \cellcolor{p}53.33 $-$ & 62.73 $\sim$ & 64.09 $\sim$ & 64.39 \\
\texttt{hrt} & \cellcolor{g}100.0 $+$ & \cellcolor{g}100.0 $+$ & \cellcolor{p}90.53 $-$ & 94.66 $-$ & 94.35 $-$ & 95.05 & \cellcolor{g}83.55 $\sim$ & \cellcolor{p}80.32 $\sim$ & 81.08 $\sim$ & 81.83 $\sim$ & 82.69 $\sim$ & 81.83 \\
\texttt{hpt} & 99.98 $+$ & \cellcolor{g}100.0 $+$ & 90.10 $-$ & 91.46 $-$ & \cellcolor{p}89.88 $-$ & 92.33 & \cellcolor{g}67.50 $\sim$ & 60.62 $\sim$ & \cellcolor{p}57.08 $-$ & 64.58 $\sim$ & 62.08 $\sim$ & 65.62 \\
\texttt{hcl} & \cellcolor{g}100.0 $+$ & \cellcolor{g}100.0 $+$ & 89.86 $+$ & 87.25 $-$ & \cellcolor{p}84.19 $-$ & 87.76 & \cellcolor{g}86.40 $+$ & 86.31 $+$ & \cellcolor{p}63.69 $-$ & 71.80 $\sim$ & 69.37 $\sim$ & 71.62 \\
\texttt{irs} & 99.98 $+$ & \cellcolor{g}100.0 $+$ & \cellcolor{p}84.17 $-$ & 94.81 $-$ & 95.51 $\sim$ & 95.26 & \cellcolor{g}94.89 $\sim$ & 94.00 $\sim$ & \cellcolor{p}81.78 $-$ & 94.00 $\sim$ & \cellcolor{g}94.89 $\sim$ & 94.22 \\
\texttt{lnd} & \cellcolor{g}100.0 $+$ & \cellcolor{g}100.0 $+$ & 55.29 $+$ & \cellcolor{p}32.05 $-$ & 39.18 $-$ & 46.27 & 50.78 $+$ & \cellcolor{g}58.92 $+$ & 42.06 $\sim$ & \cellcolor{p}38.24 $\sim$ & 40.10 $\sim$ & 40.20 \\
\texttt{mam} & \cellcolor{g}94.79 $+$ & 93.08 $+$ & 83.10 $+$ & 80.89 $-$ & \cellcolor{p}77.89 $-$ & 81.07 & 79.18 $\sim$ & 78.83 $\sim$ & \cellcolor{g}80.72 $\sim$ & 79.73 $\sim$ & \cellcolor{p}76.77 $-$ & 80.03 \\
\texttt{pdy} & \cellcolor{g}100.0 $+$ & 99.95 $+$ & 89.16 $+$ & 50.94 $-$ & \cellcolor{p}50.70 $-$ & 53.94 & \cellcolor{g}95.37 $+$ & 94.91 $+$ & 88.24 $+$ & 51.46 $-$ & \cellcolor{p}51.36 $-$ & 54.63 \\
\texttt{pis} & 99.99 $+$ & \cellcolor{g}100.0 $+$ & 87.54 $+$ & 86.89 $\sim$ & \cellcolor{p}85.53 $-$ & 86.91 & \cellcolor{g}86.90 $+$ & 86.79 $+$ & 86.12 $\sim$ & 85.88 $\sim$ & \cellcolor{p}84.67 $-$ & 85.92 \\
\texttt{pha} & \cellcolor{g}99.99 $+$ & 99.93 $+$ & \cellcolor{p}79.17 $-$ & 82.10 $-$ & 86.80 $+$ & 85.64 & 65.64 $\sim$ & 63.48 $-$ & \cellcolor{p}62.55 $-$ & \cellcolor{g}66.96 $+$ & 64.41 $\sim$ & 66.13 \\
\texttt{pre} & \cellcolor{g}100.0 $+$ & \cellcolor{g}100.0 $+$ & 99.63 $\sim$ & \cellcolor{p}98.01 $-$ & \cellcolor{g}100.0 $+$ & 99.76 & \cellcolor{g}100.0 $+$ & \cellcolor{g}100.0 $+$ & 99.64 $\sim$ & \cellcolor{p}98.03 $-$ & \cellcolor{g}100.0 $+$ & 99.79 \\
\texttt{pmp} & \cellcolor{g}99.99 $+$ & 99.97 $+$ & 87.24 $\sim$ & 86.99 $\sim$ & \cellcolor{p}86.89 $-$ & 87.03 & \cellcolor{g}88.39 $+$ & 87.57 $+$ & \cellcolor{p}86.01 $\sim$ & 86.17 $\sim$ & 86.05 $\sim$ & 86.21 \\
\texttt{rsn} & \cellcolor{g}100.0 $+$ & 99.89 $+$ & 85.02 $\sim$ & \cellcolor{p}84.33 $-$ & 85.51 $+$ & 85.13 & \cellcolor{g}86.48 $+$ & 86.07 $+$ & 84.63 $\sim$ & \cellcolor{p}84.19 $-$ & 85.30 $\sim$ & 85.07 \\
\texttt{seg} & \cellcolor{g}100.0 $+$ & \cellcolor{g}100.0 $+$ & 95.40 $+$ & \cellcolor{p}90.10 $-$ & 90.52 $\sim$ & 90.42 & 98.31 $+$ & \cellcolor{g}98.67 $+$ & 93.02 $+$ & \cellcolor{p}89.88 $-$ & 90.01 $\sim$ & 90.30 \\
\texttt{sir} & \cellcolor{g}100.0 $+$ & \cellcolor{g}100.0 $+$ & \cellcolor{p}76.96 $-$ & 84.78 $\sim$ & 85.00 $\sim$ & 84.78 & \cellcolor{g}85.33 $\sim$ & 83.33 $\sim$ & \cellcolor{p}70.67 $-$ & 81.67 $\sim$ & 81.33 $\sim$ & 82.00 \\
\texttt{smk} & \cellcolor{g}100.0 $+$ & 99.99 $+$ & 97.66 $+$ & 95.39 $-$ & \cellcolor{p}93.49 $-$ & 95.71 & \cellcolor{g}99.29 $+$ & 99.26 $+$ & 96.50 $+$ & 95.08 $\sim$ & \cellcolor{p}93.14 $-$ & 95.11 \\
\texttt{tae} & \cellcolor{g}96.91 $+$ & 96.37 $+$ & \cellcolor{p}59.23 $-$ & 63.63 $-$ & 64.15 $\sim$ & 64.35 & \cellcolor{g}66.88 $+$ & 65.00 $+$ & \cellcolor{p}49.17 $-$ & 54.58 $\sim$ & 53.75 $\sim$ & 56.46 \\
\texttt{tip} & \cellcolor{g}91.85 $+$ & 89.64 $+$ & 82.71 $+$ & 81.09 $-$ & \cellcolor{p}80.24 $-$ & 81.20 & 79.53 $\sim$ & 80.35 $\sim$ & \cellcolor{g}80.47 $\sim$ & 80.00 $-$ & \cellcolor{p}78.88 $-$ & 80.23 \\
\texttt{tit} & \cellcolor{g}95.93 $+$ & 93.28 $+$ & 73.25 $+$ & 72.28 $\sim$ & \cellcolor{p}70.57 $-$ & 72.38 & 70.81 $\sim$ & 69.96 $\sim$ & \cellcolor{g}72.30 $\sim$ & 71.30 $\sim$ & \cellcolor{p}68.93 $-$ & 71.52 \\
\texttt{wne} & \cellcolor{g}100.0 $+$ & \cellcolor{g}100.0 $+$ & \cellcolor{p}93.62 $-$ & 98.31 $-$ & 98.58 $\sim$ & 98.79 & \cellcolor{g}98.15 $\sim$ & 96.48 $\sim$ & \cellcolor{p}88.52 $-$ & 96.48 $\sim$ & 96.11 $\sim$ & 96.67 \\
\texttt{wbc} & \cellcolor{g}100.0 $+$ & \cellcolor{g}100.0 $+$ & 97.59 $+$ & 97.12 $\sim$ & \cellcolor{p}96.42 $-$ & 97.07 & \cellcolor{g}96.67 $+$ & \cellcolor{p}95.29 $\sim$ & 96.19 $\sim$ & 96.19 $\sim$ & \cellcolor{p}95.29 $\sim$ & 95.95 \\
\texttt{wpb} & \cellcolor{g}100.0 $+$ & \cellcolor{g}100.0 $+$ & 95.21 $+$ & \cellcolor{p}91.57 $-$ & 93.60 $+$ & 92.58 & 79.50 $+$ & \cellcolor{g}81.17 $+$ & \cellcolor{p}70.00 $\sim$ & 72.17 $\sim$ & 71.67 $\sim$ & 72.00 \\
\texttt{yst} & \cellcolor{g}100.0 $+$ & 99.86 $+$ & 60.73 $+$ & \cellcolor{p}53.26 $-$ & 56.22 $-$ & 58.78 & \cellcolor{g}62.80 $+$ & 59.60 $+$ & 56.89 $\sim$ & \cellcolor{p}52.04 $-$ & 53.65 $-$ & 57.61 \\
\bhline{1pt}
Rank & \cellcolor{g}\textit{1.38}$\uparrow^{\dag\dag}$ & \textit{1.65}$\uparrow^{\dag\dag}$ & \textit{4.10}$\downarrow$ & \cellcolor{p}\textit{4.98}$\downarrow^{\dag\dag}$ & \textit{4.90}$\downarrow^{\dag\dag}$ & \textit{3.98} & \cellcolor{g}\textit{1.63}$\uparrow^{\dag\dag}$ & \textit{2.77}$\uparrow^{\dag\dag}$ & \textit{4.42}$\downarrow^{\dag\dag}$ & \textit{4.07}$\downarrow^{\dag\dag}$ & \cellcolor{p}\textit{4.70}$\downarrow^{\dag\dag}$ & \textit{3.42} \\
Position & \textit{1} & \textit{2} & \textit{4} & \textit{6} & \textit{5} & \textit{3} & \textit{1} & \textit{2} & \textit{5} & \textit{4} & \textit{6} & \textit{3} \\
$+/-/\sim$ & 30/0/0 & 30/0/0 & 17/10/3 & 1/20/9 & 5/18/7 & - & 19/0/11 & 18/1/11 & 4/11/15 & 1/8/21 & 1/11/18 & - \\
\bhline{1pt}
$p$-value & 1.86E-09 & 1.86E-09 & 0.968 & 4.80E-05 & 0.00148 & - & 1.07E-05 & 0.00385 & 0.0483 & 0.0121 & 0.000529 & - \\
$p_\text{Holm}$-value & 9.31E-09 & 9.31E-09 & 0.968 & 0.000144 & 0.00297 & - & 5.37E-05 & 0.0116 & 0.0483 & 0.0243 & 0.00211 & - \\

\bhline{1pt}

\end{tabular}
}
\end{center}
\end{table*}
{
Table \ref{tb: sup sota} presents each system's average training and test classification accuracy, respectively, and average rank across all datasets.
``$+$'', ``$-$'', ``$\sim$'', ``Rank'', ``Position'', arrows, ``$\dag(\dag\dag)$'', ``$p$-value'', ''$p_\text{Holm}$-value'', and green- and peach-shaded values should be interpreted as in Tables \ref{tb: sup exstracs 10epoch} and \ref{tb: sup exstracs 50epoch}.

From Table \ref{tb: sup sota}, Random Forest, XGBoost, and \ds\ ranked first, second, and third in average rank during both training and testing phases. Notably, during training, Random Forest and XGBoost achieved 100\% classification accuracy on many datasets.

These results highlight the fundamental differences in design philosophy between these approaches. Random Forest and XGBoost are ensemble learning techniques that aggregate multiple complex decision trees, excelling in capturing intricate patterns within data. In contrast, \all\ (an LFCS) is designed to prioritize interpretability through the use of fuzzy rules with easy-to-understand linguistic terms.

\begin{table*}[t]
\begin{center}
\caption{Results Displaying Average Model Complexity Across 30 Runs. The columns Describe: Number of Trees per Model ($\#$\textsc{Trees}), Average Depth per Tree (\textsc{Depth}), Number of Nodes per Model ($\#$\textsc{Nodes}) for Random Forest and XGBoost; and Number of Macro-Rules ($\#$\textsc{Macro-Rules}) for UCS and \all.}
\vspace{2mm}
\label{tb: sup vs rfxgb}
\scalebox{0.8}{
\begin{tabular}{c|rrr|rrr|rr}
\bhline{1pt}
&\multicolumn{3}{c|}{Random Forest} & \multicolumn{3}{c|}{XGBoost} & UCS & \all \\
&$\#$\textsc{Trees} & \textsc{Depth} & $\#$\textsc{Nodes} & $\#$\textsc{Trees} & \textsc{Depth} & $\#$\textsc{Nodes} & \multicolumn{2}{c}{$\#$\textsc{Macro-Rules}} \\
\bhline{1pt}
\texttt{bnk} & 100 & 8.2  & 5144  & 100 & 2.3  & 800  & 1285  & 1244  \\
\texttt{can} & 100 & 7.3  & 3872  & 100 & 2.0  & 646  & 1750  & 1744  \\
\texttt{car} & 100 & 5.3  & 2181  & 100 & 2.2  & 599  & 219  & 258  \\
\texttt{col} & 100 & 10.2  & 6770  & 300 & 3.3  & 3279  & 1275  & 1084  \\
\texttt{dbt} & 100 & 15.2  & 23688  & 100 & 6.0  & 3400  & 1584  & 1546  \\
\texttt{ecl} & 100 & 11.0  & 9637  & 800 & 2.1  & 5291  & 1467  & 1357  \\
\texttt{frt} & 100 & 12.4  & 13842  & 700 & 2.3  & 5501  & 1826  & 1858  \\
\texttt{gls} & 100 & 10.6  & 7554  & 600 & 2.2  & 4114  & 1430  & 1313  \\
\texttt{hrt} & 100 & 10.1  & 9402  & 100 & 4.3  & 1677  & 1622  & 1620  \\
\texttt{hpt} & 100 & 11.1  & 6544  & 100 & 4.0  & 1282  & 1683  & 1690  \\
\texttt{hcl} & 100 & 10.8  & 8860  & 100 & 3.9  & 1367  & 1777  & 1786  \\
\texttt{irs} & 100 & 5.2  & 1554  & 300 & 1.3  & 1138  & 477  & 357  \\
\texttt{lnd} & 100 & 13.7  & 24562  & 500 & 5.6  & 11724  & 1374  & 723  \\
\texttt{mam} & 100 & 17.7  & 31553  & 100 & 6.0  & 3510  & 1478  & 1412  \\
\texttt{pdy} & 100 & 19.6  & 105739  & 400 & 6.0  & 18517  & 1403  & 912  \\
\texttt{pis} & 100 & 16.0  & 33351  & 100 & 6.0  & 4178  & 1668  & 1704  \\
\texttt{pha} & 100 & 20.0  & 36535  & 100 & 6.0  & 3959  & 1692  & 1725  \\
\texttt{pre} & 100 & 10.1  & 9681  & 300 & 0.6  & 793  & 958  & 990  \\
\texttt{pmp} & 100 & 19.4  & 38273  & 100 & 6.0  & 3994  & 1630  & 1669  \\
\texttt{rsn} & 100 & 15.1  & 16520  & 100 & 6.0  & 2865  & 1512  & 1464  \\
\texttt{seg} & 100 & 14.8  & 17913  & 700 & 2.2  & 5582  & 1701  & 1709  \\
\texttt{sir} & 100 & 6.2  & 2280  & 100 & 2.1  & 574  & 444  & 441  \\
\texttt{smk} & 100 & 7.6  & 3527  & 100 & 2.5  & 712  & 1425  & 1482  \\
\texttt{tae} & 100 & 12.0  & 9248  & 300 & 5.4  & 5969  & 1073  & 908  \\
\texttt{tip} & 100 & 20.1  & 78694  & 100 & 6.0  & 5626  & 1569  & 1615  \\
\texttt{tit} & 100 & 19.8  & 39364  & 100 & 6.0  & 3601  & 1538  & 1521  \\
\texttt{wne} & 100 & 5.2  & 2068  & 300 & 0.6  & 785  & 1396  & 1340  \\
\texttt{wbc} & 100 & 9.1  & 5666  & 100 & 3.8  & 1098  & 1538  & 1527  \\
\texttt{wpb} & 100 & 8.8  & 4820  & 100 & 2.9  & 887  & 1826  & 1782  \\
\texttt{yst} & 100 & 24.0  & 83275  & 1000 & 4.8  & 25418  & 1632  & 1548  \\
\bhline{1pt}
Average & 100 & 12.6 & 21404 & 267 & 3.8 & 4296 & 1408 & 1344 \\
\bhline{1pt}
\end{tabular}
}
\end{center}
\end{table*}

\label{r1-2-2}
A recent survey on explainable artificial intelligence (XAI) \cite{ortigossa2024explainable} emphasizes that while methods like Random Forest and XGBoost offer superior learning performance compared to rule-based systems (including LFCSs), they provide significantly lower model interpretability. This underscores the trade-off between performance and interpretability in machine learning models. {Table \ref{tb: sup vs rfxgb} presents a quantitative comparison of model complexity across different methods:
\begin{itemize}
    \item Random Forest generated an average of 100 trees per model, with an average depth of 12.6 per tree, resulting in an average of 21,404 decision nodes per model.
    \item XGBoost generated an average of 267 trees per model, with an average depth of 3.8 per tree, resulting in an average of 4,296 decision nodes per model.
    \item UCS generated an average of 1,408 non-linguistic macro-rules.
    \item \all\ generated an average of 1,344 linguistic macro-rules.
\end{itemize}
This comparison demonstrates that while Random Forest and XGBoost achieve generally better performance, they lead to significantly higher model complexity. The linguistic rules in \all\ provide interpretable decision-making with good performance while maintaining lower model complexity.

Thus, }our proposed \ds\ system strikes a balance between these competing objectives. While it may not achieve the same classification accuracy as Random Forest or XGBoost on some datasets, it offers enhanced interpretability and the ability to quantify uncertainty. These features are crucial in scenarios where understanding the decision-making process is as important as the final classification result.

It is important to emphasize that the primary contribution of our work is the improvement of the decision-making mechanism in LFCSs. By incorporating the DS theory-based class inference scheme, we enhanced the ability of LFCSs to handle uncertainty and provide more robust classifications, all while maintaining the interpretability that is a characteristic of these systems.

These findings highlight the trade-off between interpretability and classification performance. While \ds\ may not outperform Random Forest or XGBoost in classification accuracy, it offers unique advantages in managing uncertainty and providing interpretable results. This makes it particularly valuable in domains where decision transparency is crucial, such as healthcare or finance. Future work includes the proposed DS theory-based class inference scheme with established methods like Random Forest and XGBoost, which could further enhance its performance across diverse datasets. 
}

\end{document}